\documentclass[preprint,review,12pt]{elsarticle}
\usepackage[margin=2.5cm]{geometry} 

\usepackage[utf8]{inputenc} 
\usepackage[T1]{fontenc}    
\usepackage{hyperref}       
\hypersetup{
    colorlinks=true,       
    linkcolor=black,          
    citecolor=blue,        
    urlcolor=blue           
}
\usepackage{url}            
\usepackage{booktabs}       
\usepackage{amsfonts}       
\usepackage{nicefrac}       
\usepackage{microtype}      
\usepackage{lipsum}		
\usepackage{graphicx}
\usepackage{doi}
\usepackage{amsmath}
\usepackage{bm}
\usepackage{subfig}

\usepackage[english]{babel}
\usepackage{blindtext}
\usepackage{amssymb}
\usepackage{cleveref}
\usepackage{comment}
\usepackage[numbers]{natbib}
\usepackage[linesnumbered,ruled,vlined]{algorithm2e}
\usepackage[table,xcdraw]{xcolor}
\definecolor{newcolor}{rgb}{.8,.349,.1}
\usepackage{amsthm}
\theoremstyle{definition}

\usepackage{graphicx}
\journal{}

\begin{document}

\begin{frontmatter}



\title{Non-overlapping, Schwarz-type Domain Decomposition Method for Physics and Equality Constrained Artificial Neural Networks}


\author[inst1]{Qifeng Hu}
\ead{qih56@pitt.edu}

\author[inst1]{Shamsulhaq Basir}
\ead{shb105@pitt.edu}

\author[inst1]{Inanc Senocak \corref{cor1}}
\cortext[cor1]{corresponding author:~senocak@pitt.edu (Inanc Senocak)}

\affiliation[inst1]{organization={Department of Mechanical Engineering and Materials Science, \\ University of Pittsburgh},
            city={Pittsburgh},
            postcode={15261}, 
            state={PA},
            country={USA}}
            

\begin{abstract}
We present a non-overlapping, Schwarz-type domain decomposition method with a generalized interface condition, designed for physics-informed machine learning of partial differential equations (PDEs) in both forward and inverse contexts. Our approach employs physics and equality-constrained artificial neural networks (PECANN) within each subdomain. Unlike the original PECANN method, which relies solely on initial and boundary conditions to constrain PDEs, our method uses both boundary conditions and the governing PDE to constrain a unique interface loss function for each subdomain. This modification improves the learning of subdomain-specific interface parameters while reducing communication overhead by delaying information exchange between neighboring subdomains.
To address the constrained optimization in each subdomain, we apply an augmented Lagrangian method with a conditionally adaptive update strategy, transforming the problem into an unconstrained dual optimization. A distinct advantage of our domain decomposition method is its ability to learn solutions to both Poisson’s and Helmholtz equations, even in cases with high-wavenumber and complex-valued solutions. Through numerical experiments with up to 64 subdomains, we demonstrate that our method consistently generalizes well as the number of subdomains increases.
\end{abstract}


\end{frontmatter}

\section{Introduction}
Deep learning with artificial neural networks (ANNs) has transformed many fields of science and engineering. The functional expressivity of ANNs was established by universal approximation theory \cite{hornik1989multilayer}. Since then, ANNs have emerged as a meshless method to solve partial differential equations (PDEs) for both forward and inverse problems \cite{dissanayake1994neural, van1995neural, Monterola1998lagrange, lagaris1998artificial}. With the introduction of easily accessible software tools for automatic differentiation and optimization, the use of ANNs to solve PDEs has grown rapidly in recent years as physics-informed neural networks (PINNs) \cite{Raissi2019, Karniadakis2021}. Numerous works have been published since the introduction of the PINN framework to address the shortcomings of the framework as well as expand it with different features such as uncertainty quantification. 

PINNs offer several advantages over conventional numerical methods such as the finite element and volume methods when applied to data-driven modeling, inverse and parameter estimation problems. Unlike conventional numerical methods that have been developed and advanced over several decades as predictive-science techniques for challenging problems, PINNs have thus far been mostly applied to two-dimensional problems. Several issues stand in the way of extending PINNs to large, three-dimensional, multi-physics problems, including difficulties with nonlinear non-convex optimization, respecting conservation laws strictly, and long training times. In the present work, we focus on the application of domain decomposition methods (DDM) to PINNs, which are motivated by solving forward and inverse problems that can be computationally large and may involve multiple physics.  

Domain decomposition has become an essential strategy for solving complex PDE problems that are too large to be solved on a single computer or that have complex geometries with multiple physics. DDM can be constructed as overlapping or non-overlapping. The earliest instance of an overlapping domain decomposition method is attributed to \citet{schwarz1870uber}, whose work later became known as the alternating Schwarz method (ASM). The multiplicative Schwarz method is a generalization of ASM, while the additive Schwarz method \cite{additive_schwarz} introduces a modification that allows for parallel computations. In the additive approach, both subdomains are solved concurrently using information from the previous iteration. However, these variants of the Schwarz method are computationally slow and fail to converge when applied to non-overlapping subdomains \cite{dolean2002optimized}. Moreover, even for overlapping subdomains, these methods do not converge for acoustic problems \cite{gander2006optimized}. To address these limitations, \citet{japhet1998} optimized the transmission conditions, resulting in faster convergence. This variant of the method is now referred to as the optimized Schwarz method (OSM) \cite{gander2006optimized}. Since there are similarities between our proposed approach and OSM, we will introduce OSM in the next section. Detailed discussions on various approaches to domain decomposition can be found in textbooks on DDM \cite{smith1998domain, quarteroni1999DDM, dolean2015introduction}. 

Domain decomposition in the context of PINNs is a new and active research area that has been the subject of several recent works. 
\citet{li2019d3m} proposed an overlapping DDM for the DeepRitz method \cite{weinan2017proposal},  which is an alternative formulation of PINNs for learning the solution of PDEs. In their approach, the ASM with a Dirichlet-type overlapping interface condition was used, and the arising loss term was incorporated into the objective function of the DeepRitz method. \citet{li2020ddmelliptic} solved the Poisson's equation on overlapping decomposed domains with a complex interface using the baseline PINN approach. In their approach a classical ASM was used as well. The loss term arising from satisfying the interface conditions was added to the PINN's objective function in a composite fashion along with loss terms arising from the residual forms of the boundary conditions and the governing PDE.

\citet{jagtap2020conservative} decomposed a spatial domain into smaller domains and used the baseline PINN method to learn the solution of a PDEs on the whole domain. A separate neural network was adopted in each subdomain and flux continuity across subdomain interfaces were enforced in strong form. The average value of the solution between two subdomains sharing an interface was also enforced as an additional condition. Since, the neural network models associated for subdomains exchange information at each epoch, makes it not strictly a Schwarz-type DDM. In the spirit of the baseline PINNs, loss terms arising from the flux continuity across subdomain interfaces are lumped into a single composite objective function with tunable weights. In a followup work, \citet{jagtap2020xpinns} extended the work presented in \cite{jagtap2020conservative} to include the time domain. Furthermore, in this followup work, the interface conditions were simplified to make the method applicable to PDEs that may not represent conservation laws. \textcolor{black}{However, challenges related to the interface loss term in XPINN were acknowledged in a recent work and a trainable gate network with a flexible domain decomposition strategy was proposed \cite{HU2023_APINN}.} \citet{hu_xpinn_2022} explored the conditions and mechanisms by which Extended PINN (XPINN) with domain decomposition enhances generalization compared to PINN without domain decomposition. Their analysis indicated that domain decomposition can improve predictions by breaking the overall problem into simpler, more manageable subproblems. However, they also noted that this strategy reaches a limit as the availability of training points decreases with an increasing number of subdomains. A parallel implementation of XPINN was presented in \cite{shukla2021parallel_pinns} showing decent scalability and speedup. 

Recently, \citet{Moseley2023} introduced finite-basis physics-informed neural networks (FBPINNs) for solving PDEs on overlapping subdomains. FBPINNs draw inspiration from traditional finite element methods, in which the solution to a differential equation is represented as a sum of a finite number of basis functions that have compact support. In FBPINNs, neural networks are employed to learn these basis functions, which are defined across small, overlapping subdomains. In \citet{DOLEAN2024}, this approach was extended by adding multiple levels of domain decompositions to their solution ansatz, which was shown to improve the generalization of FBPINNs. However, to improve the accuracy of the local solutions, the authors also trained a neural network with hard constraints for the entire domain to serve as a coarse correction. 

Clearly, domain decomposition in the context of scientific machine learning or physics-informed neural networks is a growing area of focus, because it enables neural networks to tackle larger and complex problems or reduce training times substantially. Additionally, empirical evidence shows that training separate neural networks on smaller domains is much more feasible and likely to converge than training a single neural network on a large domain with many points \cite{hu_xpinn_2022}, a feature that we confirm in the present study as well. In what follows, we present the theory behind the optimized Schwarz methods \cite{gander2006optimized} as it inspires our proposed approach. We then propose a non-overlapping, Schwarz-type domain decomposition method with an interface condition with learnable, subdomain-specific parameters. We apply the resulting distributed learning method to forward and inverse PDE problems. Notably, our examples include Poisson's and Helmholtz equations, with high wavenumber and complex-valued solutions.  
\section{Optimized Schwarz method}\label{sec:osm}
Our proposed DDM for learning the solution of physics and equality constrained artificial neural networks \cite{PECANN_2022} has important parallels with the optimized Schwarz methods (OSM), but also differ from the OSM in major ways. Therefore, we briefly explain OSM and discuss some of the key works in OSM.  
\begin{figure}
    \centering
    \includegraphics[scale=0.6]{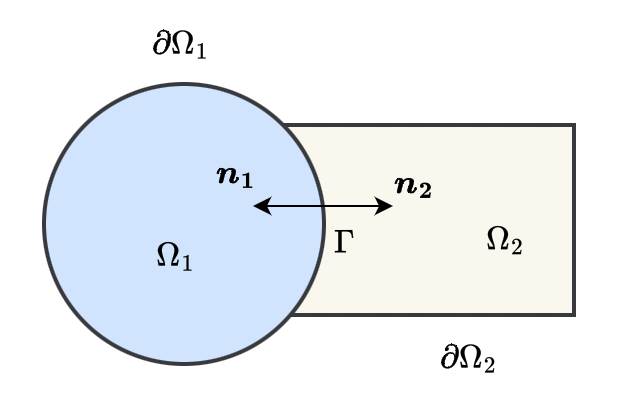}
    \caption{Domain decomposition with a non-overlapping subdomains}
    \label{fig:overlapping_non_overlapping_domains}
\end{figure}

Let us consider solving a second-order elliptic PDE on two non-overlapping subdomains, as shown in Fig. \ref{fig:overlapping_non_overlapping_domains}. On the first subdomain, we consider
\begin{equation}
\begin{aligned}
    -\Delta (u_1 ^{n+1})  &= s_1 &&\text{in} \quad \Omega_1,\\
    u_1^{n+1} &= 0  &&\text{on} \quad \partial \Omega_1 \cap \partial \Omega,\\
    (\mathcal{A}_1 + \beta_1 \frac{\partial}{\partial \boldsymbol{n_1}}) u_1 ^{n+1} &= (\mathcal{A}_1 - \beta_1 \frac{\partial}{\partial \boldsymbol{n_2}}) u_2 ^{n} &&\text{on} \quad \Gamma_1,
\end{aligned}
\end{equation}
and on the second subdomain
\begin{equation}
\begin{aligned}
    -\Delta (u_2 ^{n+1})  &= s_2 &&\text{in} \quad \Omega_2,\\
    u_2^{n+1} &= 0  &&\text{on} \quad \partial \Omega_2 \cap \partial \Omega,\\
    (\mathcal{A}_2 + \beta_2 \frac{\partial}{\partial \boldsymbol{n_2}}) u_2^{n+1} &= (\mathcal{A}_2 - \beta_2 \frac{\partial}{\partial \boldsymbol{n_1}}) u_1^{n} &&\text{on} \quad \Gamma_2,
\end{aligned}
\end{equation}
where $\Delta$ is the Laplacian operator applied to the function $u$, and $s_{1,2}$ are given source terms. $\boldsymbol{n_1}$ and $\boldsymbol{n_2}$ are the outward normal directions on the subdomain boundaries of $\Omega_1$ and $\Omega_2$, respectively. Note that $\boldsymbol{n_1}=-\boldsymbol{n_2}$. $\Gamma_1$ and $\Gamma_2$ represents the subdomain interfaces corresponding to $\Omega_1$ and $\Omega_2$, respectively. In the case of a non-overlapping domain decomposition $\Gamma_1$ and $\Gamma_2$ are identical.  $\mathcal{A}_1$ and $\mathcal{A}_2$ are operators act along the interfaces $\Gamma_1$ and $\Gamma_2$, respectively. $\beta_1$ and $\beta_2$ are real valued functions. With $\beta_1 = \beta_2 = 0$ and $\mathcal{A}_1$ and $\mathcal{A}_2$ being identity operators, the original Schwarz method is recovered. As a remedy to the drawbacks of classical Schwarz methods, \citet{lions1990schwarz} proposed to replace Dirichlet interface conditions with Robin interface conditions with a tunable parameter $\alpha$. In the above interface formulation, we see that with $\beta_1 = \beta_2 = 1$ and $\mathcal{A}_1 = \mathcal{A}_2 = \alpha$, where $\alpha>0.0$, we recover the Robin interface conditions proposed by \citet{lions1990schwarz}.

The essence of optimized Schwarz method (OSM) \cite{gander2006optimized} is to determine optimal operators $\mathcal{A}$ and the parameter $\beta$ such that the convergence rate of the Schwarz algorithm is maximized. This is often achieved by theoretically deriving an expression for the convergence rate for a representative problem with a simple decomposition (e.g. two subdomains) and optimizing the interface parameters with respect to that convergence rate. The extension of this approach to complex domains with challenging decompositions with many subdomains is admittedly a formidable task. However, numerical experiments have shown that optimal interface conditions, once derived from canonical problems, can be used in complex problems with a general decomposition, as shown in several works \cite{gander2002optimized, gander2007_two_sided_osm, nataf2007, MadayOSM2007}. 

\section{Technical Formulation}\label{sec:formulation}
Our primary objective is to determine the solutions of both forward and inverse partial differential equation (PDE) problems in a distributed manner using neural networks. To achieve this, we pursue a constrained optimization formalism and develop a non-overlapping Schwarz-type DDM and implement it using the Message Passing Interface (MPI) standard. Specifically, our proposed method extends the constrained optimization formulation of the Physics and Equality Constrained Artificial Neural Networks 
(PECANN) framework \cite{PECANN_2022} to a generalized interface loss term arising from domain decomposition. A distinctive feature of our approach is the simultaneous learning of the parameters of the interface conditions and the solution of the PDE within each subdomain. We demonstrate the distinct advantege of this by its ability to learn the solutions of both Poisson's and Helmholtz equations, with high wavenumber and complex-valued solutions. We also apply our method to inverse PDE problems.

\textcolor{black}{For ease of presentation, we consider the same prototype problem illustrated in Fig. \ref{fig:overlapping_non_overlapping_domains}, and benefit from the optimized interface conditions for convection-diffusion equations as proposed in the works of \citet{JAPHET2001} and \citet{dolean_optimized_2002}. Different from those works, in our interface condition, we ignore the term for convection and the second order tangential derivative to adapt the interface condition for Poisson's and Helmholtz equations.} For the first subdomain $\Omega_1$ we have  
\begin{equation}
\begin{aligned}
    \alpha_1 u_1 ^{n+1} + \beta_1 \frac{\partial u_1 ^{n+1}}{\partial \boldsymbol{n_1}} + \gamma_1 \frac{\partial u_1 ^{n+1}}{\partial \boldsymbol{\tau_1}}  &= \alpha_1 u_2 ^{n} + \beta_1\frac{\partial u_2 ^{n}}{\partial \boldsymbol{n_1}} + \gamma_1 \frac{\partial u_2 ^{n}}{\partial \boldsymbol{\tau_1}}  &&\text{on} \quad \Gamma, \label{eq:interface_cond_subdomain1}
\end{aligned}
\end{equation}
and for the second subdomain $\Omega_2$ we have
\begin{equation}
\begin{aligned}
    \alpha_2 u_2^{n+1} + \beta_2\frac{\partial u_2^{n+1}}{\partial \boldsymbol{n_2}} + \gamma_2 \frac{\partial u_2 ^{n+1}}{\partial \boldsymbol{\tau_2}}  &= 
    \alpha_2 u_1^{n} + \beta_2 \frac{\partial u_1^{n}}{\partial \boldsymbol{n_2}} + \gamma_2 \frac{\partial u_1^{n}}{\partial \boldsymbol{\tau_2}}  &&\text{on} \quad \Gamma, \label{eq:interface_cond_subdomain2}
\end{aligned}
\end{equation}
where $\alpha_{1,2}, \beta_{1,2}, \gamma_{1,2} > 0 $ are ``learnable'' scalar parameters of the transmission conditions imposed on the interface $\Gamma$ between adjacent subdomains $\Omega_1$ and $\Omega_2$, respectively. The superscript $n$ represents the iteration level.

We initialize the scalar parameters (i.e., $\alpha_{1,2}$, $\beta_{1,2}$ \& $\gamma_{1,2}$) as $1.0$ so as not to favor any of the terms in the interface condition. 
It is important to highlight that these parameters are independent and assigned separately for each subdomain, as the solution and its gradient may vary significantly across different domains during the learning phase. To this end, employing the same $\alpha$, $\beta$ and $\gamma$ for all the subdomains is not desirable, as it would impede convergence of the gradient-based optimizer.  We should mention that using different parameters in transmission conditions is not uncommon. For instance, \citet{gander2007_two_sided_osm} proposed a two-sided Robin condition for the Helmholtz equation on non-overlapping domains in which different parameters were adopted for the Dirichlet term in the Robin transmission conditions for each subdomain, which led to better convergence rates with two-sided Robin condition compared to using the same parameters in the Robin transmission condition.

\subsection{PECANN: Physics and Equality Constrained Artificial Neural Networks}\label{sec:pecanns}
In learning the solution of a PDE for the entire domain $\Omega$ in a distributed manner, each subdomain utilizes an independent neural network, but the architecture of the neural network remains consistent across all subdomains. As mentioned earlier, we adopt the constrained optimization formulation of the PECANN framework \cite{PECANN_2022}. However, extending this framework to incorporate distributed learning with a Schwarz-type domain decomposition method necessitates a new formulation. To this end, our formulation enables the learning of the parameters of the transmission conditions, constrained by the governing PDE and its boundary conditions. This approach is distinct from the original formulation adopted in the PECANN framework \cite{PECANN_2022}, where the solution to the PDE is learned solely under the constraints imposed by its boundary conditions.

To better generalize our formulation, the physical problem for the \(k\)th decomposed subdomain \(\Omega_k\) is defined using the following operators:
\begin{equation}
\begin{aligned}
    \mathcal{P}(\textbf{x}; \theta) &:= \mathcal{L}_{\textbf{x}}(u; \theta) - s(\textbf{x}), \quad \quad &&\forall \textbf{x} \in \Omega_k, \\
    \mathcal{D}(\textbf{x}; \theta) &:= u(\textbf{x}; \theta) - g(\textbf{x}), &&\forall \textbf{x} \in \partial_D \Omega_k \cup \Gamma_k, \\
    \mathcal{N}(\textbf{x}; \theta) &:= \frac{\partial u}{\partial \boldsymbol{n}}(\textbf{x}; \theta) - h(\textbf{x}), &&\forall \textbf{x} \in \partial_N \Omega_k \cup \Gamma_k, \\
    \mathcal{T}(\textbf{x}; \theta) &:= \frac{\partial u}{\partial \boldsymbol{\tau}}(\textbf{x}; \theta) - r(\textbf{x}), &&\forall \textbf{x} \in \Gamma_k. \label{eq:operators} \\
\end{aligned}
\end{equation}
Here, $\mathbf{x}$ represent the spatial vector, which may also include time \(t\) for time-dependent problems. We generalize the representation of a PDE by the differential operator \(\mathcal{L}_{\mathbf{x}}\) and the source term \(s(\mathbf{x})\). For a neural network model with learnable parameters $\theta$, 
the operator \(\mathcal{P}(\mathbf{x}; \theta)\) represents the residual form of the PDE. The operator \(\mathcal{D}(\mathbf{x}; \theta)\) imposes Dirichlet conditions on domain boundaries \(\partial_D \Omega_k\) and interfaces \(\Gamma_k\), with prescribed values \(g(\mathbf{x})\). Similarly, \(\mathcal{N}(\mathbf{x}; \theta)\) applies Neumann conditions on \(\partial_N \Omega_k\) and \(\Gamma_k\), with prescribed values \(h(\mathbf{x})\), while \(\mathcal{T}(\mathbf{x}; \theta)\) enforces continuity of tangential derivatives on \(\Gamma_k\), with prescribed values \(r(\mathbf{x})\). \textcolor{black}{Note that functions $g, h, r$ are initialized to zero on $\Gamma_k$ and can be prescribed specific values on domain boundary $\partial \Omega$.}

The subdomain interface condition operator $\mathcal{I}(\mathbf{x}; \theta)$ on $\Gamma_k$ can be formulated by a linear combination of the Dirichlet $\mathcal{D}$, Neumann $\mathcal{N}$, and tangential derivative continuity $\mathcal{T}$ operators with the interface condition parameters represented by the vector $\textbf{q} = [\alpha, \beta, \gamma]^T$. The prescribed values of these operators act on the information transferred from neighboring subdomains after communication:
\begin{equation}
\begin{aligned}
    g^{n+1}(\mathbf{x}) &\leftarrow u^{n}(\mathbf{x}; \theta_{adj}), && \forall \mathbf{x} \in \Gamma_k, \\
    h^{n+1}(\mathbf{x}) &\leftarrow \frac{\partial u^{n}}{\partial \boldsymbol{n}}(\mathbf{x}; \theta_{adj}), && \forall \mathbf{x} \in \Gamma_k, \\
    r^{n+1}(\mathbf{x}) &\leftarrow \frac{\partial u^{n}}{\partial \boldsymbol{\tau}}(\mathbf{x}; \theta_{adj}), && \forall \mathbf{x} \in \Gamma_k,
\end{aligned}
\label{eq:interface_info_commu}
\end{equation}
where \(\theta_{adj}\) represents the neural network parameters of the subdomain adjacent to $\Omega_k$. This information is summarized as a vector, \(\mathbf{f} = [g^{n+1}(\mathbf{x}), h^{n+1}(\mathbf{x}), r^{n+1}(\mathbf{x})]^T\).
In terms of inverse problems, measurement data are typically handled as a Dirichlet operator, denoted by \(\mathcal{M}(\textbf{x}; \theta)\).
Notation wise, these conditions are represented as follows:
\begin{equation}
\begin{aligned}
    \mathcal{B}(\textbf{x}; \theta) & =
\left\{\begin{array}{l}
    \mathcal{D}(\textbf{x}; \theta),\\
    \mathcal{N}(\textbf{x}; \theta),
\end{array}\right. && \begin{array}{l}
    \forall \textbf{x} \in \partial_D \Omega_k, \\
    \forall \textbf{x} \in \partial_N \Omega_k, 
\end{array}\\
    \mathcal{I}(\textbf{x}; \theta) &= \alpha \mathcal{D}(\textbf{x}; \theta) + \beta \mathcal{N}(\textbf{x}; \theta) + \gamma \mathcal{T}(\textbf{x}; \theta), \quad && \forall \textbf{x} \in \Gamma_k, \\
    \mathcal{M}(\textbf{x}; \theta) &= \mathcal{D}(\textbf{x}; \theta), \quad &&\forall \textbf{x} \in M_k \subset \Omega_k.
\end{aligned}
\end{equation}

With the mean squared error (MSE) metric as a distance function, we define the equality-constrained optimization problem for the subdomain \(\Omega_k\) as follows:
\begin{equation}
\begin{aligned}
    \min_{\theta} \quad & \mathcal{J}(\theta; \textbf{x}) = \frac{1}{N_{\Gamma_k}} \sum_{i=1}^{N_{\Gamma_k}} \|\mathcal{I}(\theta; \textbf{x}_i)\|_2^2 && ~\text{on} \quad \Gamma_k, \\
    \text{subject to} \quad & \mathcal{C}_B(\theta; \textbf{x}) = \frac{1}{N_{\partial \Omega_k}} \sum_{i=1}^{N_{\partial \Omega_k}} \| \mathcal{B}(\theta; \textbf{x}_i) \|_2^2 && ~\text{on} \quad \partial \Omega_k, \\
        & \mathcal{C}_P(\theta; \textbf{x}) = \frac{1}{N_{\Omega_k}} \sum_{i=1}^{N_{\Omega_k}} \| \mathcal{P}(\theta; \textbf{x}_i) \|_2^2 && ~\text{in} \quad \Omega_k, \\
        & \mathcal{C}_M(\theta; \textbf{x}) = \frac{1}{N_{M_k}} \sum_{i=1}^{N_{M_k}} \| \mathcal{M}(\theta; \textbf{x}_i) \|_2^2 && ~\text{in} \quad M_k ,
    \label{eq:constrained_problem}
\end{aligned}
\end{equation}
where \(\|\cdot \|_2\) is the Euclidean norm, $N_{\Gamma_k}, N_{\partial \Omega_k}$ and $ N_{\Omega_k}$ denote the number of collocation points on the subdomain interface with index $i$ representing the individual collocation point, domain boundaries, and in the subdomain, respectively. Additionally, $N_{M_k}$ represents the number of observed data for data-driven or inverse problems.

It is worthwhile to note a critical aspect of the formulation presented in Eq. \ref{eq:constrained_problem}. The objective function \(\mathcal{J}\) incorporates an interface transmission condition, while the constraint function \(\mathcal{C}_j\), $j \in \{B, P, M\}$, encompasses boundary and/or data constraints, along with physics constraints that represent the governing equation at hand. This feature sets our approach apart from the original PECANN \cite{PECANN_2022}, PINN \cite{raissi2019deep} and its distributed variant XPINN \cite{jagtap2020xpinns, hu_xpinn_2022} frameworks. \textcolor{black}{Notably, we treat the governing PDE as a constraint, rather than the subdomain interface conditions. We find this approach to be much more robust than using interface conditions as constraints, particularly because interface predictions can be inaccurate during the initial stages of training when the interface parameters are still undetermined.}

\subsection{Formulation of the Interface Loss Function}\label{sec:interface_loss}
To analyze the optimality conditions of the objective function \(\mathcal{J}\) with respect to the interface parameters in conditions given in Eq.~\ref{eq:interface_cond_subdomain1}, we consider the derivative of the interface loss at the $i$th collocation point \(\textbf{x}_i\) on a subdomain interface.
In summation form, the interface operator is rewritten as:
\begin{equation}
    \mathcal{I}_i = \sum_{j=1}^3 q^j \mathcal{O}_i^j,
\end{equation}
where $q^j$ is the $j$th interface parameter in $\textbf{q} = [\alpha, \beta, \gamma]^T$ and $\mathcal{O}_i^j$ is the corresponding interface operator in $\bm{\mathcal{O}}_i = [\mathcal{D}_i, \mathcal{N}_i, \mathcal{T}_i]^T$. Since we adopt the MSE as a metric, the loss function will be $\mathcal{J}_i = \mathcal{I}_i^2$.
The gradient of the loss function with respect to the $j$th interface parameter is given by:
\begin{equation}
    \frac{\partial \mathcal{J}_i}{\partial q^j} = 2 \mathcal{O}_i^j \mathcal{I}_i = 2 \mathcal{O}_i^j  \sum_{j=1}^3 q^j \mathcal{O}_i^j.
    \label{eq:optimality_objective_I}
\end{equation}
At optimality, we expect the above gradient of the loss function to vanish, meaning either \(\mathcal{I}_i\) or \(\mathcal{O}_i^j\) for the $i$th interface must be zero. The presence of \(\mathcal{I}_i\) in Eq. \ref{eq:optimality_objective_I} naturally couples the Dirichlet, Neumann, and continuity of tangential derivative conditions with interface parameters $\textbf{q}$. The presence of \(\mathcal{I}_i\) inherently couples the Dirichlet, Neumann, and continuity of tangential derivative conditions. This coupling implies that the elements of \(\mathbf{q}\) are not truly independent, hindering the attainment of optimality  in practice. To address this issue, we introduce an approximation to the interface loss function, as detailed below.

Let us start with expanding the interface loss as follows:
\begin{equation}
\mathcal{J}_i = \sum_{j=1}^3 (q^j \mathcal{O}_i^j)^2 + 2 \sum_{j\neq m} q^j q^m \mathcal{O}_i^j \mathcal{O}_i^m,
\label{eq:full_interface_loss}
\end{equation}
which we will refer to as the \textbf{full interface loss}. When taking the gradient of the loss function $\mathcal{J}_i$ with respect to interface parameters $\textbf{q}$, 
the cross-product terms (i.e. $q^j q^m \mathcal{O}_i^j \mathcal{O}_i^m$) in Eq. \ref{eq:full_interface_loss} are responsible for this problematic coupling issue.
To address this issue, we consider the following two alternatives to break the dependency and enhance convergence.

Our first approach involves using an absolute distance function metric to formulate a loss function. Although less efficient, the absolute distance function metric does not lead to the cross-product terms, thereby preventing the coupling of interface parameters. The loss function is expressed as:
\begin{equation}
\mathcal{J}_i = |\mathcal{I}_i| = \sum_{j=1}^3 q^j |\mathcal{O}_i^j|,
\label{eq:abs_interface_loss}
\end{equation}
which we refer to as the \textbf{absolute interface loss}. While this is not our proposed solution, the formulation illustrates an alternative means of representing the generalized interface condition as a loss term for the optimization problem.

For our second approach, we propose the following approximation to Eq.~\ref{eq:full_interface_loss}:
\begin{equation}
    \begin{aligned}
        \mathcal{J}_i &= \mathcal{I}_i^2 \approx \sum_{j=1}^3 (q^{j} \mathcal{O}_i^j)^2.
    \end{aligned}\label{eq:app_interface_loss}
\end{equation}
In arriving at the above approximation, we neglected the cross-product terms in Eq.~\ref{eq:full_interface_loss}. We will refer to Eq. \ref{eq:app_interface_loss} as \textbf{approximate interface loss}. With this approximation to the loss function, the optimality condition can now be expressed as
\begin{equation}
    \begin{aligned}
        \frac{\partial \mathcal{J}_i}{\partial q^j} &= 2 q^j (\mathcal{O}_i^j)^2.
    \label{eq:optimality_objective_final}
    \end{aligned}
\end{equation}
Again, at optimality, we expect each element of $\bm{\mathcal{O}}_i$ to vanish. Therefore, each element of $\textbf{q}$ does indeed become independent parameters, which aligns with our aim. We should also note that the independent learning of the interface parameters facilitates proper adaptation to the distinct physical scales or dimensions inherent in their corresponding operators. 

\textcolor{black}{We find the use Eq. \ref{eq:app_interface_loss} as a generalized interface loss term and the adaptability of its subdomain specific parameters to be key for effectively tackling both Poisson's and Helmholtz equations using the same PECANN framework. We note that the resulting objective function will be unique to the type of PDE problem because the interface loss is constrained by the target PDE and its boundary conditions.} Moreover, when $\mathcal{O}_i^j$ is small or converging toward to small values, a larger \(q^j\) helps maintain a large gradient, thus enhancing the convergence rate. This adjustment leverages the efficiency of the gradient-based optimizer, improving the overall performance of the solution process.
In other words, these parameters serve as multipliers for the Dirichlet, Neumann and tangential derivative continuity operators, respectively. However, it is crucial to emphasize that the method by which these parameters are predicted in our approach differs fundamentally from their determination in optimized Schwarz methods. 

\textcolor{black}{To further examine the convergence behavior of the interface parameters as multipliers, we analyze the ratios $\dfrac{\beta}{\alpha}$ and $\dfrac{\gamma}{\alpha}$ for the $k$th subdomain. These ratios are expected to asymptotically stabilize to constant values upon the method’s convergence. This expectation follows from Eq. \ref{eq:interface_cond_subdomain1} and Eq. \ref{eq:interface_cond_subdomain2}, where the interface conditions should be satisfied at convergence. Consequently, we can treat one of the parameters as a multiplier; normalizing both sides by this parameter yields two independent ratios that the network must learn.}

\subsection{Conditionally Adaptive Augmented Lagrangian Method}
With the approximation of Eq.~\ref{eq:app_interface_loss}, the final objective function for each subdomain becomes
\begin{equation}
    \begin{aligned}
        \min_{\theta} \max_{\textbf{q}>1} \mathcal{J}(\theta, \textbf{q}; \textbf{x})  \approx \frac{1}{N_{\Gamma_k}} \sum_{i=1}^{N_{\Gamma_k}} \sum_{j=1}^3 \left(q^{j} \mathcal{O}^j(\theta; \textbf{x}_i)\right)^2  && ~\text{on } \Gamma_k.
    \end{aligned}
    \label{eq:objective_final}
\end{equation}
We can cast the constrained optimization problem \eqref{eq:constrained_problem} into an unconstrained optimization problem using the augmented Lagrangian formalism \cite{hestenes1969multiplier, powell1969method} as follows:
\begin{equation}
    \min_{\theta} \max_{\textbf{q}>1,\bm{\lambda}} \mathcal{L}(\theta,\bm{\lambda},\textbf{q};\textbf{x}, \bm{\mu}) =  \mathcal{J}(\theta,\mathbf{q}; \textbf{x}) + \bm{\lambda}^T \bm{\mathcal{C}}(\theta; \textbf{x})  + \frac{1}{2} \bm{\mu}^T \big(\bm{\mathcal{C}}(\theta; \textbf{x}) \odot \bm{\mathcal{C}}(\theta; \textbf{x})\big), \label{eq:unconst_problem}
\end{equation}
where $\odot$ denotes an element-wise product, and $\bm{\lambda}$ and $\bm{\mu}$ are the vectors of Lagrange multipliers and penalty parameters corresponding to the vector of constraints $\bm{\mathcal{C}} = [\mathcal{C}_B, \mathcal{C}_P, \mathcal{C}_D]^T$, respectively. Note that unlike in the conventional ALM, we employ a vector containing of independent penalty parameters to address the different characteristics of the constraints. 

\textcolor{black}{The minimization of Eq.~\ref{eq:unconst_problem} can be performed using a variant of gradient descent type optimizer for Lagrange multipliers, inspired by the adaptive augmented Lagrangian method in \citet{basir2023adaptive}.
This process is outlined in Algorithm~\ref{alg:adaptive_training_algorithm}.
In each outer iteration of the algorithm, a specific unconstrained optimization problem, given \(\bm{\lambda}^{p-1}\) and \(\bm{\mu}^{p-1}\), is solved to satisfy a convergence criterion over a limited number of inner epochs $P$ through training.
The solution is considered converged if the current augmented Lagrangian loss function \(\mathcal{L}^p\) exceeds a fraction \(\omega\) (default value 0.999) of the previous loss \(\mathcal{L}^{p-1}\). Upon convergence, updates to the Lagrange multipliers and penalty parameters are made, and the process proceeds to the next unconstrained optimization problem with updated \(\bm{\lambda}^p\) and \(\bm{\mu}^p\). 
For domain decomposition, an additional condition requires that the optimization progress meets a minimum epoch threshold, \( epoch_{min} \), to prevent excessive communication between the adjacent subdomains.}

\textcolor{black}{The update of the penalty parameters \(\bm{\mu}\) is adaptive based on the information from the individual constraints. The term \(\bar{\textbf{v}}\) measures the weighted moving average of the squared constraints, initialized at zero, with \(\zeta\) (default value 0.99) as the smoothing constant. \(\bm{\mu}\) is increased by multiplying it with the ratio of the current constraints to the square root of \(\bar{\textbf{v}}\) when this ratio exceeds 1, but the adjustments are capped by \(\bm{\mu}_{\text{max}}\).
The vector of penalty scaling factors \(\bm{\eta}\) regulates the upper bounds of the penalty parameters $\bm{\mu}_{max}$. It uses default values of 1 for boundary condition (BC) and data constraints, and a smaller value of \(10^{-2}\) for the physics (PDE) constraint. This configuration biases the optimization process towards stricter satisfaction of the boundary and data constraints over the PDE constraints. 
A small constant \(\epsilon\) (set to \(10^{-16}\)) is added to the denominator to ensure numerical stability and prevent division by zero.}

\IncMargin{1em}
\begin{algorithm}[!h]
\SetAlgoLined
\SetKwInOut{KwInput}{Input}
\SetKwInOut{KwOutput}{Output}
\SetKwInOut{KwDefaults}{Defaults}
\KwDefaults{$\bm{\eta} = [1, 10^{-2}, 1]^T, ~\zeta = 0.99,~\omega = 0.999,~ \epsilon = 10^{-16}$}

\KwInput{$\theta^0$, number of outer iterations $N$, number of inner epochs $P$}
$\bm{\lambda}^0 = \textbf{1}$ \hspace{9em}
\tcc{Initializing Lagrange multipliers}
$\bm{\mu}^0 = \textbf{1}$ \hspace{9em}
\tcc{Initializing penalty parameters}
$\bm{\mu}_{\max}^0 = \textbf{1}$ \hspace{8em}
\tcc{Initializing penalty bounds}
$\bar{\textbf{v}}^0 = \textbf{0}$ \hspace{9em}
\tcc{Initializing averaged square-gradients}

\BlankLine
\For{$n \gets 1  ~ \KwTo ~N$}{
    $\mathcal{L}^0 \gets \infty$ \hspace{7em} 
    \tcc{Initializing augmented Lagrangian loss}
    \For{$p\gets 1  ~ \KwTo ~P$}{
        $\theta_k^p \gets \text{Update}(\theta_k^{p-1})$\hspace{11em}
        \tcc{primal update}
        $\mathcal{L}^p \gets$  Eq.~\ref{eq:unconst_problem} with $\theta_k^p, \bm{\lambda}^{p-1}, \bm{\mu}^{p-1}, \cdots$\hspace{2em}
        \tcc{augmented loss update}
        $\bar{\textbf{v}}^p  \xleftarrow{} \zeta ~\bar{\textbf{v}}^{p-1} + (1 - \zeta)~
        \big(\bm{\mathcal{C}}(\theta^p) \odot \bm{\mathcal{C}}(\theta^p)\big)$\hspace{1em}
        \tcc{square-gradient update}
        \uIf{$\mathcal{L}^p >= \omega \mathcal{L}^{p-1}$ \textbf{OR} $p ==P$}{
            $\bm{\lambda}^p \xleftarrow{} \bm{\lambda}^{p-1} + \bm{\mu}^{p-1} \bm{\mathcal{C}}(\theta^p)$ \hspace{6em}
            \tcc{dual update}
            $\bm{\mu}_{\max}  \xleftarrow{} \max(\bm{\mu}_{\max}, \dfrac{\bm{\eta}}{\sqrt{\bar{\textbf{v}}^p} + \epsilon})$\hspace{5em}
            \tcc{penalty bound update}
            $\bm{\mu}^p \xleftarrow{} \min\big( \bm{\mu}_{\max}, \max(\dfrac{\bm{\mathcal{C}}(\theta^p)}{\sqrt{\bar{\textbf{v}}^p} + \epsilon},\textbf{1}) ~\bm{\mu}^{p-1}\big)$ \hspace{0em}
            \tcc{penalty update}
            \uIf{$p \geq epoch_{min}$}{
                \textbf{break}
                }
            }
        }
    }
\KwOutput{$\theta^N$}
\caption{\textcolor{black}{Conditionally Adaptive Augmented Lagrangian Method}}
\label{alg:adaptive_training_algorithm}
\end{algorithm}

It is crucial to highlight the robustness of the PECANN framework in formulating and solving both forward and inverse PDE problems through a constrained optimization formalism. This stands in stark contrast to the unconstrained optimization formalism adopted in the PINN approach. PECANN strictly adheres to a constrained optimization formalism, from which an equivalent dual unconstrained optimization problem is derived using the augmented Lagrangian method \cite{hestenes1969multiplier, powell1969method}.
This distinctive approach enables PECANN to systematically integrate diverse constraints into the learning process, eliminating the need for heuristic methods to balance terms within a composite objective function, as would be required in an unconstrained optimization framework.

\subsection{Proposed Domain Decomposition Method}
\textcolor{black}{Algorithm~\ref{alg:ddm} presents our domain decomposition training procedure for learning the solution of PDEs in a distributed computing setting using the PECANN framework with a conditionally adaptive augmented Lagrangian method. 
The algorithm begins by taking inputs such as domain $\Omega$, collocation points $D$, the number of subdomains $K$, the number of inner epochs $P$ for local training, the number of outer iterations $N$ for interface communication, and the local neural network parameter $\theta$.
Each computing processor (rank), indexed by $k$, initializes its respective subdomain $\Omega_k$ and collocation points $D_k$ by partitioning the global problem.
It then sets up a local model with parameters $\theta_k^0$ with random Xavier initialization scheme \cite{pmlr-v9-glorot10a}. and initializes the interface parameters $\textbf{q}_k$ and interface information $\textbf{f}_k$.
Additionally, each rank initializes Lagrange multipliers $\bm{\lambda}$ and penalty parameters $\bm{\mu}$ for managing the various constraint functions specific to its subdomain.}

\textcolor{black}{The training process is conducted in parallel across all ranks. Each local model is independently trained for at least \(epoch_{min}\) inner epochs using the previously provided interface information. At the end of each local training cycle or upon reaching convergence (i.e. line 13 in Alg.~\ref{alg:ddm}), the algorithm exchanges locally optimized interface data--comprising the predicted solution, fluxes, and tangential derivatives--with neighboring subdomains, following the relationship specified in Eq.~\ref{eq:interface_info_commu}. This exchange only takes place after each local training cycle and is repeated for \(N\) outer iterations.
The independent training of subdomain networks for a fixed number of inner iterations defers communication overhead to the end of the local training and ensures that each subdomain model is specifically adapted to its segment of the problem, without the immediate influence of neighboring subdomains. This method enhances both the specificity and efficiency of the learning process across the distributed network}

The output from this iterative and parallel training process is a set of trained local neural network models, each finely tuned to its part of the global problem, but collectively forming a coherent solution across the entire domain.

\begin{algorithm}[!h]
    \SetKwInOut{Input}{Input}
    \SetKwInOut{Output}{Output}
    \SetKwInOut{Defaults}{Defaults}

    \Input{Domain $\Omega$, collocation points $D$, number of subdomains $K$, number of inner epochs $P$, number of outer iterations $N$, local model parameters $\theta$}

    \For{each rank $k \in \{1, 2, ..., K\}$}{
        $\Omega_k \& D_k \gets \Omega \& D$ \hspace{2em}
        \tcc{Assigning subdomain \& collocation points}
        $\theta_k^0$ \hspace{10em}
        \tcc{Initializing local model parameters} 
        $\textbf{q}_k^0 = \textbf{1}$ \hspace{9em}
        \tcc{Initializing interface parameters}
        $\textbf{f}_k^0 = \textbf{0}$ \hspace{9em}
        \tcc{Initializing interface information}
        $\bm{\lambda}_k^0 = \textbf{1}$\hspace{9em}
        \tcc{Initializing Lagrange multipliers}
        $\bm{\mu}_k^0 = \textbf{1}$\hspace{9em}
        \tcc{Initializing penalty parameters}
    }
    
    \For{$n \gets 1$ to $N$}{
        \For{each rank $k \in \{1, 2, ..., K\}$}{
            \For{$p \gets 1$ to $P$}{
                $\theta_k^p \gets \text{Update}(\theta_k^{p-1})$ \\
                $\textbf{q}_k^p \gets \text{Update}(\textbf{q}_k^{p-1})$ \\
                
                \uIf{$\mathcal{L}_k^p >= \omega \mathcal{L}_k^{p-1}$ \textbf{OR} $p ==P$}{
                    Update $\bm{\lambda}_k^p, \bm{\mu}_k^p$  \hspace{4em}\tcc{via lines 9-16 from Algorithm \ref{alg:adaptive_training_algorithm}}
                }
            }
        }
        Update $\textbf{f}_k^n$ \hspace{10em}\tcc{via Eq.~\ref{eq:interface_info_commu}} 
    }
    \Output{$\theta_k^N$ and $\textbf{q}_k^N$}
    \caption{\textcolor{black}{Domain Decomposition Training Procedure}} \label{alg:ddm}
\end{algorithm}

\section{Numerical Experiments}
We apply our domain decomposition method to solve both forward and inverse PDE problems in a parallel fashion using the Message Passing Interface (MPI) standard.
To evaluate the accuracy of our predictions, we consider an \(n\)-dimensional vector of predictions \(\boldsymbol{\hat{u}} \in \mathbb{R}^n\) and the corresponding exact values \(\boldsymbol{u}\). We define the relative Euclidean (or \(\mathit{l^2}\)) norm and the infinity norm (\(\mathit{l^\infty}\)) as follows:
\begin{align}
    \mathcal{E}_r(\boldsymbol{\hat{u}},\boldsymbol{u}) = \frac{\| \boldsymbol{\hat{u}} - \boldsymbol{u}\|_2}{\|\boldsymbol{u}\|_2}, 
    \label{eq:relative_L2_Error} \quad
    \mathcal{E}_{\infty}(\boldsymbol{\hat{u}},\boldsymbol{u}) = \| \boldsymbol{\hat{u}}-\boldsymbol{u}\|_{\infty},
\end{align}
where \(\| \cdot \|_{\infty}\) denotes the maximum norm. 
Additionally, we evaluate the parallel performance of our method through both weak and strong scaling analyses.

\subsection{Forward Problems}
Here, we focus on two types of forward problems: Poisson’s and Helmholtz equations. Among them, solving PDE problems with solutions that exhibit multi-scale and high wavenumber features has been a significant challenge for PINNs due to the mean computation of the global loss terms \cite{wang_multi-scale2021, wang2022respecting}. It is crucial to highlight that Poisson's and Helmholtz equations are considered foundational in the study of domain decomposition methods \cite{smith1998domain, quarteroni1999DDM, dolean2015introduction}. For instance, the classical Schwarz method for Poisson’s equation necessitates overlap between subdomains for convergence, with the rate of convergence being influenced by the overlap size \cite{dolean2015introduction}. In the present method, we pursue a DDM with no overlap between subdomains and aim to tackle both types of equations with a common framework.

\subsubsection{Poisson's equation with a low-wavenumber solution}
\label{sec:experiment_psn_simple}
We consider the following Poisson's equation on the domain $\Omega = \{ (x,y) ~ | ~ -1 \le x \le 1, -1 \le y \le 1 \}$:
\begin{equation}
    \begin{aligned}
        - \Delta u & = s, && \text{in} \quad \Omega, \\
                 u & = g, && \text{on} \quad \partial \Omega,
         \label{eq:poisson_equation}
    \end{aligned}
\end{equation}
An exact solution that satisfies the Poisson's equation~\eqref{eq:poisson_equation} can be assumed as follows:
\begin{equation}
u(x,y) = \sin(4  \pi  x) \cos(2  \pi  y) + \cos (2 \pi x) \sin(2 \pi y), \quad \forall (x,y) \in \Omega.
 \label{eq:psn_sol_low_freq}
\end{equation}
The corresponding source functions $s(x,y)$ and $g(x,y)$ can be calculated exactly by substituting the solution into Eq.~\eqref{eq:poisson_equation}. 

\begin{figure}
    \centering
    \subfloat[]{\includegraphics[scale=0.45]{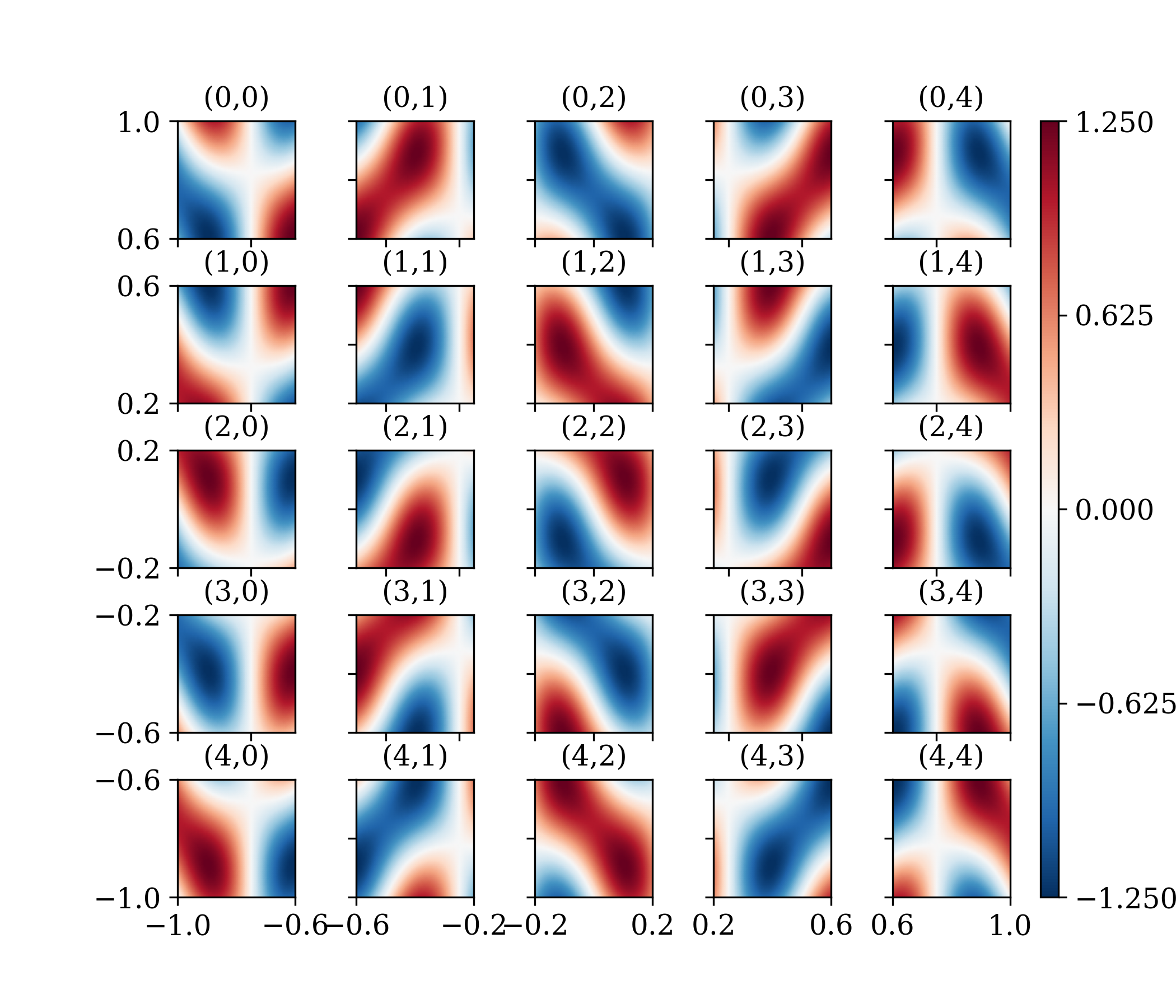}}
    \subfloat[]{\includegraphics[scale=0.45]{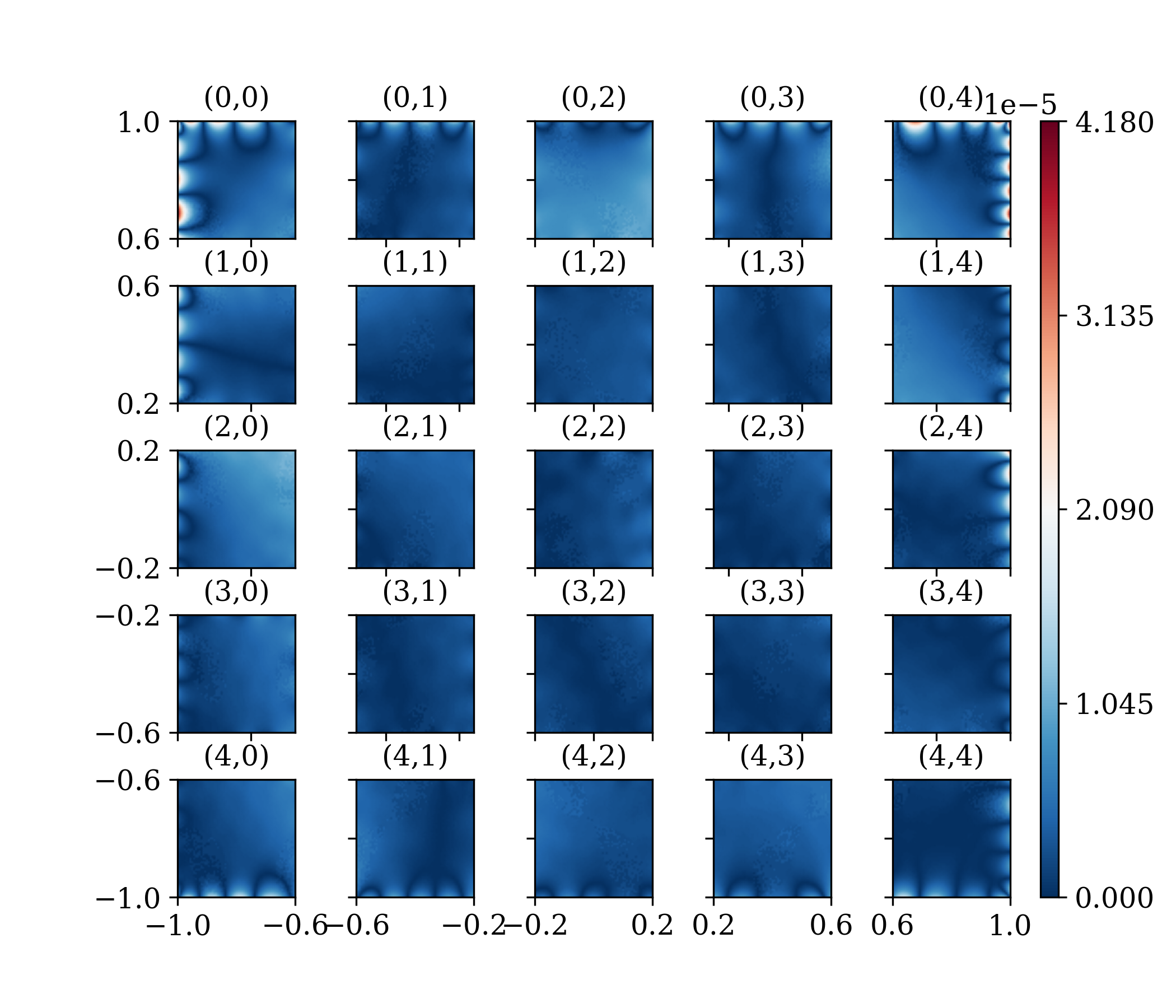}}
    \caption{Poisson's equation with a low-wavenumber solution: (a) predicted solution on a $5\times5$ partitioned domain, (b) absolute point-wise error. Each subdomain is labeled as \textit{(row, column)} following matrix notation. The relative $l^2$ norm is $5.43\times10^{-6}$.}
    \label{fig:poisson_simple_solution}
\end{figure}

We employ a $5\times5$ Cartesian decomposition and utilize a shallow feed-forward neural network consisting of three hidden layers, with each layer containing 20 neurons and tangent hyperbolic activation function in each subdomain. 
The local neural network models are trained for a minimum of 50 inner epochs, capped at 100, after which we justify whether to terminate the current subproblem optimization. Subsequently, the interface information is exchanged between neighboring subdomains.
The neural network parameters are optimized by the L-BFGS optimizer with \textit{strong Wolfe} line search function, while the Adam optimizer is used for the Robin parameters.

In the global domain, we utilize $160 \times 160$ randomly distributed collocation points, including residual, boundary, and interface points. The exact number of points for each subdomain is determined by rounding off calculations based on the specific domain decomposition strategy employed. 
In this setting, each subdomain consists of 1024 residual points and 32 points for each boundary and interface.

Figure~\ref{fig:poisson_simple_solution}(a,b) illustrates the distributions of the predicted solution and the absolute point-wise error of one trial with a $5\times5$ partitioning, respectively. 
To identify the position of subdomains within the global domain, we label each subdomain by its row and column number using a matrix notation.
From Fig.~\ref{fig:poisson_simple_solution}(b), it is observed that the maximum error reaches \(4.180\times10^{-5}\). Subdomains that are devoid of the boundary data, exhibit the same low errors as the subdomains processing boundaries, indicating effective information exchange on the interfaces.

\begin{figure}[!h]
\centering
    \subfloat[]{\includegraphics[scale=0.45]{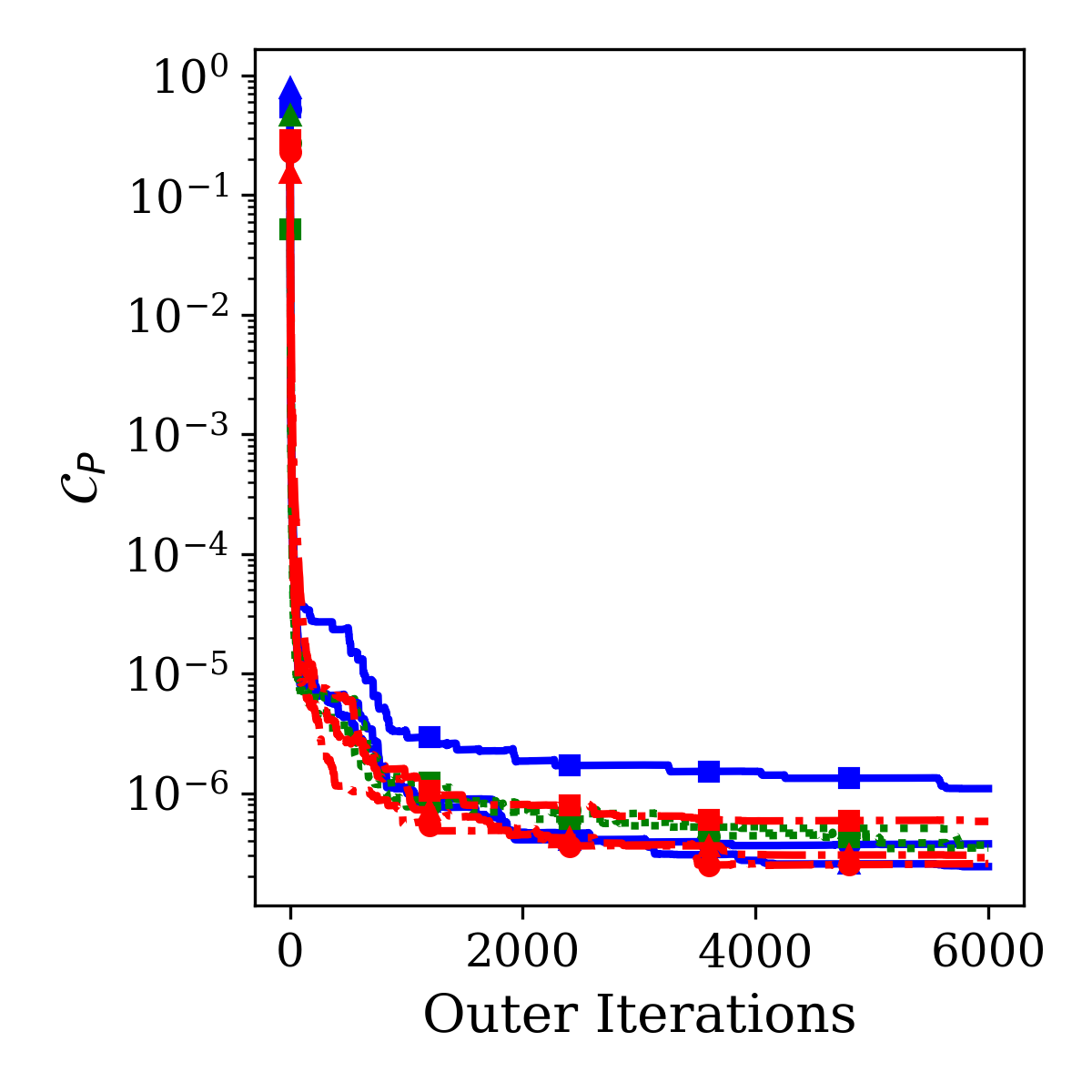}}\quad
    \subfloat[]{\includegraphics[scale=0.45]{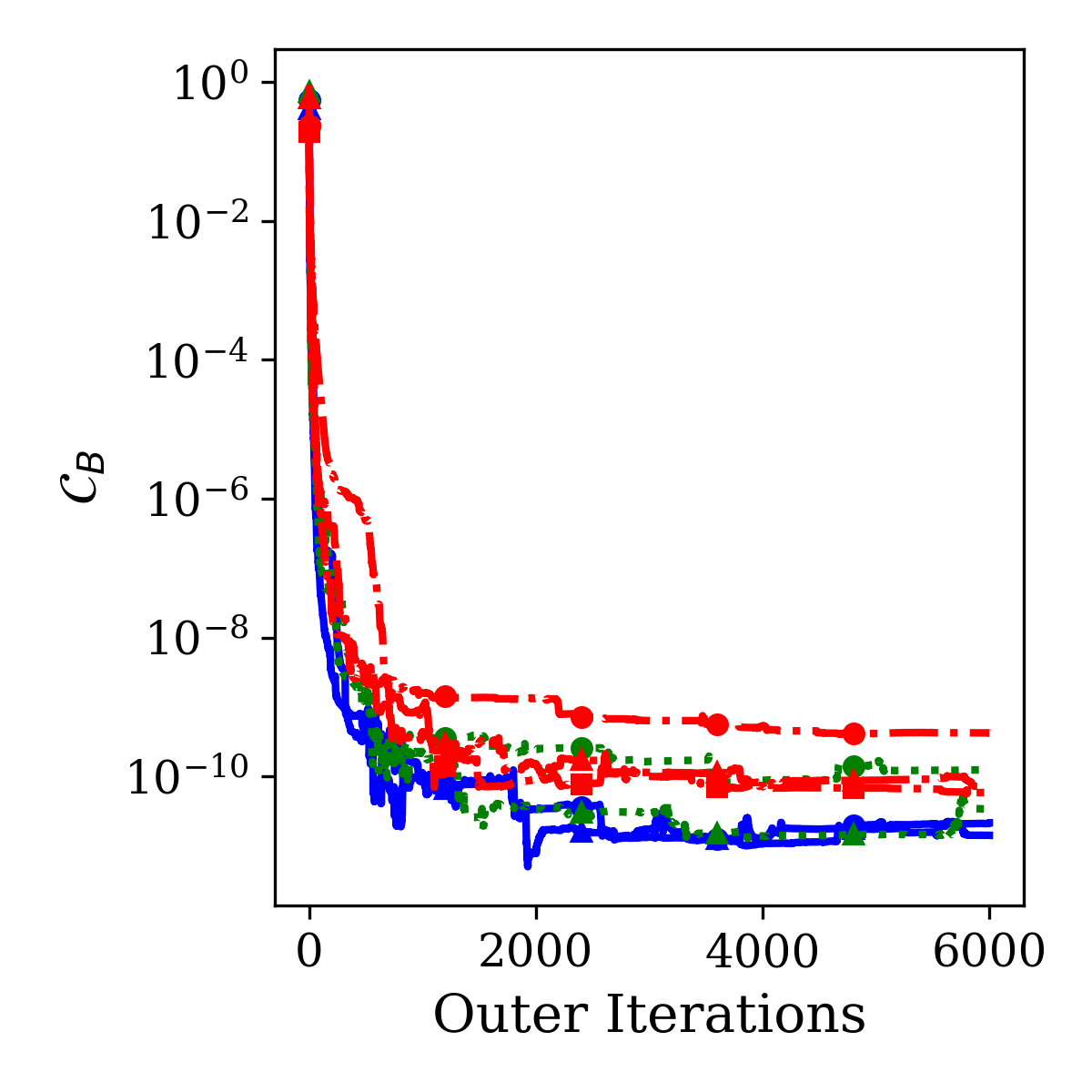}}\quad
    \subfloat[]{\includegraphics[scale=0.45]{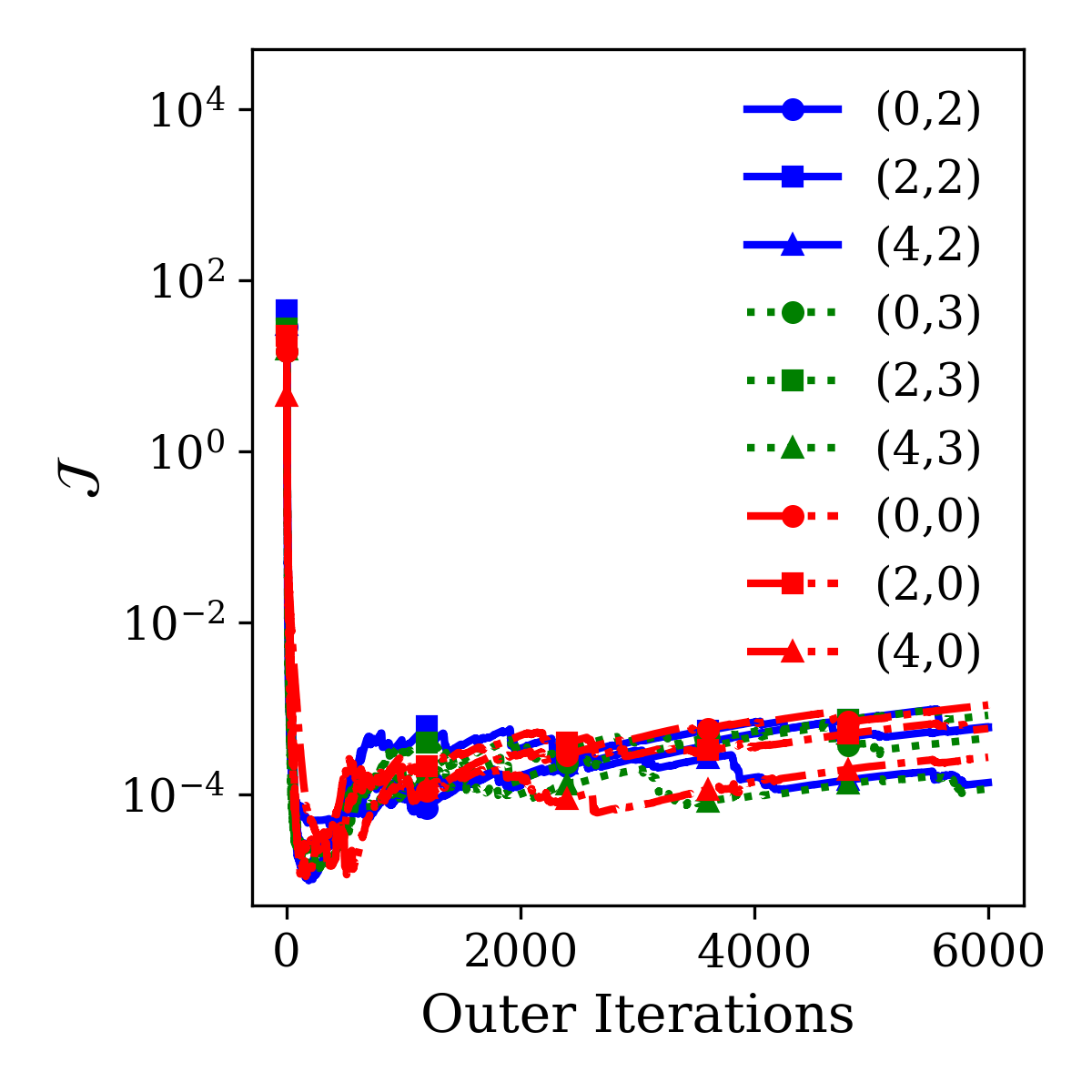}}\quad
    \subfloat[]{\includegraphics[scale=0.45]{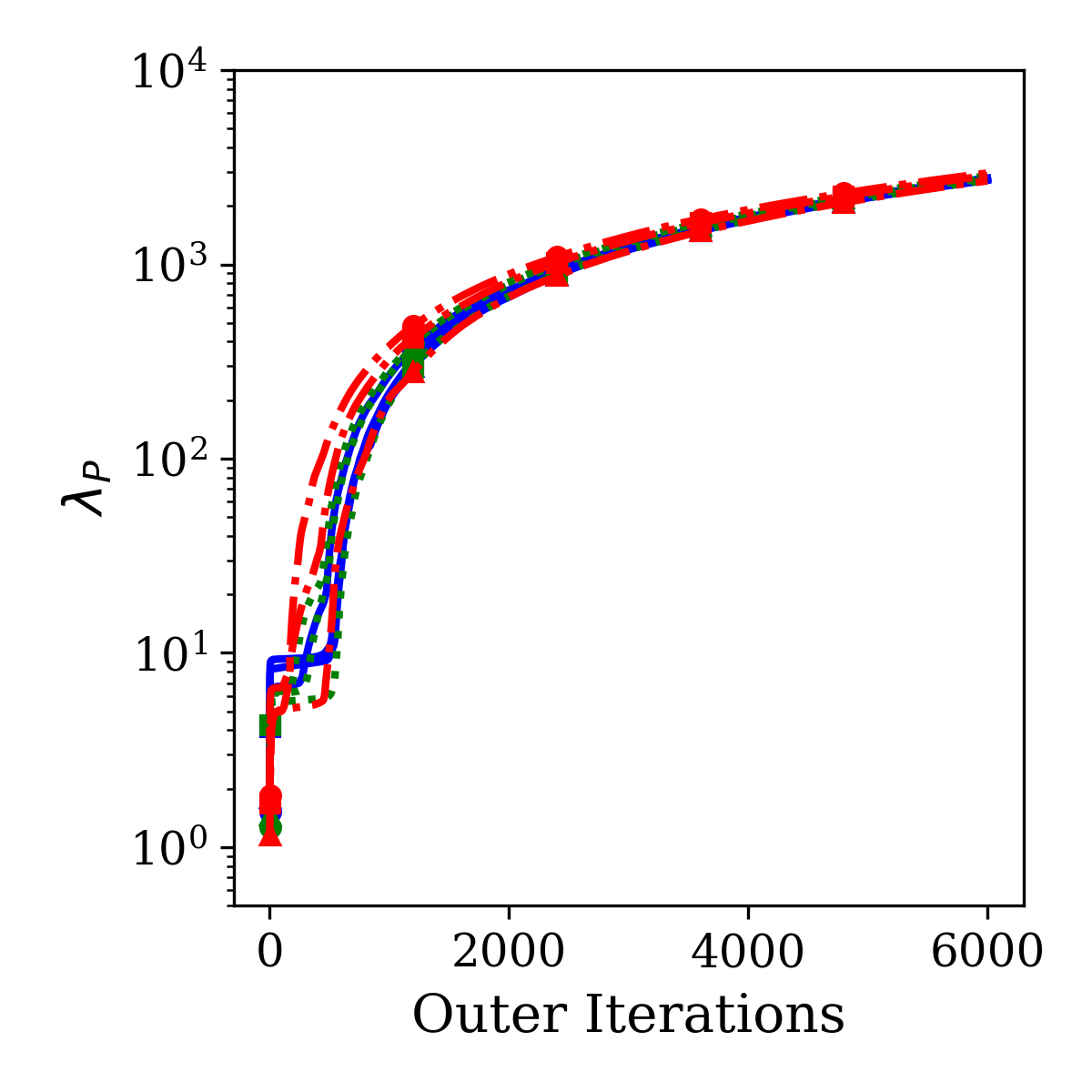}}\quad
    \subfloat[]{\includegraphics[scale=0.45]{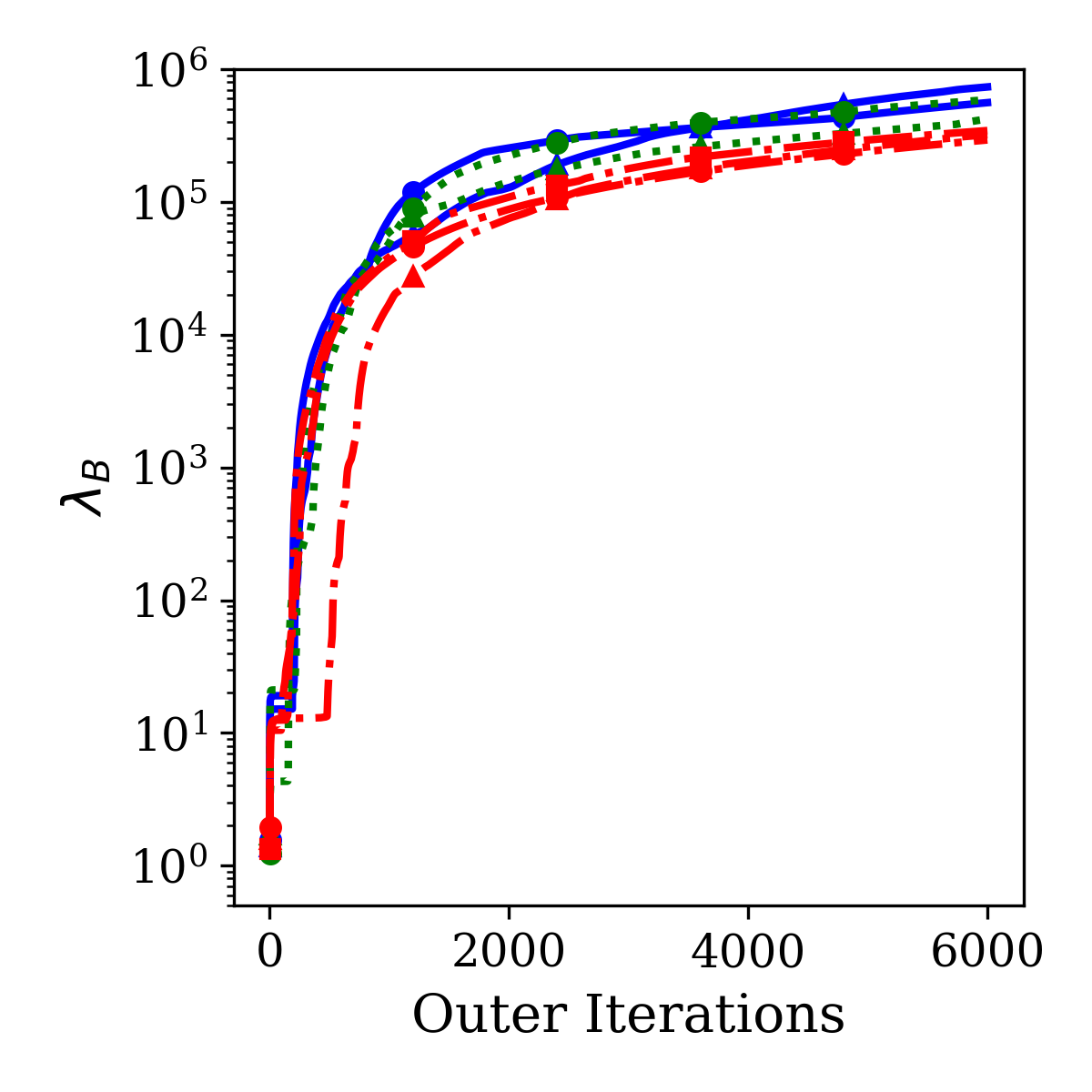}}\quad
    \subfloat[]{\includegraphics[scale=0.45]{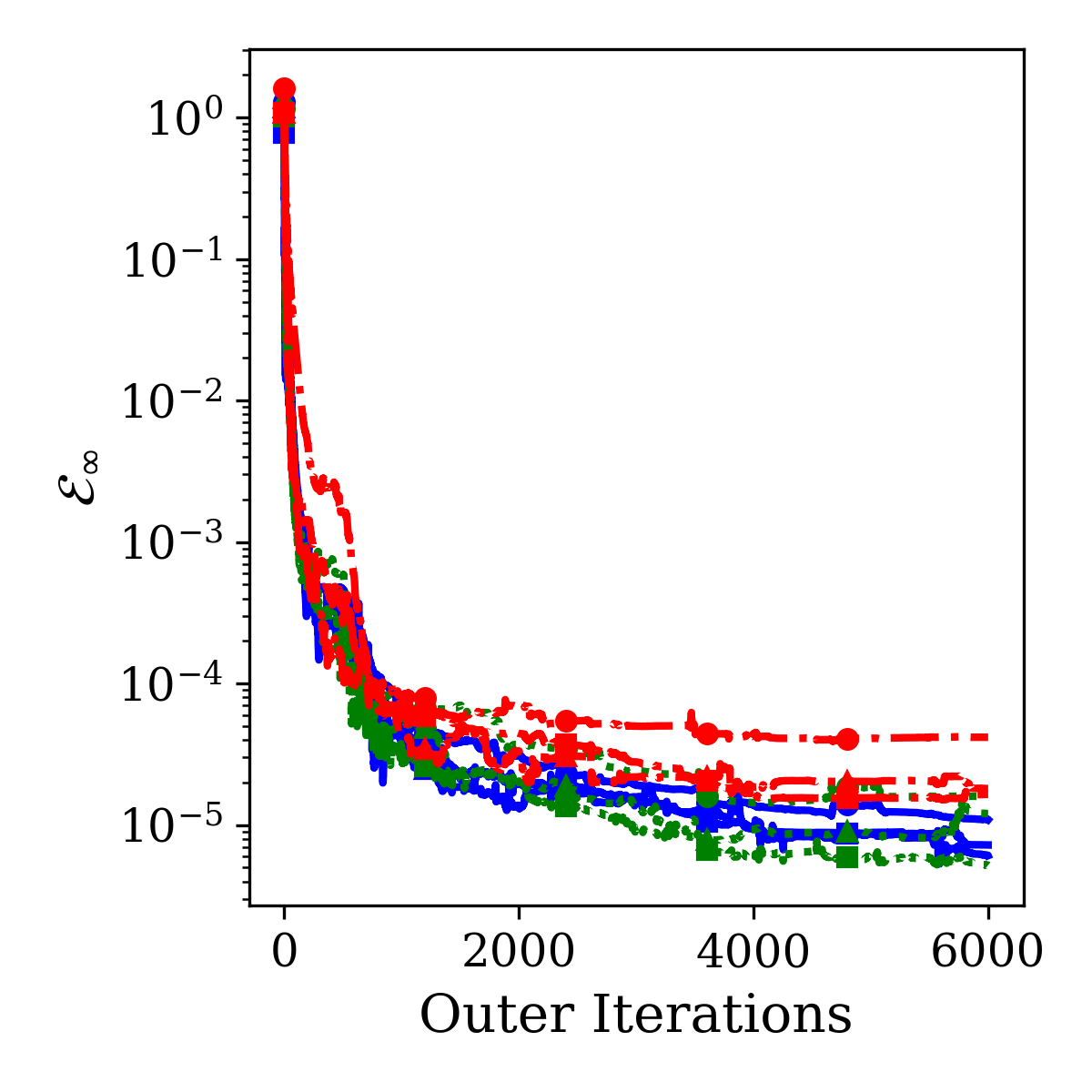}}\quad
    \subfloat[]{\includegraphics[scale=0.45]{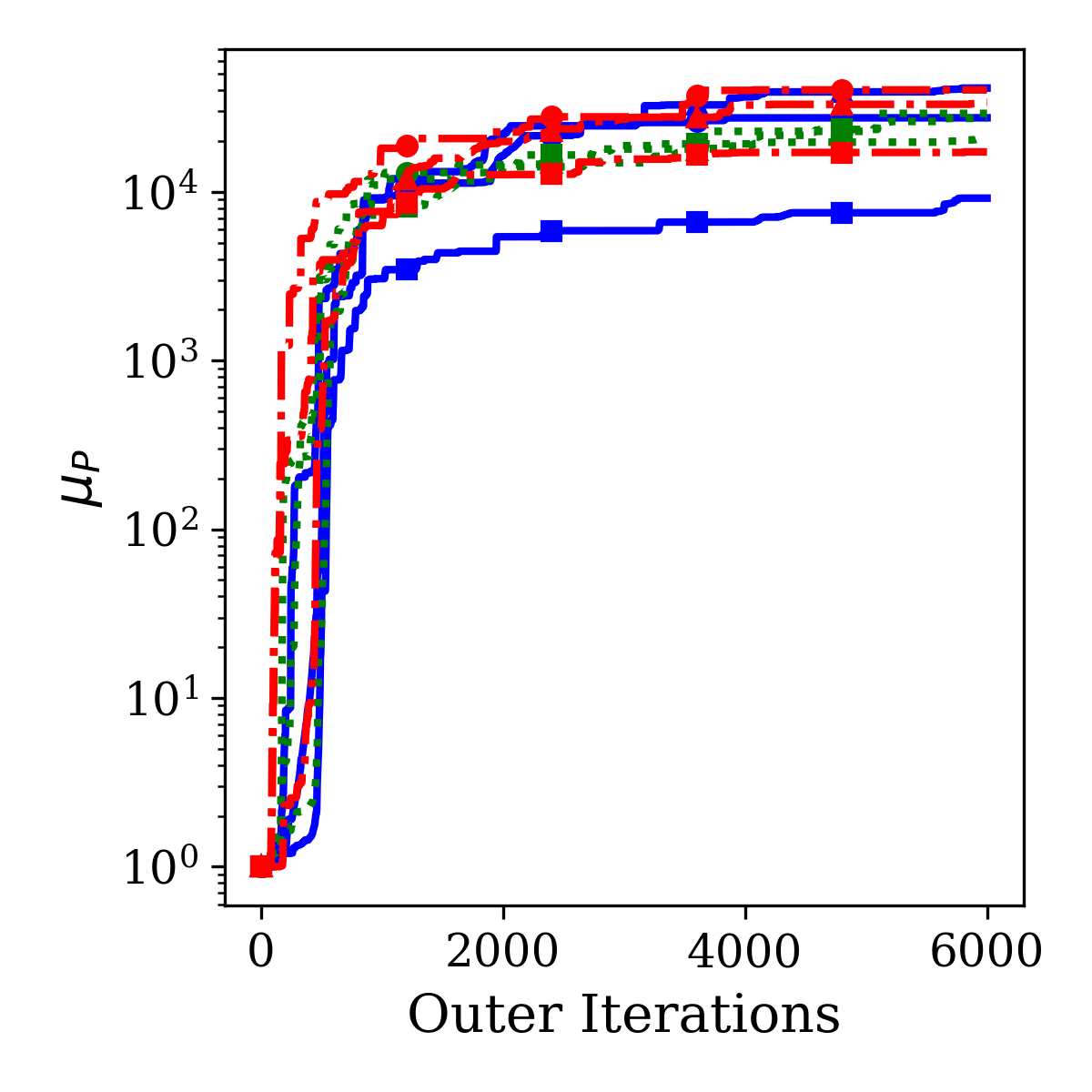}}\quad
    \subfloat[]{\includegraphics[scale=0.45]{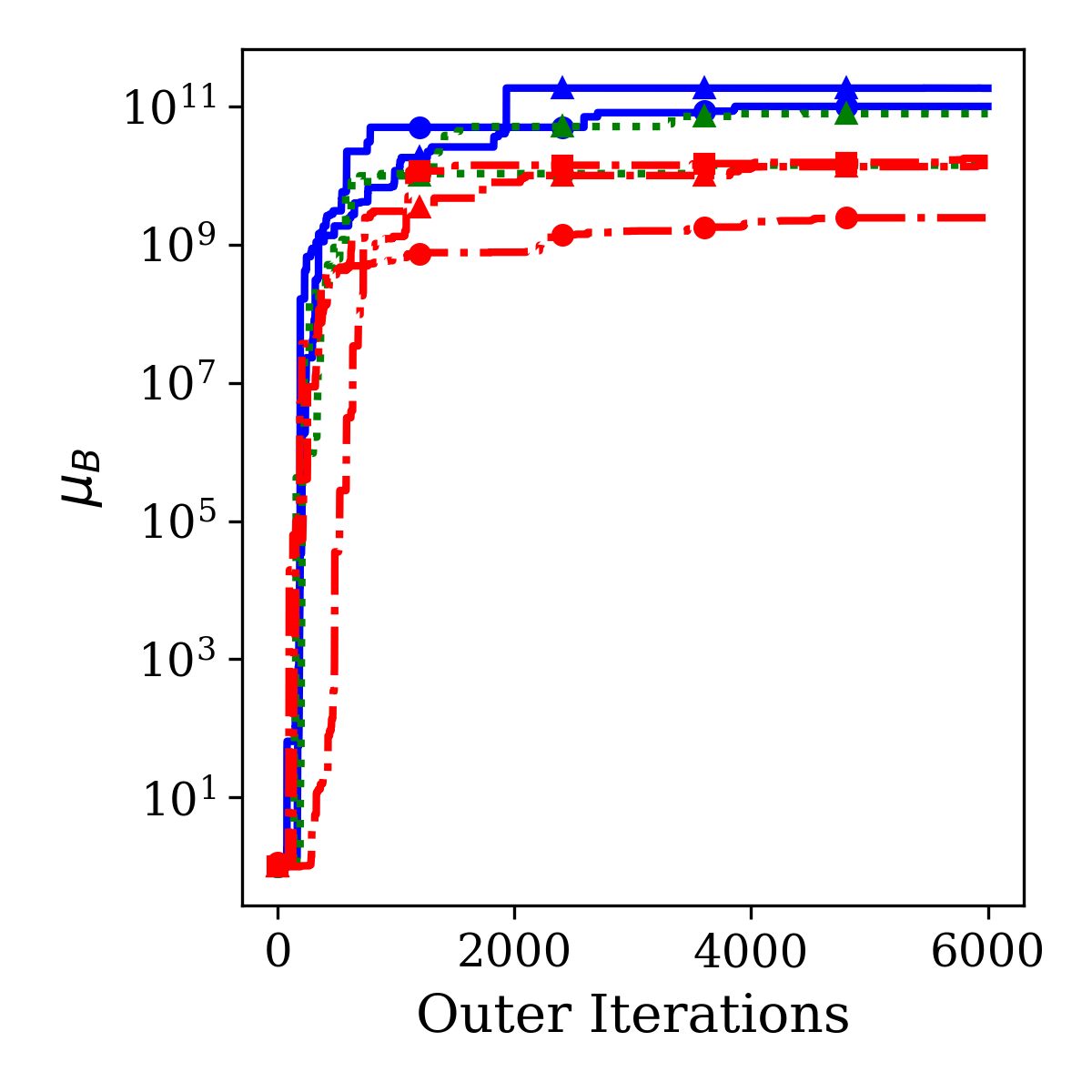}}\quad
    \subfloat[]{\includegraphics[scale=0.45]{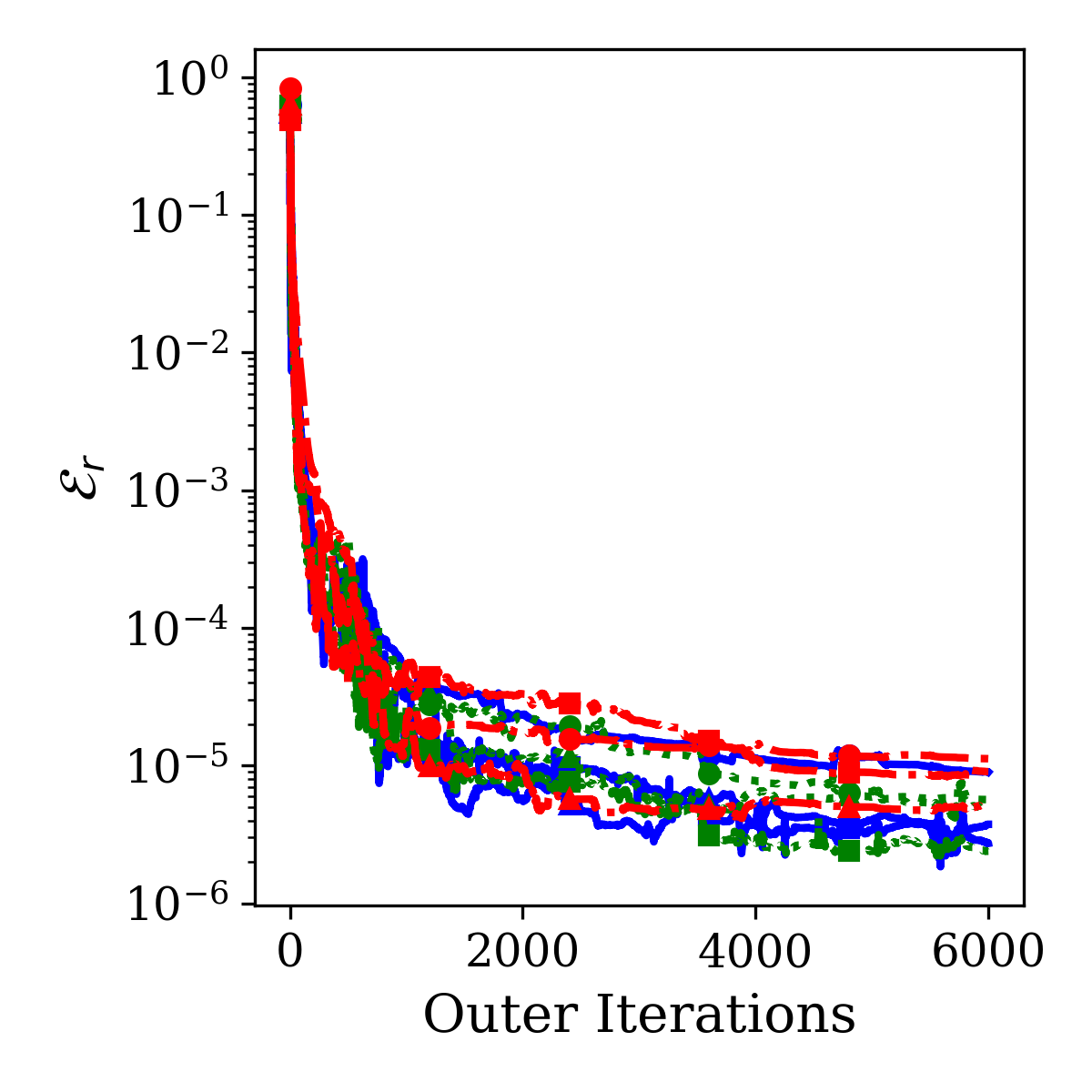}}\quad
    \caption{Poisson's equation with a low-wavenumber solution: Evolution of (a) PDE constraints, (b) boundary condition constraints, (c) objective functions, (d) Lagrange multipliers for PDE constraints, (e) Lagrange multipliers for boundary condition constraints, (f) infinity norms (g) penalty parameters for PDE constraints, (h) penalty parameters for boundary condition constraints, (i) relative $\mathit{l^2}$ norms for nine subdomains randomly selected out of the total 25 subdomains resulting from a $5\times 5$ decomposition. The labels indicate the positions of these subdomains in the global domain as (row, column).
}
\label{fig:poisson_simple_update}
\end{figure}

For ease of presentation, we randomly select nine subdomains out of the 25 subdomains and show the evolution of physics (PDE), boundary condition (BC) constraints, and objective functions as well as the corresponding Lagrange multipliers, penalty parameters, the $\mathit{l^\infty}$ and relative $\mathit{l^2}$ norms across outer iterations in Figure~\ref{fig:poisson_simple_update}. \textcolor{black}{The trends shown in Figs.~\ref{fig:poisson_simple_update}(a-b) indicate that the constraints for PDE and BC are satisfied after approximately 1000 outer iterations, achieving magnitudes on the order of $10^{-6}$, $10^{-10}$, respectively. At the same time, the objective functions drop to a satisfactory level on the order of $10^{-4}$, as shown in Fig.~\ref{fig:poisson_simple_update}(c)}
This behavior reflects the higher prioritization of boundary condition constraints over PDE constraints, with the objective function, represented by the interface losses.
We observe that minimization of the loss terms for subdomains lacking $\mathcal{C}_B$, such as subdomain $(2, 2)$, exhibited relatively larger values of convergence.
This phenomenon in the interior subdomains is attributed to their reliance solely on interface information from neighboring subdomains, which evolves after each communication step, particularly during the early stages of training.
Despite this, the effective interface condition ensures consistency and synchronization of predictions across all subdomains. Combined with the relatively rapid satisfaction of the physics constraint, this reinforces the overall integrity and effectiveness of the distributed solution.

Figs.~\ref{fig:poisson_simple_update}(d) and \ref{fig:poisson_simple_update}(e) show that the Lagrange multipliers $\bm{\lambda}$ for the PDE and boundary condition constraints converge to approximately $10^{3}$ and $10^{5}$, respectively. This difference is attributed to the varying penalty scaling factors $\bm{\eta}$ in the penalty parameters $\bm{\mu}$, with a larger $\eta_B$ for the BC, allowing the models to prioritize boundary conditions over physics, as seen in Figs.~\ref{fig:poisson_simple_update}(g) and \ref{fig:poisson_simple_update}(h). Although $\mu_B$ grow significantly, reaching values around $10^{10}$, the adaptive, gradually increasing strategy is crucial for preventing ill-conditioning, a common issue in penalty methods. This approach ensures the numerical stability and robustness of our optimization process.
Eventually, as shown in Figs.~\ref{fig:poisson_simple_update}(f) and \ref{fig:poisson_simple_update}(i), the infinity norms and relative \(\mathit{l^2}\) norms stabilize around $10^{-5}$ after approximately 2000 outer iterations.

\begin{figure}[!h]
\centering
    \subfloat[]{\includegraphics[scale=0.45]{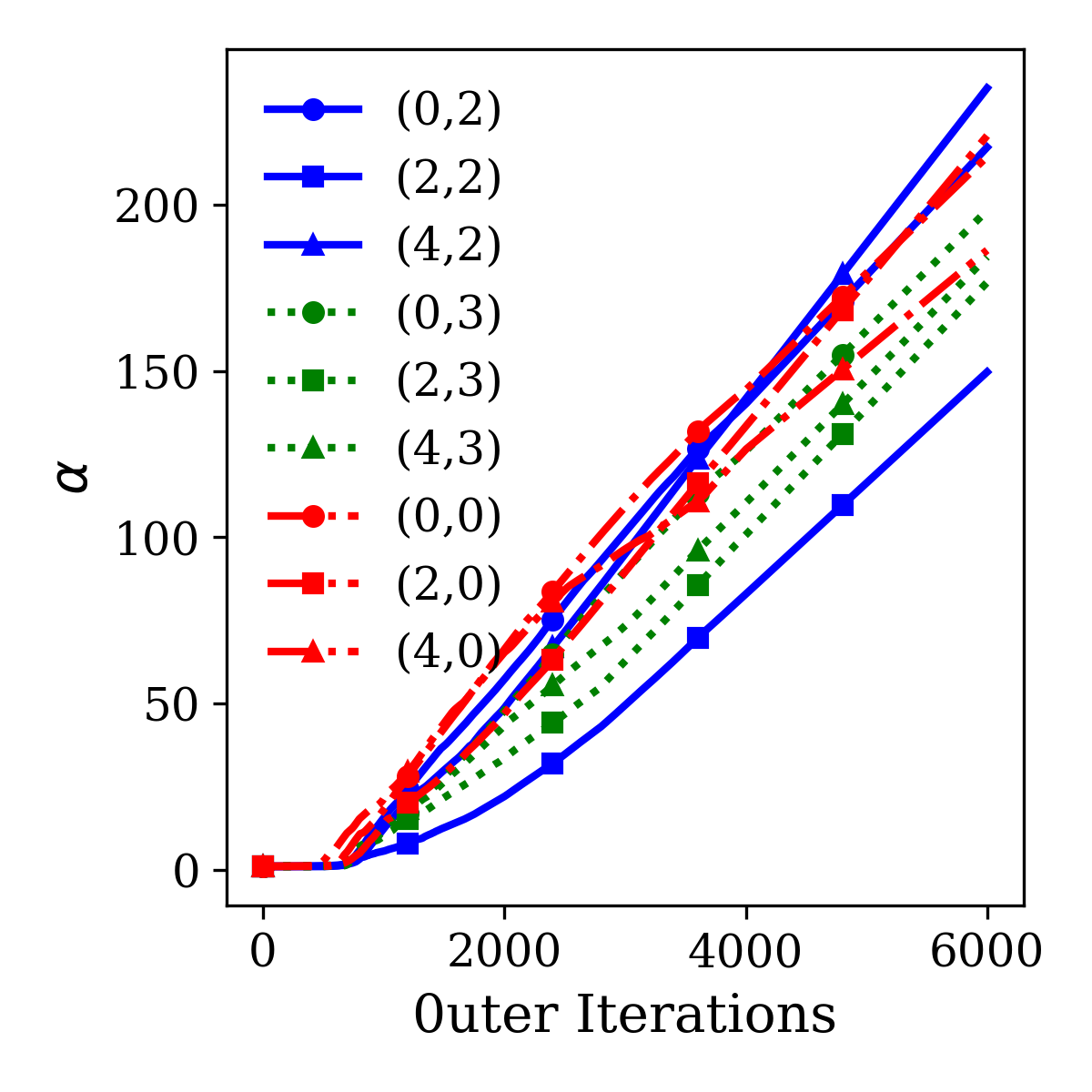}}\quad
    \subfloat[]{\includegraphics[scale=0.45]{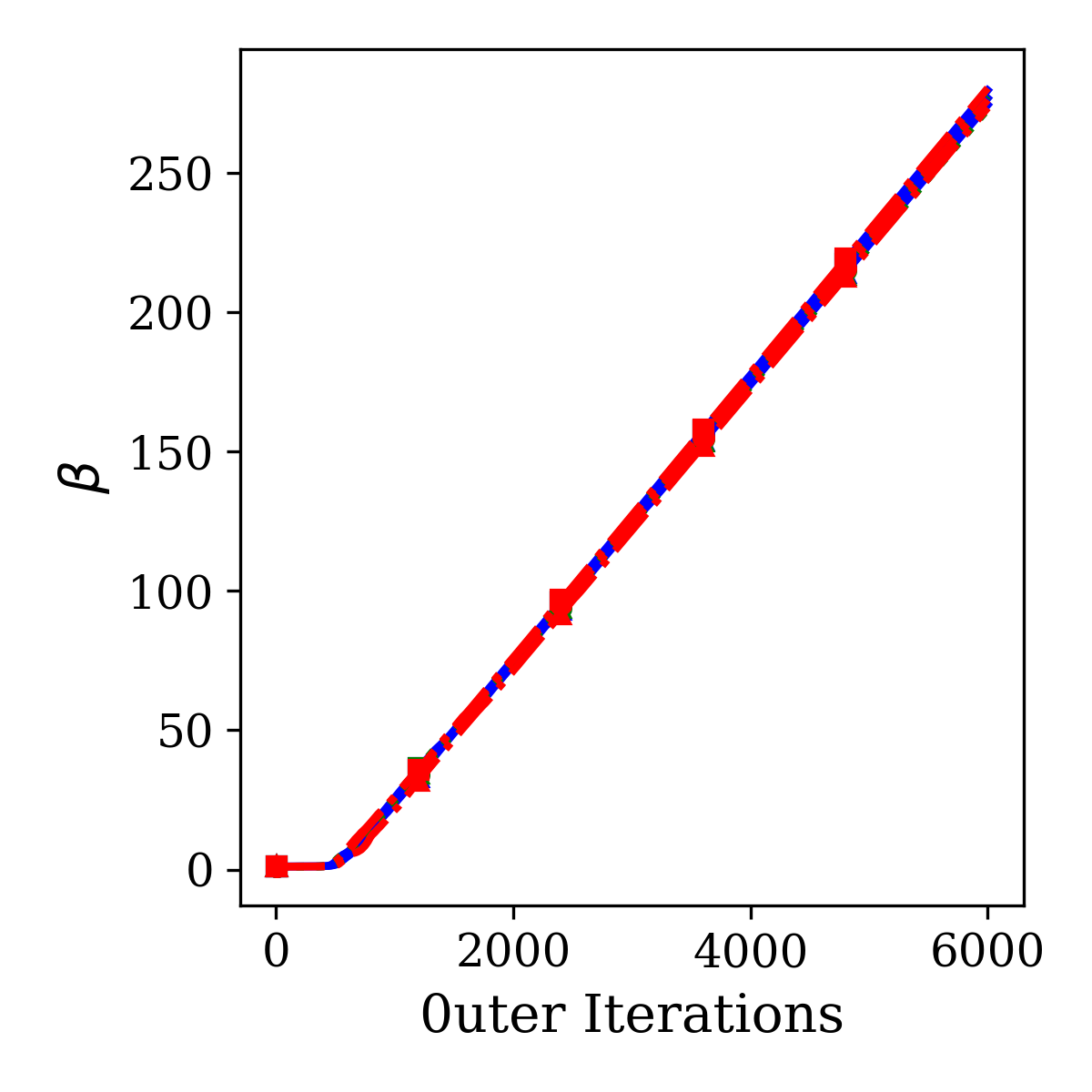}}\quad
    \subfloat[]{\includegraphics[scale=0.45]{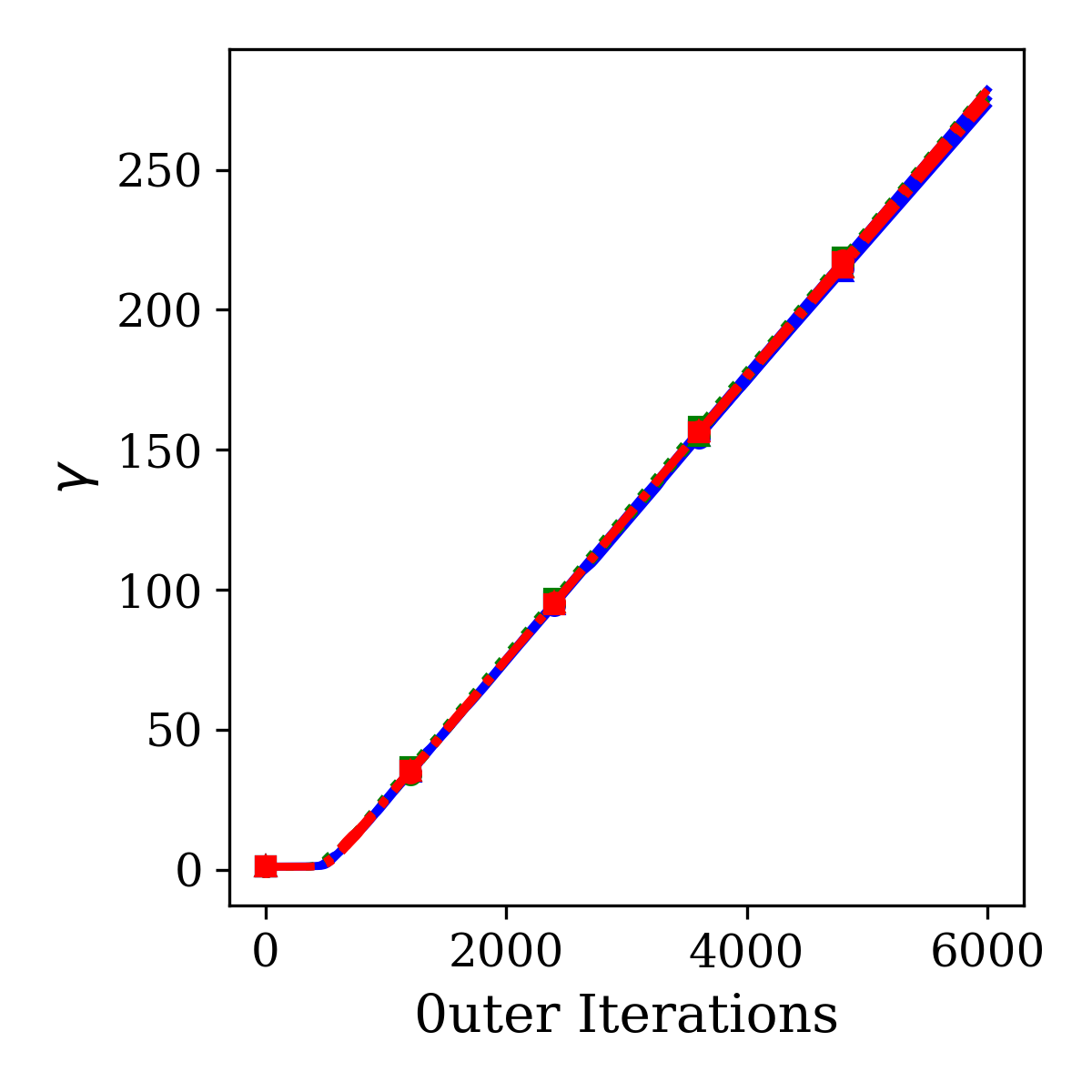}}\quad
    \subfloat[]{\includegraphics[scale=0.45]{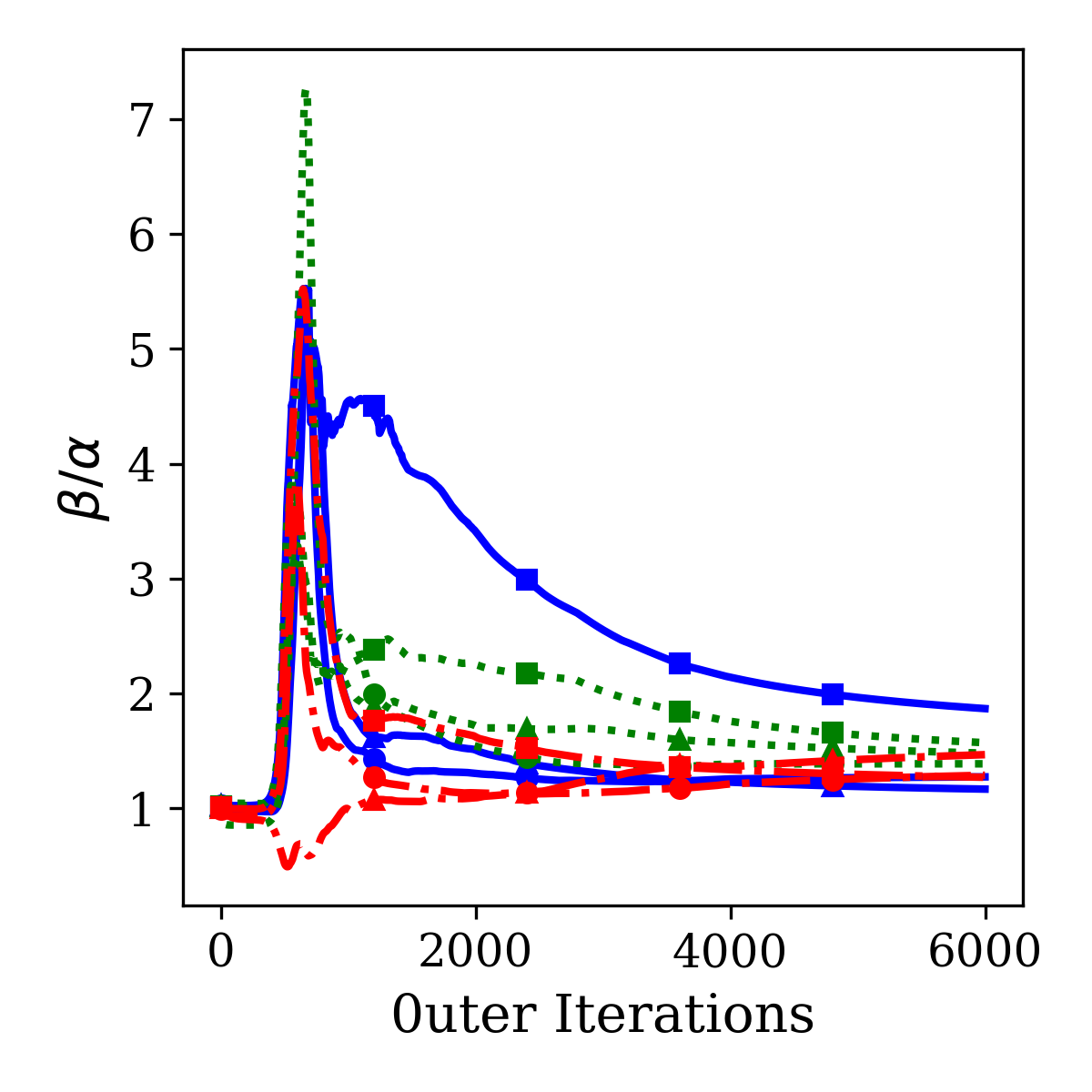}}\quad
    \subfloat[]{\includegraphics[scale=0.45]{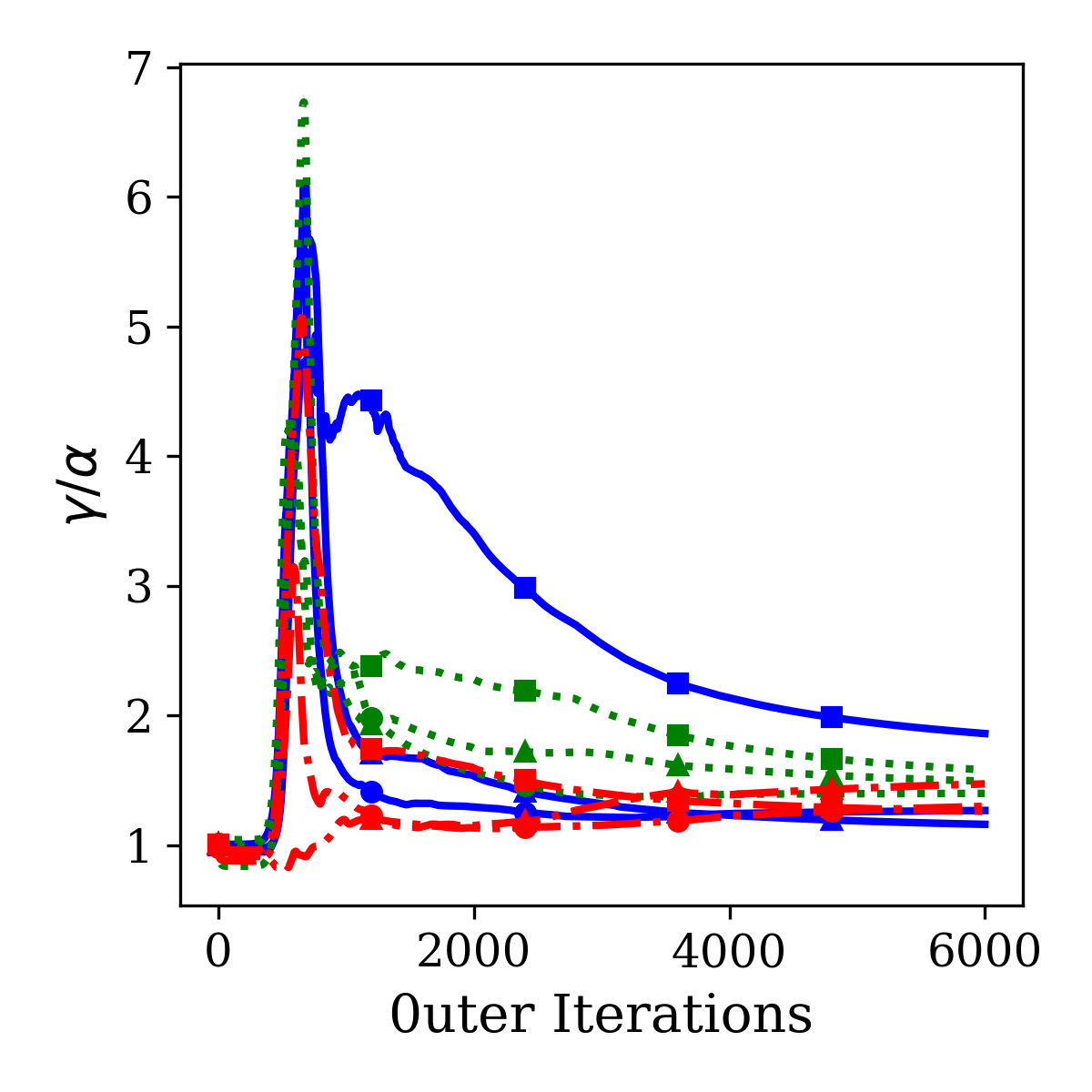}}\quad
    \caption{Poisson's equation with a low-wavenumber solution: Evolution of interface parameters (a) $\alpha$, (b) $\beta$, (c) $\gamma$, (d) the ratio $\beta / \alpha$, (e) the ratio $\gamma / \alpha$ for nine randomly selected subdomains. The labels indicate the positions of these subdomains in the global domain as (row, column).}
    \label{fig:poisson_simple_update_interface}
\end{figure}

Figure~\ref{fig:poisson_simple_update_interface} offers critical insights into the evolution of the interface parameters $\alpha$, $\beta$ and $\gamma$, as well as their ratios $\dfrac{\beta}{\alpha}$ and $\dfrac{\gamma}{\alpha}$. 
From Figs.~\ref{fig:poisson_simple_update_interface}(a-c), the interface parameters remain relatively constant during the initial rapid convergence period of the objectives.
Subsequently, they begin to increase, indicating a growing influence towards the Dirichlet, Neumann and tangential derivative continuity operators. 
Despite this increase, the corresponding objective function does not rise in Fig.~\ref{fig:multiscale_poisson_update}(c), suggesting that the MSE metrics of the operators are decreasing further. 
This is a clear indication of the effectiveness of introducing Robin parameters in accelerating the convergence of the interface loss.
The ratios in Fig.~\ref{fig:poisson_simple_update_interface}(d-e) generally stabilize at values greater than 1, eventually converging between 1 and 2 after approximately 4000 outer iterations. 
It is noteworthy, however, that the convergence of interface ratios occurs much later in the optimization process when compared to the loss terms in Figs.~\ref{fig:multiscale_poisson_update}(a-c) and the relative $\mathit{l^2}$ norm in Fig.~\ref{fig:multiscale_poisson_update}(i). 
This delay suggests that achieving a near-optimal solution does not necessarily require the interface ratios to fully converge to their near-optimal values, allowing for a reduction in computational cost in practice.

\begin{table}[!h]
\centering
\begin{tabular}{ccc}
\hline
Decomposition & Relative $\mathit{l^2}$ norm    & $\mathit{l^\infty}$ norm          \\ \hline
$2\times1$           & $3.74e-2 \pm 5.16e-2$ & $2.16e-1 \pm 2.67e-1$   \\
$2\times2$           & $2.25e-3 \pm 2.35e-3$ & $1.50e-2 \pm 1.59e-2$ \\
$3\times3$           & $3.78e-5 \pm 1.16e-5$ & $2.56e-4 \pm 5.28e-5$ \\
$4\times4$           & $2.49e-5 \pm 2.72e-6$ & $2.15e-4 \pm 6.90e-5$ \\
$5\times5$           & $2.29e-5 \pm 6.53e-6$ & $1.48e-4 \pm 4.32e-5$ \\ \hline
\end{tabular}
\caption{Generalization ability of the proposed DDM demonstrated through the Poisson's equation with a low-wavenumber solution: The figure displays the relative $\mathit{l^2}$ and $\mathit{l^\infty}$ norms of the global domain as the number of subdomains in the domain decomposition increases. Note that the number of neurons per layer has been reduced to 10, compared to 20 in the results shown in Figure \ref{fig:poisson_simple_solution}.}
\label{tab:psn_simple_l2_linf}
\end{table}

Here, we assess the generalization ability of our proposed DDM using the same low-wavenumber solution of Poisson's equation example. We strategically reduce the neural network model's complexity to 10 neurons per layer from the original 20 neurons per layer while increasing the number of subdomains in the DDM.
This reduction aims to limit the network's ability to represent intricate functions.
We keep the total number of collocation points assigned to the global domain constant, while varying the number of subdomains. 
Table~\ref{tab:psn_simple_l2_linf} presents the mean and standard deviation of the relative $\mathit{l^2}$ norms and $\mathit{l^\infty}$ norms of the global domain, over 5 different trials. From the data, it becomes evident that a $2\times1$ domain decomposition produces the highest error, indicating a weaker representation by the neural network. 
In contrast, as the number of subdomains are increased, the errors diminish markedly. 
This improvement empirically illustrates the generalization ability of our proposed DDM and the neural network's capacity to resolve the complexities of the solution when more focused, smaller subdomains are employed. 
Specifically, the transition from a $2\times1$ to a $5\times5$ decomposition reveals a marked improvement in both of the error metrics, underscoring the benefits of finely partitioned domain decomposition in reducing the complexity of neural network models. Our tests show that our DDM offers advantages that go beyond simply addressing and accelerating large computational problems; it also enhances prediction accuracy. However, the trend of improved accuracy with an increasing number of subdomains is not sustainable indefinitely. As the subdomains become smaller, errors due to potential overfitting resulting from the less collocation points are likely to dominate, ultimately limiting performance improvements.

\subsubsection{\textcolor{black}{Poisson's equation with a multi-scale solution}}
\label{sec:experiment_psn_multi}
PINNs perform poorly when applied to PDE problems with multi-scale solutions because global loss terms are computed on a mean basis \cite{wang_multi-scale2021, wang2022respecting}. This averaging process can obscure critical details in regions characterized by complex variability, thereby hindering the effective transmission of essential information from initial and boundary conditions into the target domain. The limitation can result in an inaccurate representation of the physical phenomena \cite{anagnostopoulos_residual-based_2024}. To demonstrate the effectiveness of our DDM and the overall PECANN framework in problems with multi-scale solutions, we amplify one of the wavenumbers in the exact solution from Eq.~\ref{eq:psn_sol_low_freq}, resulting in the following multi-scale solution:
\begin{equation}
u(x,y) = \sin(40 \pi x) \cos(2 \pi y) + \cos(2 \pi x) \sin(2 \pi y), \quad \forall (x,y) \in \Omega.
 \label{eq:poisson_multi_scale}
\end{equation}

\begin{figure}[!h]
    \centering
    \subfloat[]{\includegraphics[scale=0.45]{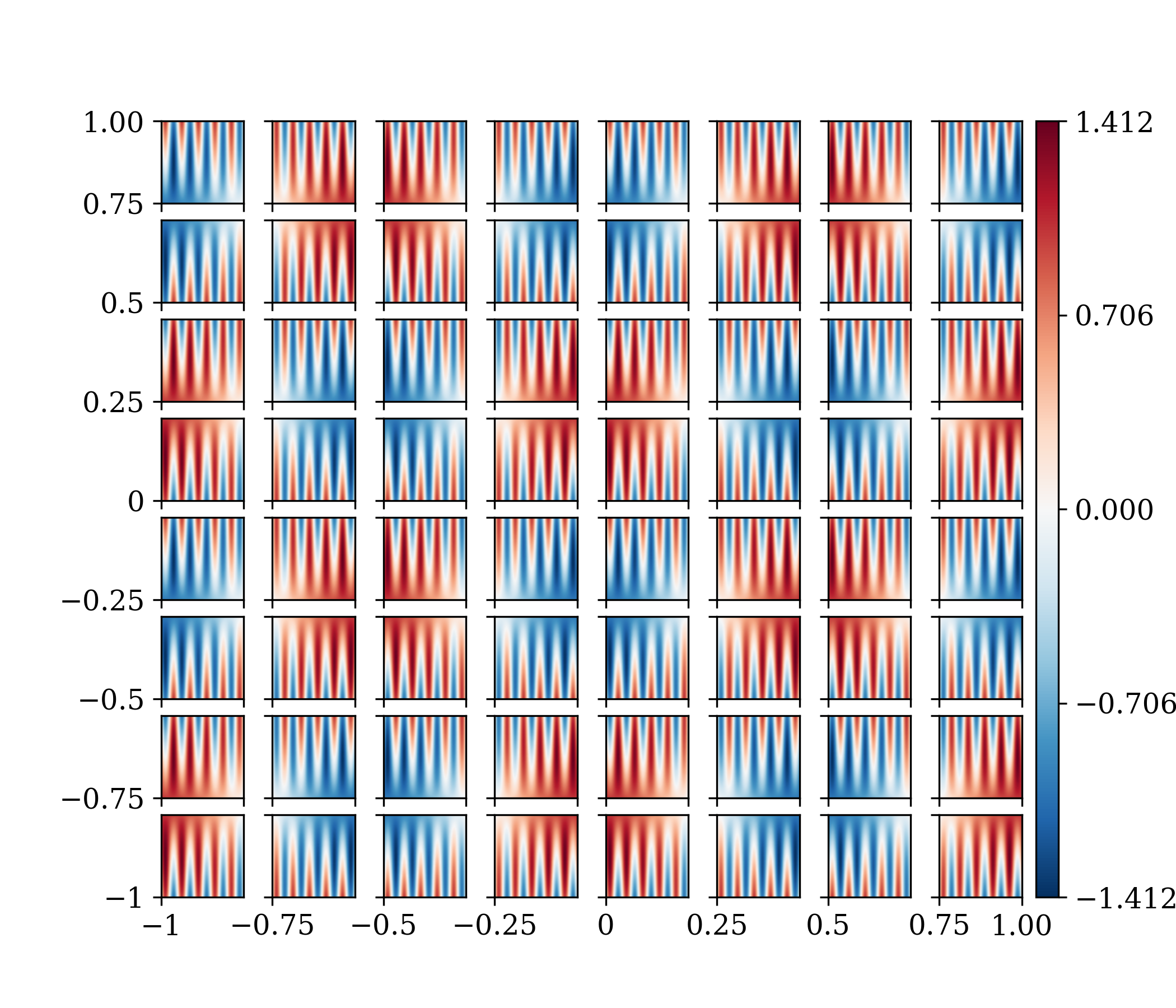}}\quad
    \subfloat[]{\includegraphics[scale=0.45]{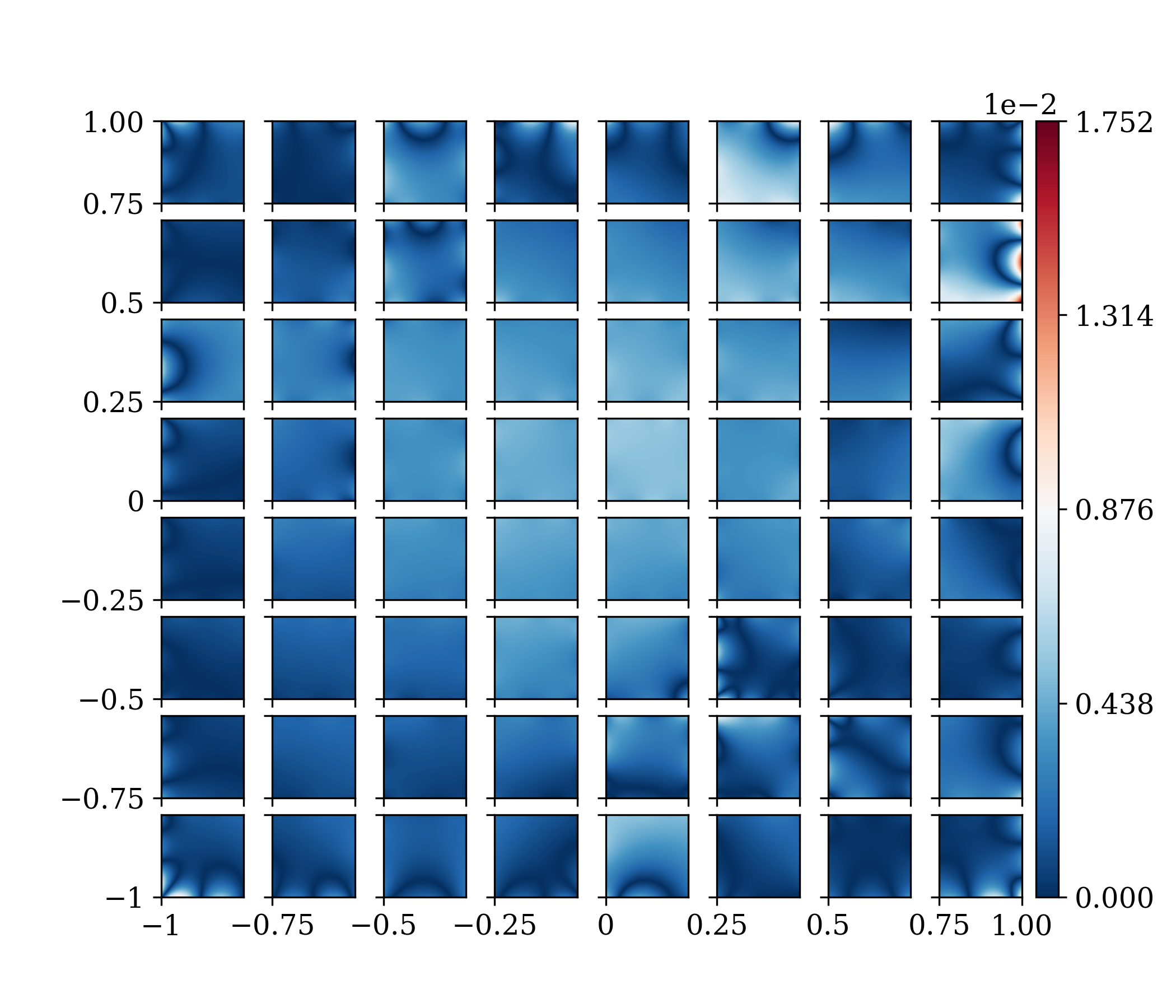}}\quad
    \caption{Poisson's equation with a multi-scale solution: (a) predicted solution on a $8\times8$ partitioned domain, (b) absolute point-wise error. The relative $l^2$ norm is $3.665\times10^{-3}$.}
    \label{fig:multiscale_poisson_solution}
\end{figure}

In this case, we utilize $640 \times 640$  randomly distributed collocation points, in the global domain, with an $8\times8$  decomposition.
All other modeling and optimizer settings remain the same as in Figure~\ref{fig:poisson_simple_solution}.
Figs.~\ref{fig:multiscale_poisson_solution}(a,b) illustrates the distributions of the predicted solution and the absolute point-wise error for the best trial, respectively. 
From Fig.~\ref{fig:multiscale_poisson_solution}(b), the maximum error is on the order of magnitude \(10^{-2}\), confined in the subdomain (1, 7). 
Subdomains lacking boundary data generally exhibit higher errors, which contrasts with the findings from the low-wavenumber case, highlighting the increased complexity of multi-scale problems.

\begin{figure}[t!]
\centering
    \subfloat[]{\includegraphics[scale=0.45]{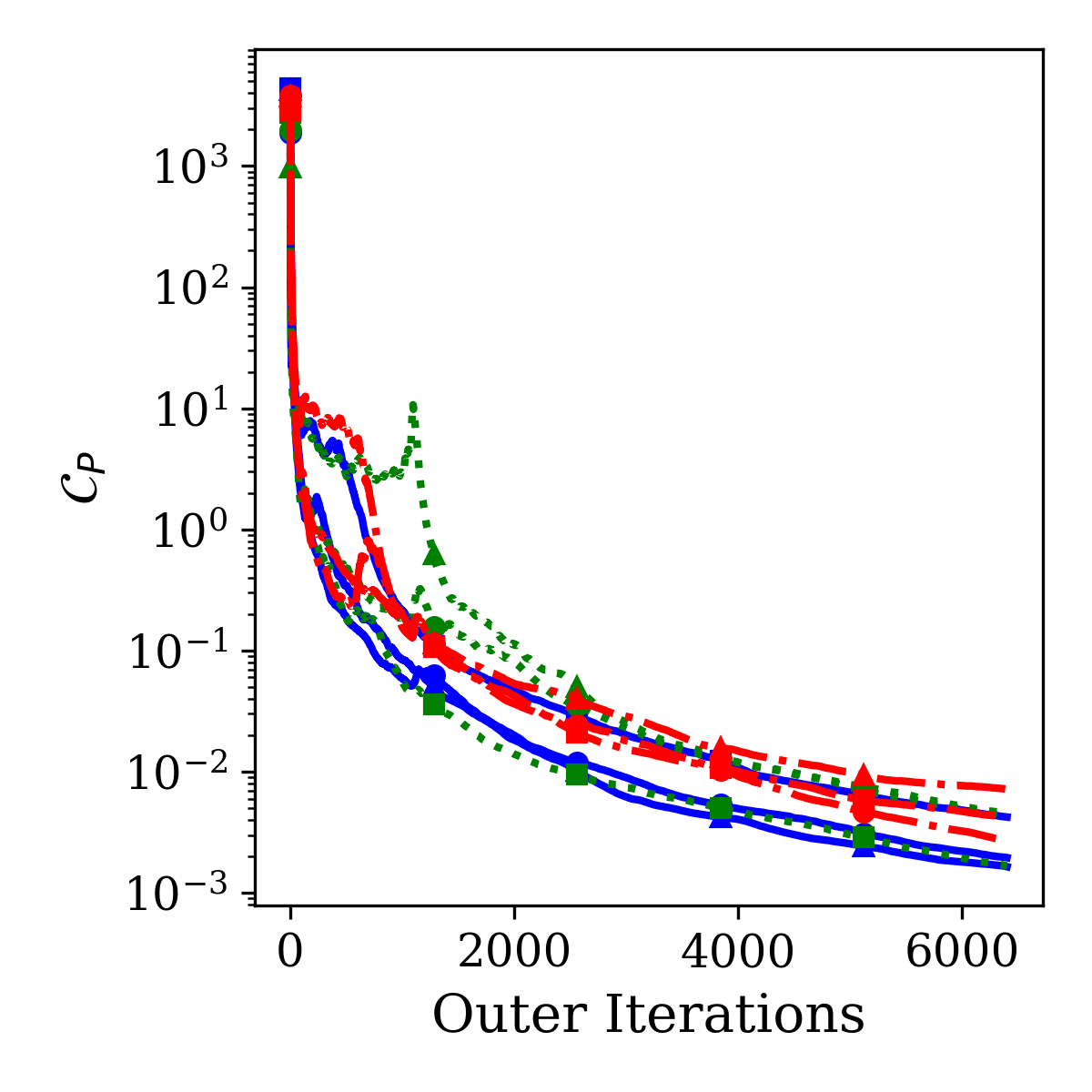}}\quad
    \subfloat[]{\includegraphics[scale=0.45]{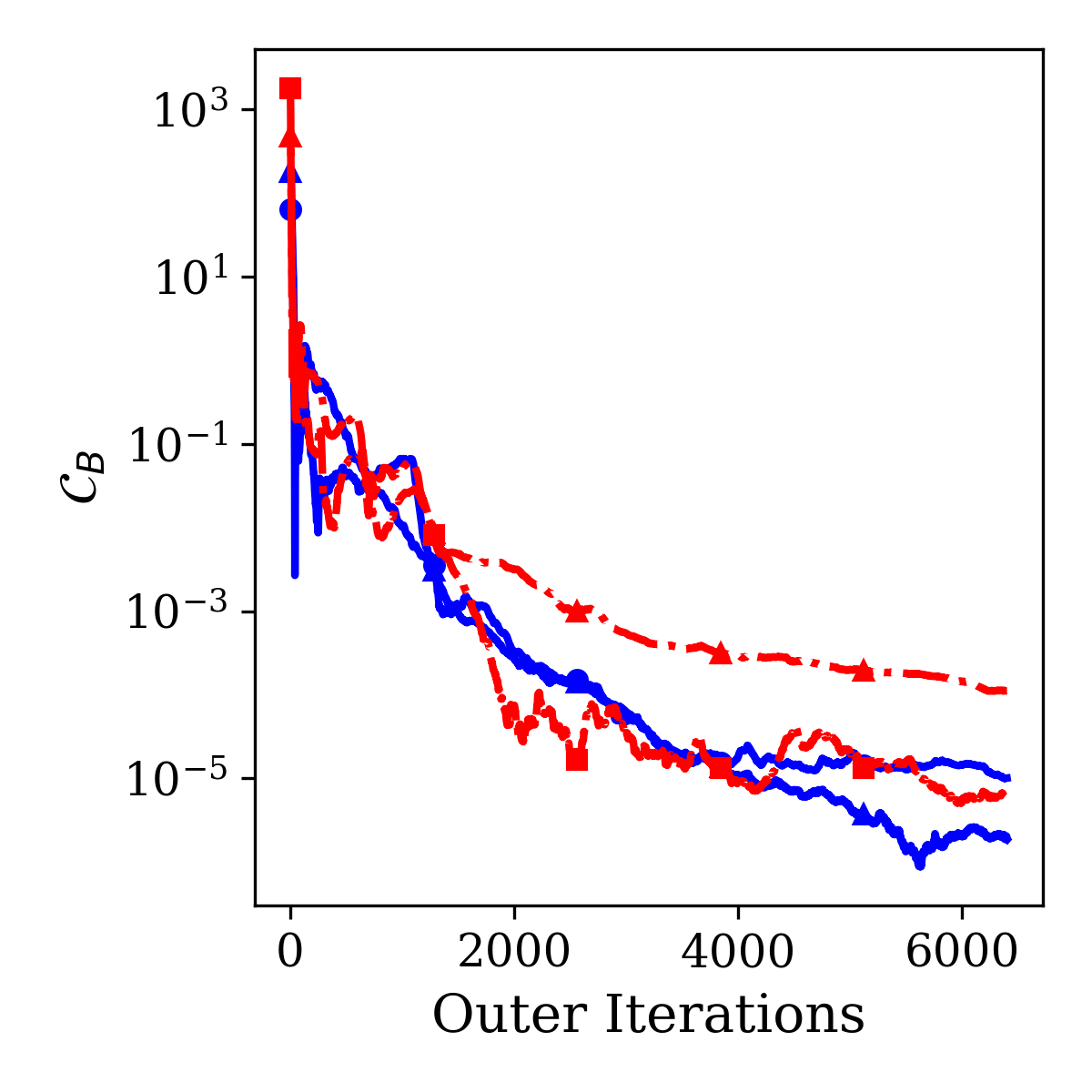}}\quad
    \subfloat[]{\includegraphics[scale=0.45]{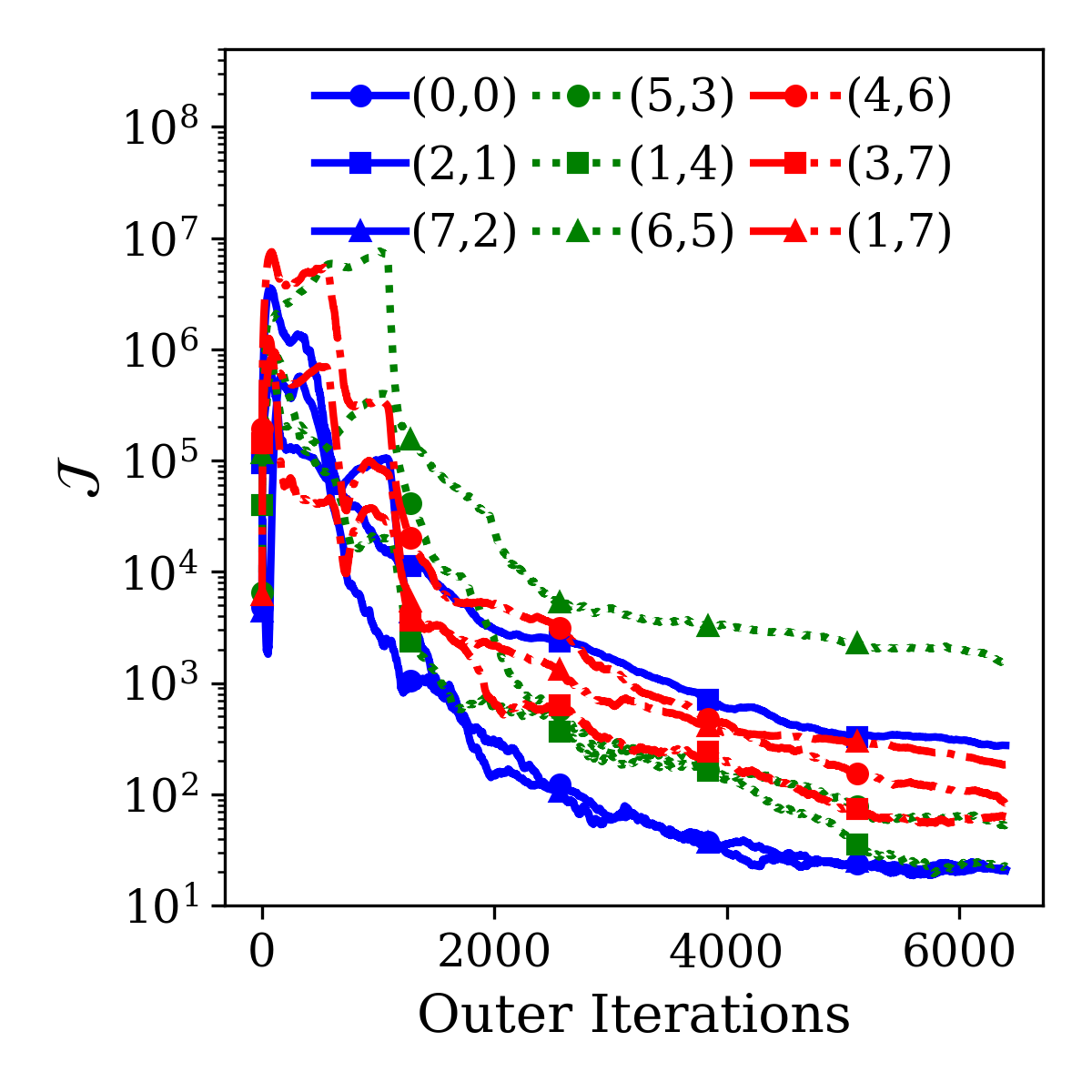}}\quad
    \subfloat[]{\includegraphics[scale=0.45]{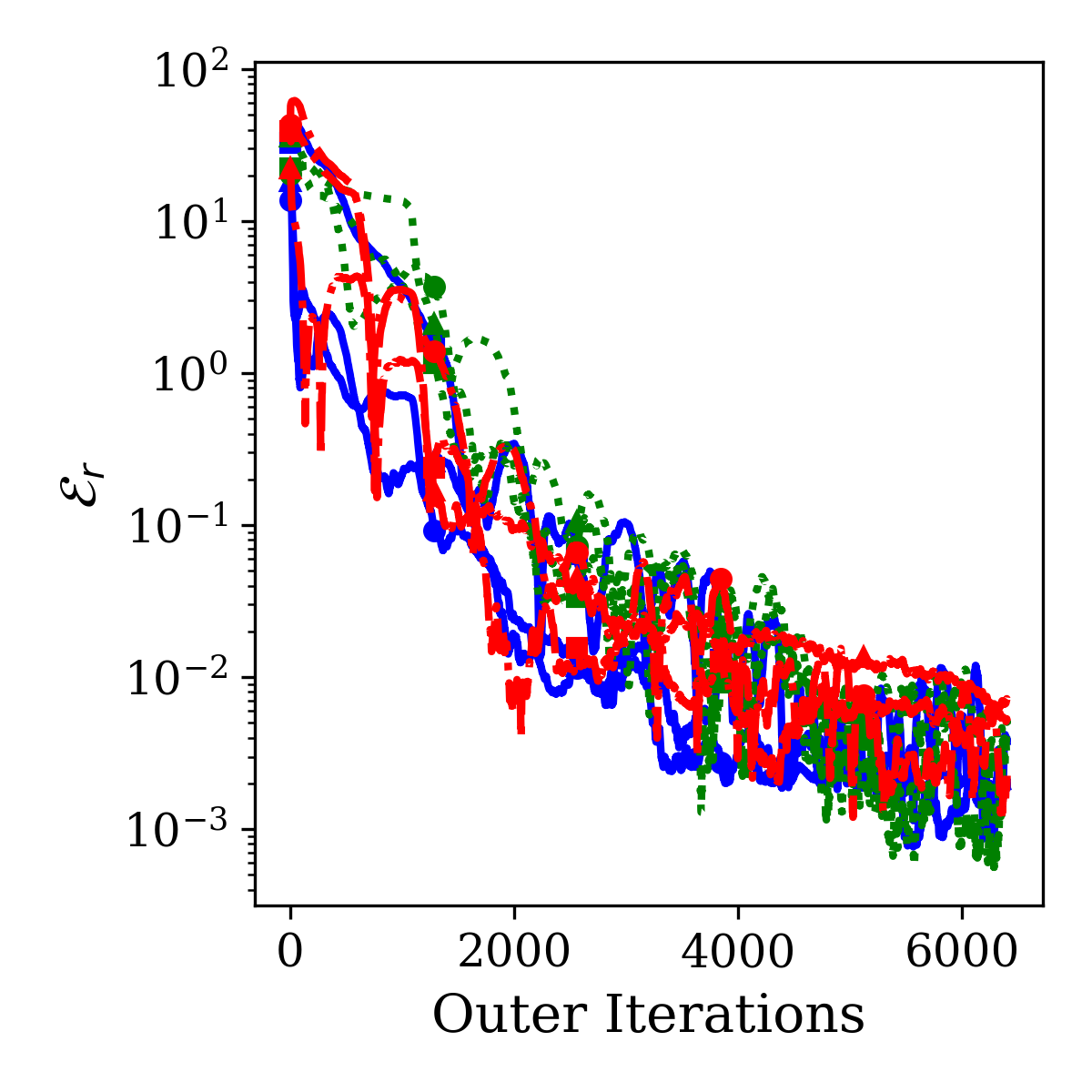}}\quad
    \subfloat[]{\includegraphics[scale=0.45]{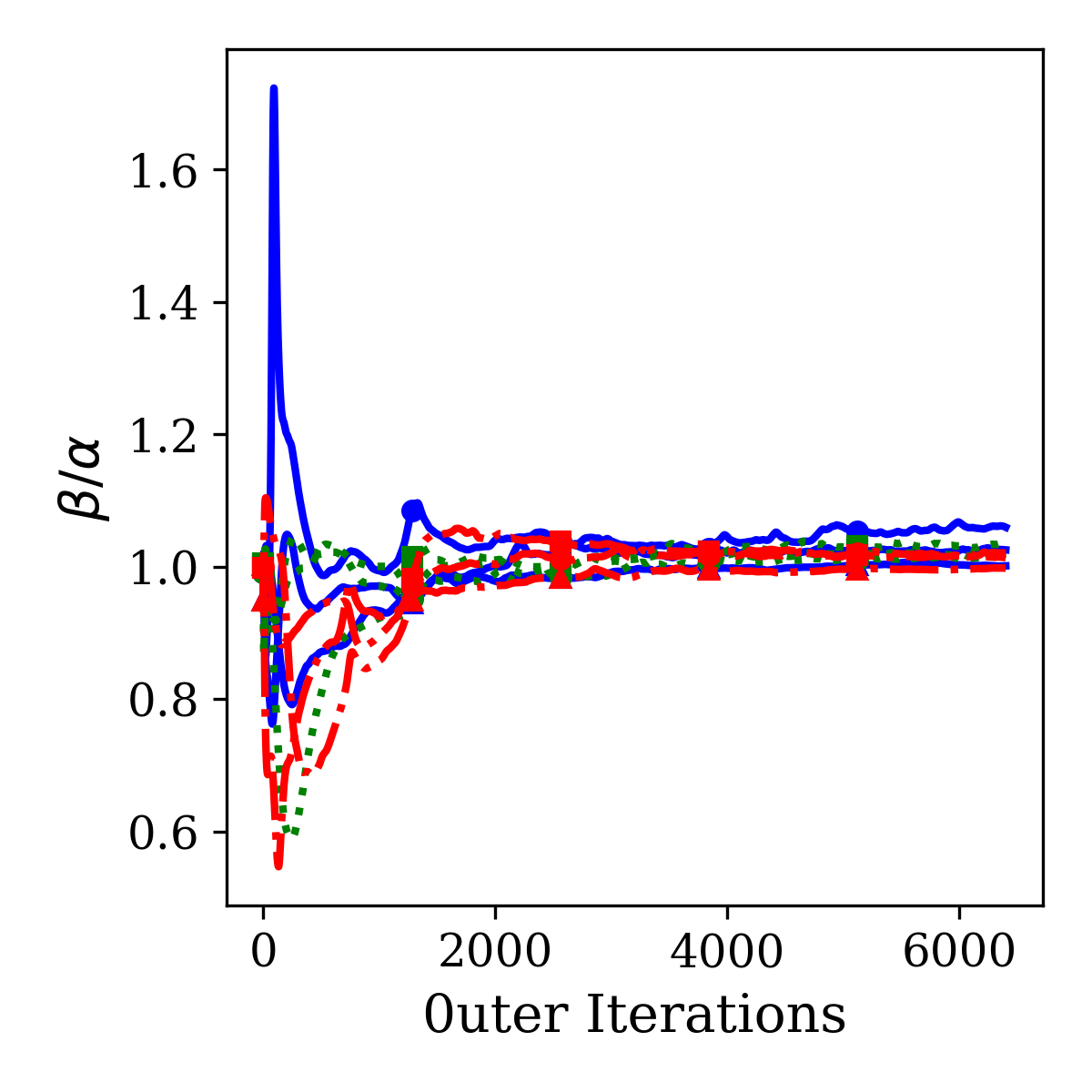}}\quad
    \subfloat[]{\includegraphics[scale=0.45]{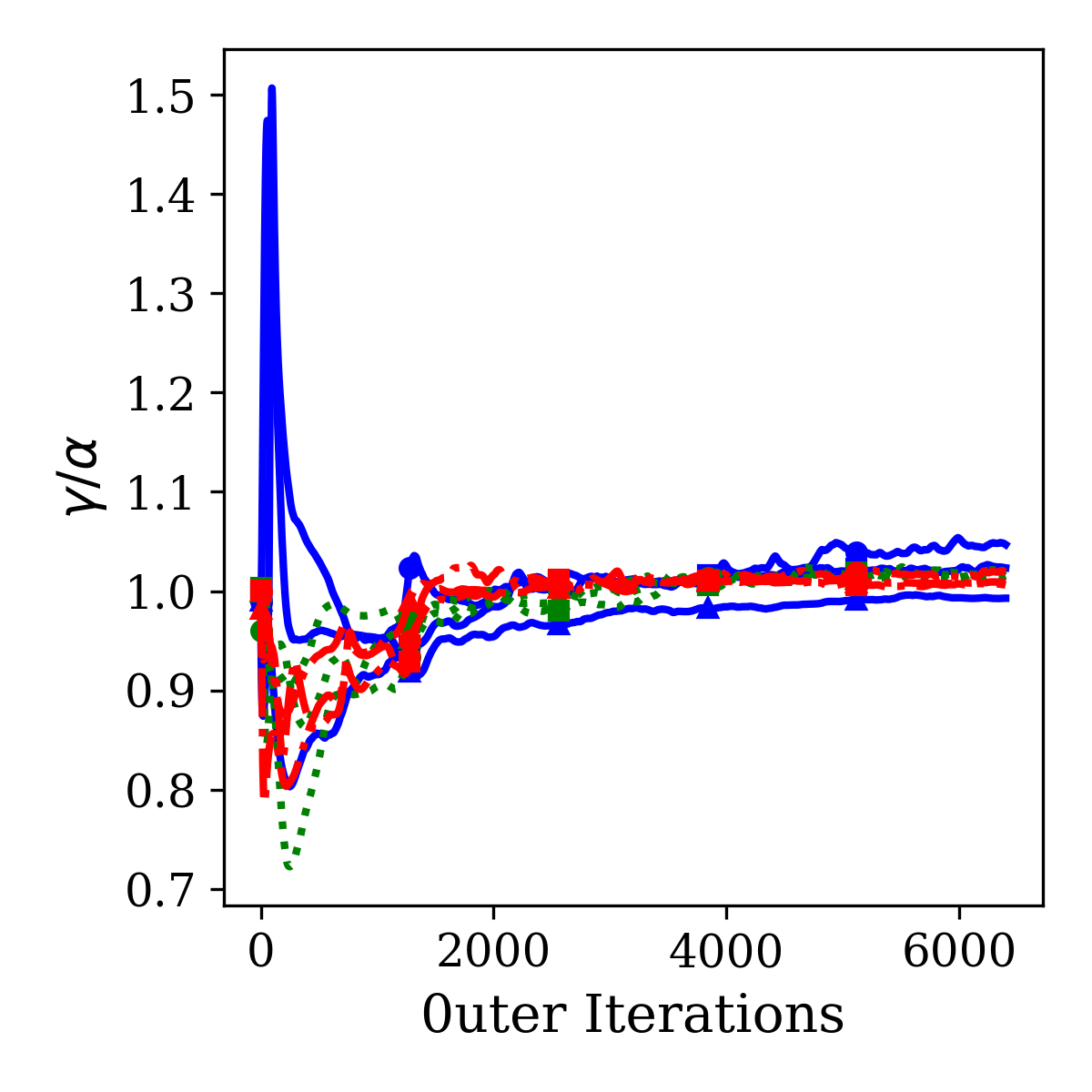}}\quad
    \caption{Poisson's equation with a multi-scale solution: Evolution of (a) PDE constraints, (b) boundary condition constraints, (c) objective functions, (d) relative $\mathit{l^2}$ norms, (e) the ratio $\beta / \alpha$, (e) the ratio $\gamma / \alpha$ for nine randomly selected subdomains.
}
\label{fig:multiscale_poisson_update}
\end{figure}

We randomly select nine out of the 64 subdomains to illustrate the evolution of the PDE and boundary condition constraints, objective functions, relative $\mathit{l^2}$ norms, and two interface ratios in Fig.~\ref{fig:multiscale_poisson_update}. \textcolor{black}{Figs.~\ref{fig:multiscale_poisson_update}(a-c) demonstrate that after roughly 5000 outer iterations, the PDE and boundary condition constraints reach magnitudes of approximately \(10^{-3}\) and \(10^{-5}\), respectively, while the objective functions decrease by more than four orders of magnitude to around \(10^{2}\).}
Compared to the previous case in Fig.~\ref{fig:poisson_simple_update}, the larger values and extended convergence time underscore the increased difficulty of the problem with a multi-scale solution. Notably, the objective function in Fig.~\ref{fig:multiscale_poisson_update}(c) spikes significantly at the start of training, indicating that the algorithm prioritizes on reducing the more challenging physics and boundary condition constraints. The subdomain (1,7), which exhibits the maximum error in Fig.~\ref{fig:multiscale_poisson_solution}(b), appears to stall in a local minimum, showing slower convergence of its boundary condition constraints.
Nonetheless, the evolution of these three loss terms follows the intended prioritization strategy, resulting in a prediction with good accuracy, as shown in the Fig.~\ref{fig:multiscale_poisson_update}(d).

Since our primary focus is on the convergence characteristics of the interface ratios rather than the interface parameters, Figs.~\ref{fig:multiscale_poisson_update}(e-f) show only the evolution of the interface ratios $\dfrac{\beta}{\alpha}$ and $\dfrac{\gamma}{\alpha}$. Both ratios exhibit significant oscillations during the initial stages of training but stabilize to values between 0.9 and 1.1 after approximately 2000 outer iterations. 
These oscillations reflect the evolution of the objective function during this period, with extremely large values indicating adjustments of the interface parameters towards a balanced state among the Dirichlet, Neumann, and tangential derivative continuity operators.

\subsubsection{\textcolor{black}{Helmholtz equation with a complex-valued solution}}\label{sec:helmholtz_complex}
\textcolor{black}{
The Helmholtz equation governs propagation phenomena. Its solution on partitioned domains requires specialized treatment due to its unique characteristics. Unlike the Poisson's equation, the Helmholtz equation's discretization typically produces a symmetric but non-positive definite matrix \cite{quarteroni1999DDM}. 
In the context of conventional numerical methods (e.g. finite difference method), separate domain decomposition methods have been proposed to tackle Poisson's and Helmholtz equations (e.g. \cite{gander2007_two_sided_osm}). This is  because transmission conditions that work well for the Poisson's equation does not readily extend to handle the Helmholtz equation.
To this end, solving the Helmholtz and Poisson's equations with a unified domain decomposition framework presents substantial technical challenges.}

\textcolor{black}{
Here, we adopt a complex-valued, 2D Helmholtz equation problem that has an exact solution. This problem was used by \citet{babuska_1995} to validate a generalized finite element method for the Helmholtz equation to reduce the rising error with higher wave number. This example was also adopted in the work of \citet{bao_2003} to validate a discrete singular convolution algorithm for solving the Helmholtz equation with high wavenumbers.}

\textcolor{black}{
The exact solution to the problem is a planar wave of the form $u(x,y)=\exp {(j(k_1x + k_2y))}$, where $j$ is the imaginary unit, in a unit square $\Omega = {(x,y)|0 \leq x,y \leq 1}$ with a Sommerfeld-type radiation boundary condition as follows:
\begin{equation}
    \begin{aligned}
    - \Delta u - k^2 u & = 0, && \text{in} \quad \Omega, \\
     jku + \frac{\partial u}{\partial n} & = g, && \text{on} \quad \partial \Omega,
    \label{eq:helmholtz_complex}
    \end{aligned}
\end{equation}
where $(k_1, k_2) = k(\cos \theta, \sin \theta)$ and the function $g$ depends on the wave direction $\theta$.
Here, we set $k=\pi*2^4$ and $\theta = \dfrac{\pi}{3}$. We use a $4\times4$ domain decomposition with a $128\times128$ global uniform mesh. The neural network architecture for each subdomain is identical to that used in the previous Poisson’s equation cases, with the exception that it now produces two outputs, corresponding to the real and imaginary parts of the solution, respectively. In terms of the definition of boundary loss terms, the squared magnitude of a complex-valued residual is equivalent to the summation of the squared real part and imaginary parts. For instance, the complex value of the boundary residual operator $\mathcal{B}(\theta;\mathbf{x}_i)$ on the $i$th boundary point is $\mathfrak{Re}[\mathcal{B}(\theta;\mathbf{x}_i)] + j * \mathfrak{Im}[\mathcal{B}(\theta;\mathbf{x}_i)]$, then the squared magnitude is
\begin{equation}
    ||\mathcal{B}(\theta;\mathbf{x}_i)||_2^2 = \mathfrak{Re}^2[\mathcal{B}(\theta;\mathbf{x}_i)] + \mathfrak{Im}^2[\mathcal{B}(\theta;\mathbf{x}_i)].
\end{equation}
In the same way, we keep the same real-valued interface parameters $\mathbf{q}$ as in previous examples, and the interface loss term on the $i$th interface point can be written as
\begin{equation}
    \mathcal{J}(\theta;\mathbf{x}_i) \approx \sum_{k=1}^3 q_k^2 ||\mathcal{O}^k_i||_2^2 = \sum_{k=1}^3 q_k^2 \mathfrak{Re}^2(\mathcal{O}^k_i) + q_k^2 \mathfrak{Im}^2(\mathcal{O}^k_i).
\end{equation}
Therefore, we separate the real and imaginary parts and avoid the occurrence of cross-product terms, as argued in Section \ref{sec:interface_loss}, when implementing the complex-valued interface parameters. We note that our proposed method remains unchanged.}

\textcolor{black}{
Table~\ref{tab:helm_complex_l2_linf} summarizes the statistical performance of the proposed method over 5 independent trials, in terms of the relative $\mathit{l^2}$ and $\mathit{l^\infty}$ norms for real and imaginary parts of the solution. The mean and standard deviation of $\mathcal{E}_r$ with an order of magnitude of $10^{-4}$ suggest that an accurate solution can be predicted consistently with each trial. The levels of $\mathit{l^\infty}$ norm, which is a stricker norm than the $\mathit{l^2}$ as a metric of accuracy, further support this outcome.
}
\begin{table}[!t]
\centering
\begin{tabular}{ccc}
\hline
Complex-valued prediction & Relative $\mathit{l^2}$ norm    & $\mathit{l^\infty}$ norm          \\ \hline
$\mathfrak{Re}$ part          & $3.76e-4 \pm 4.87e-4$ & $5.28e-2 \pm 7.08e-2$   \\
$\mathfrak{Im}$ part         & $1.46e-4 \pm 1.28e-4$ & $1.47e-2 \pm 1.27e-2$ \\
 \hline
\end{tabular}
\caption{The relative $\mathit{l^2}$ and $\mathit{l^\infty}$ norms for the real and imaginary parts of the predicted solution representing five independent trials.}
\label{tab:helm_complex_l2_linf}
\end{table}

\begin{figure}[t!]
    \centering
    \subfloat[]{\includegraphics[scale=0.45]{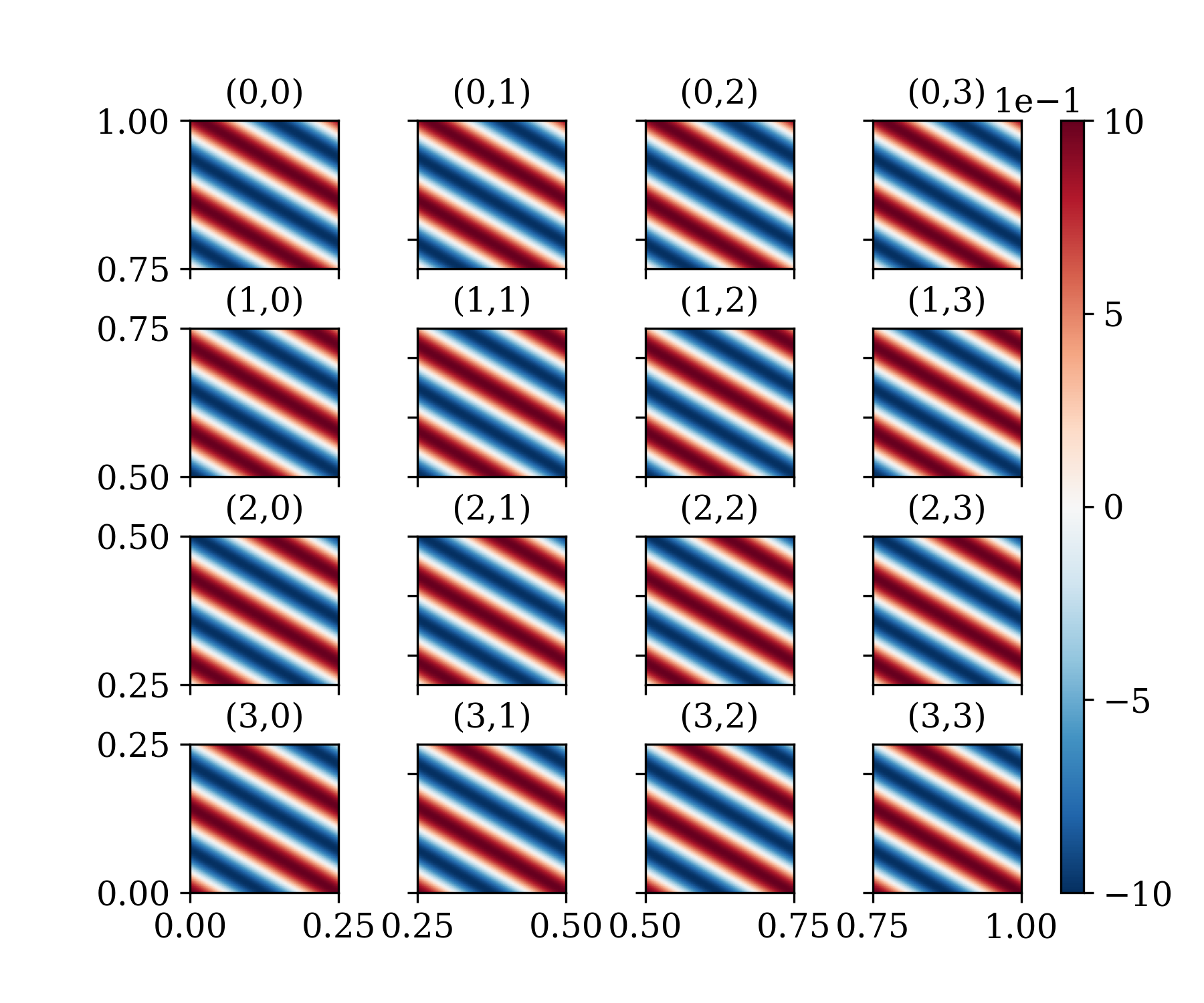}}\quad
    \subfloat[]{\includegraphics[scale=0.45]{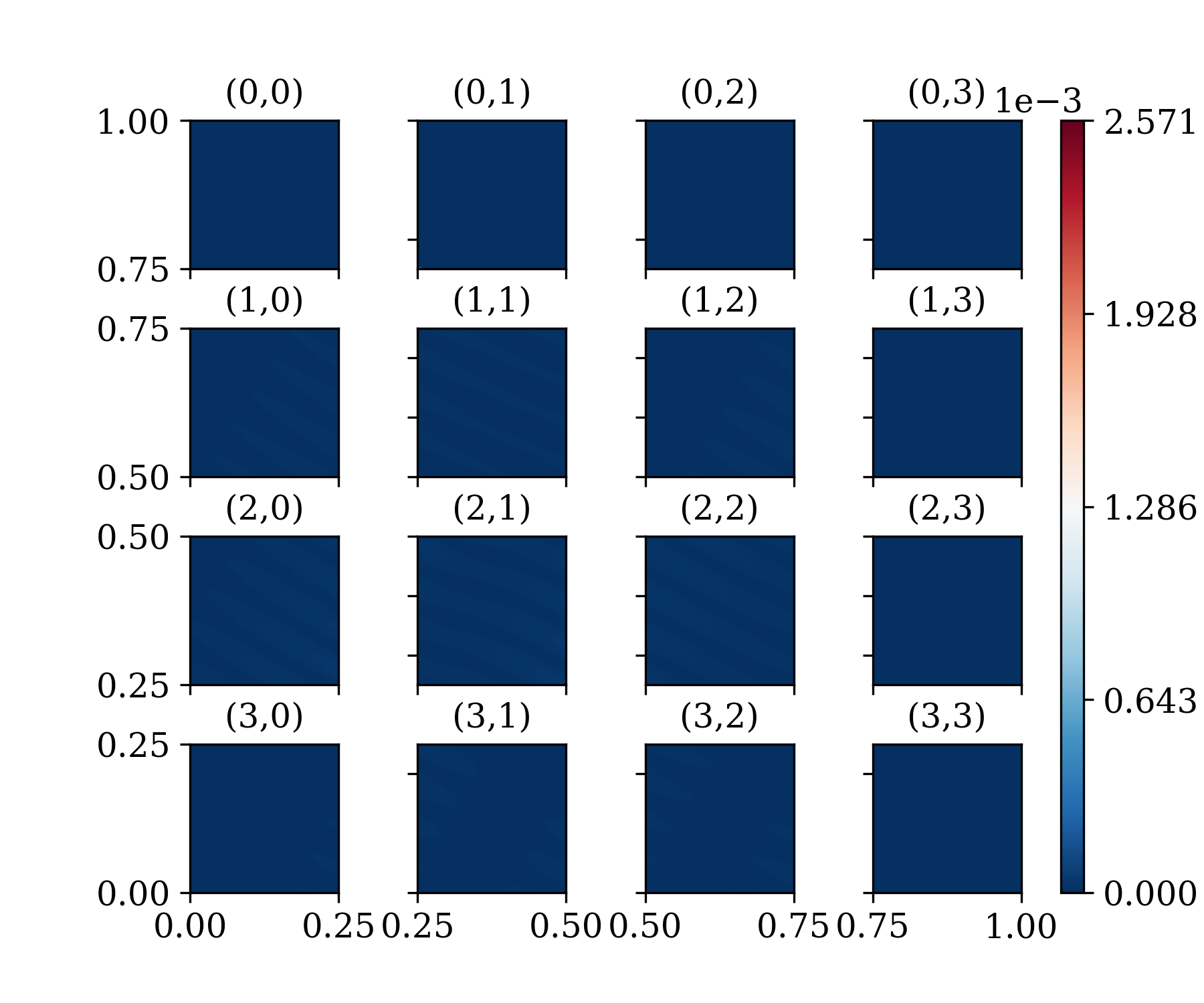}}\quad
    \subfloat[]{\includegraphics[scale=0.45]{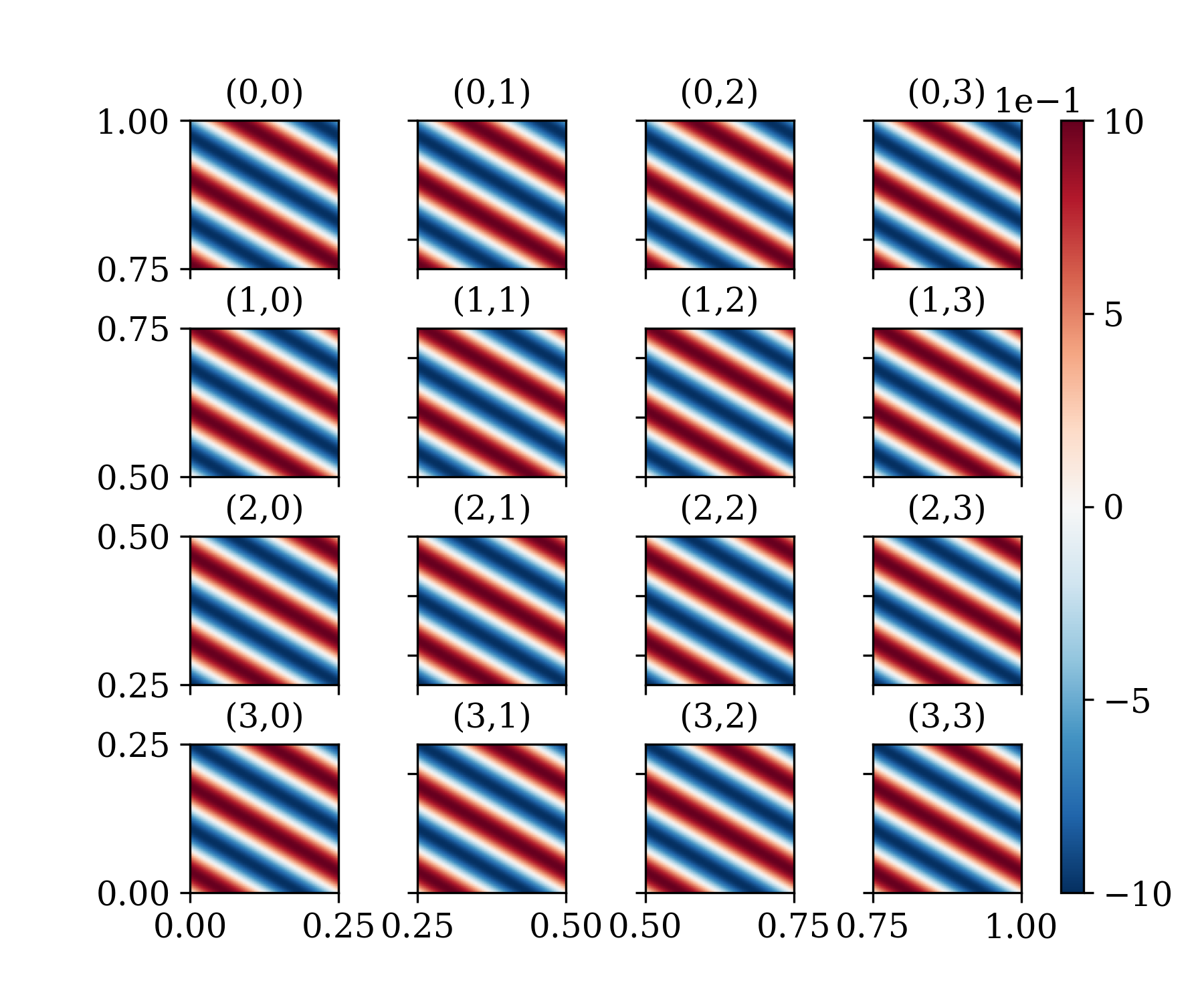}}\quad
    \subfloat[]{\includegraphics[scale=0.45]{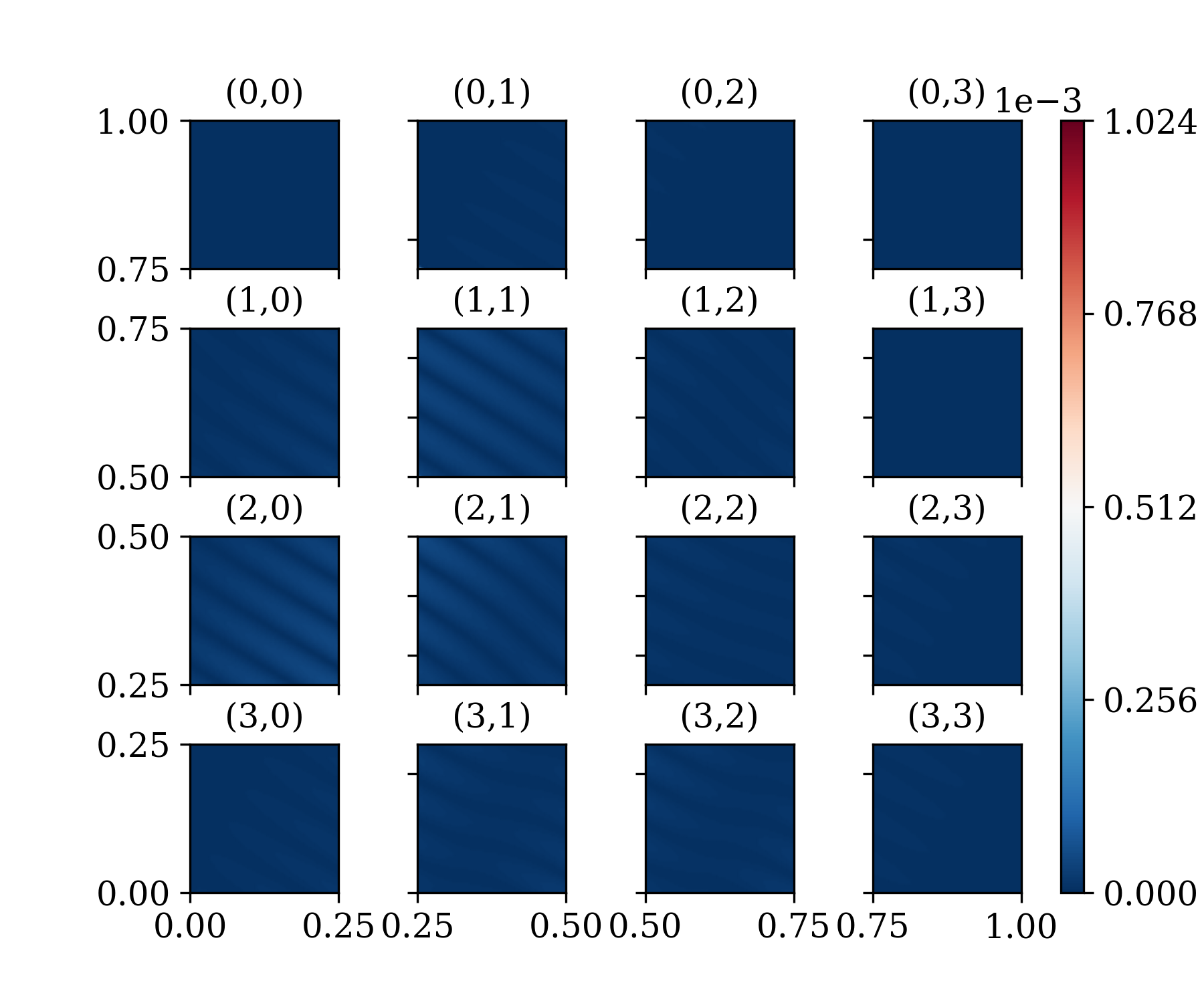}}\quad
    \caption{2D Helmholtz equation problem with a complex-valued solution: (a) real part of the predicted solution, (b) real part of the absolute point-wise error, (c) imaginary part of the predicted solution, (d) imaginary part of the absolute point-wise error from the best trial. }
    \label{fig:helmholtz_complex_solution}
\end{figure}

\textcolor{black}{
Figure~\ref{fig:helmholtz_complex_solution} shows the distribution of the predicted solution and point-wise error from the best trial. The wave propagation with a phase lag is visible on both real and imaginary parts of the solution in Fig.~\ref{fig:helmholtz_complex_solution}(a) and Fig.~\ref{fig:helmholtz_complex_solution}(c), respectively. The absence of wave reflection or distortion along the boundaries indicates that the Sommerfeld-type radiation boundary conditions are effectively handled by our PECANN method. From Fig.~\ref{fig:helmholtz_complex_solution}(b) and Fig.~\ref{fig:helmholtz_complex_solution}(d) we observe that the absolute error is evenly distributed across the domain, with no significant accumulation along the interfaces or boundaries.}

\begin{figure}[t!]
\centering
    \subfloat[]{\includegraphics[scale=0.45]{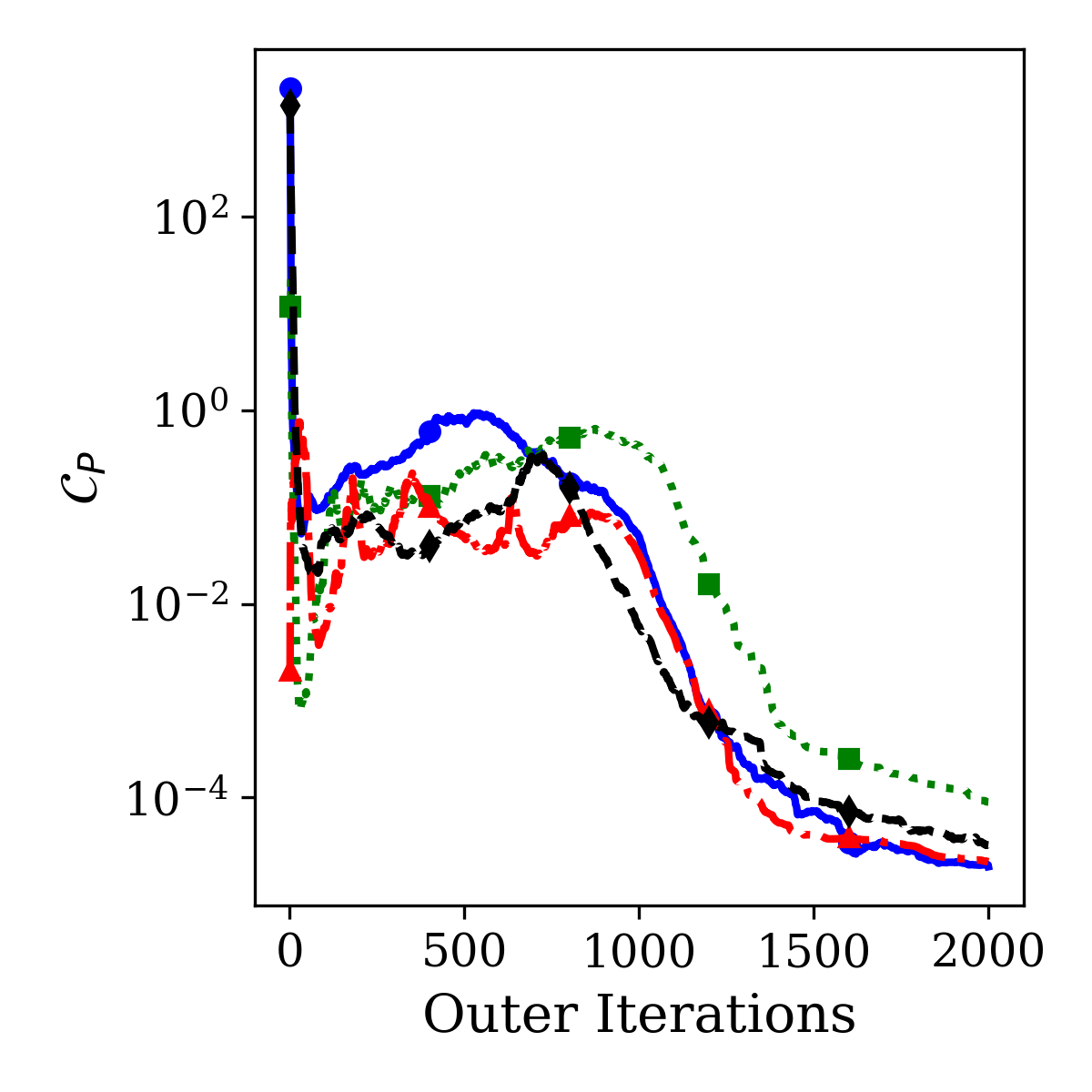}}\quad
    \subfloat[]{\includegraphics[scale=0.45]{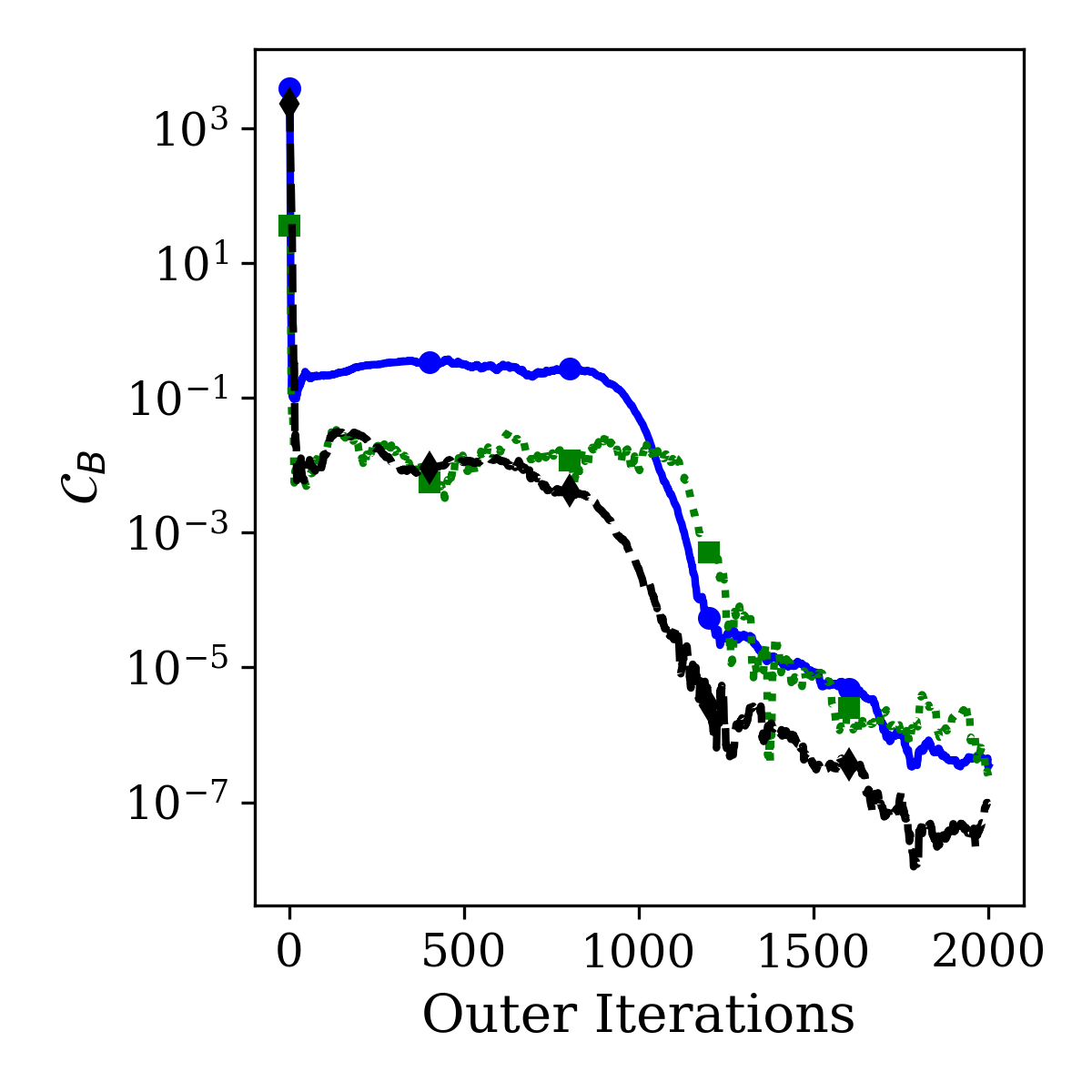}}\quad
    \subfloat[]{\includegraphics[scale=0.45]{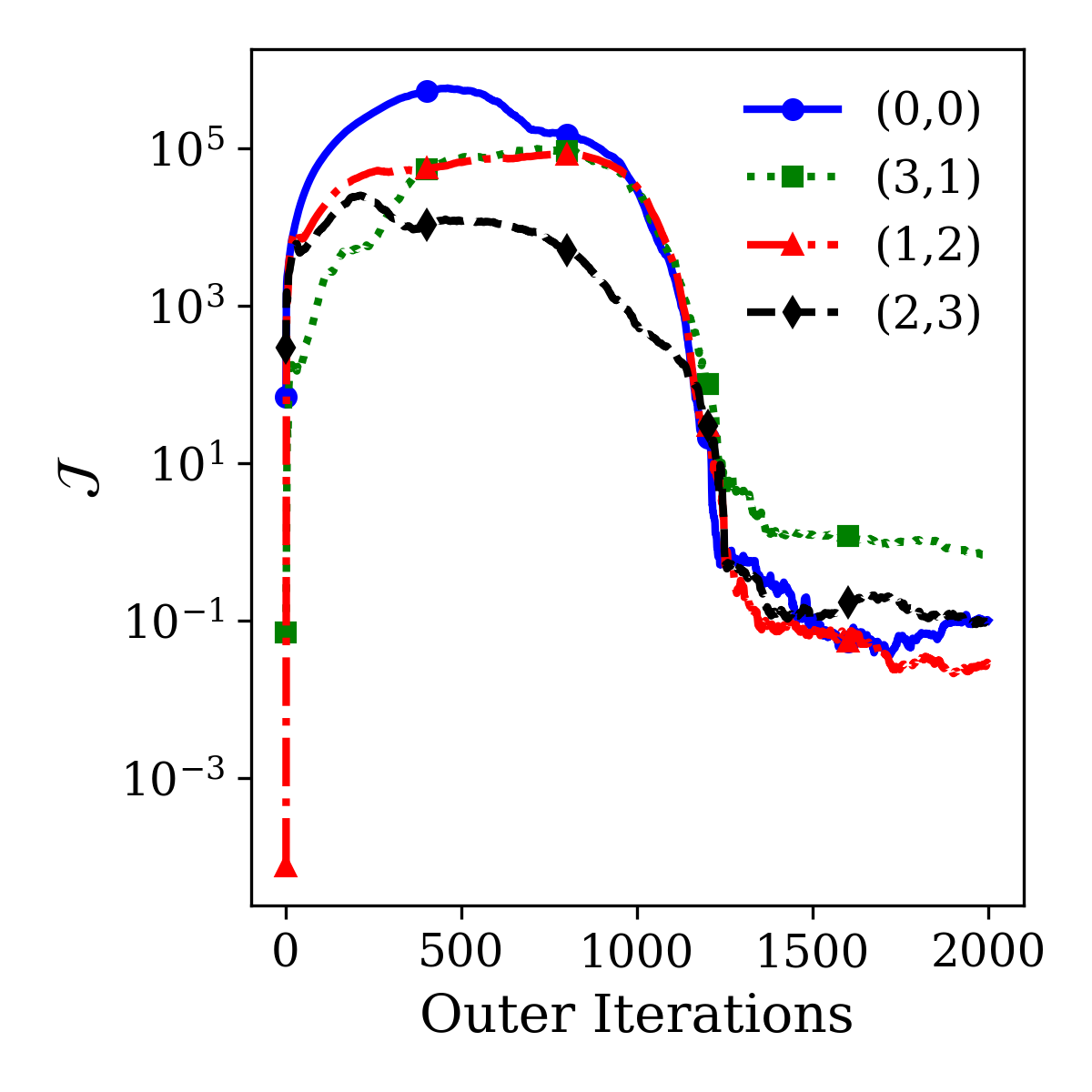}}\quad
    \subfloat[]{\includegraphics[scale=0.45]{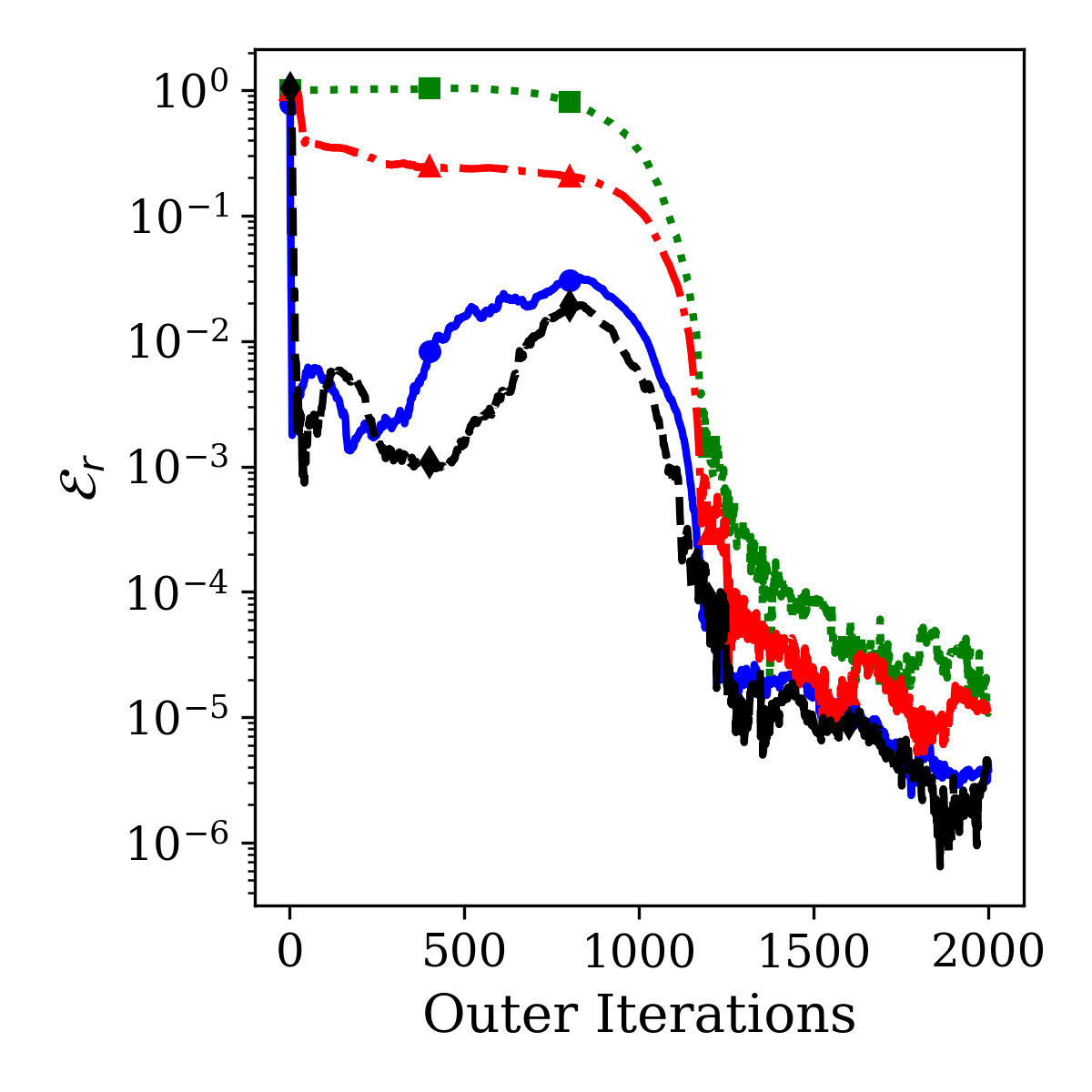}}\quad
    \subfloat[]{\includegraphics[scale=0.45]{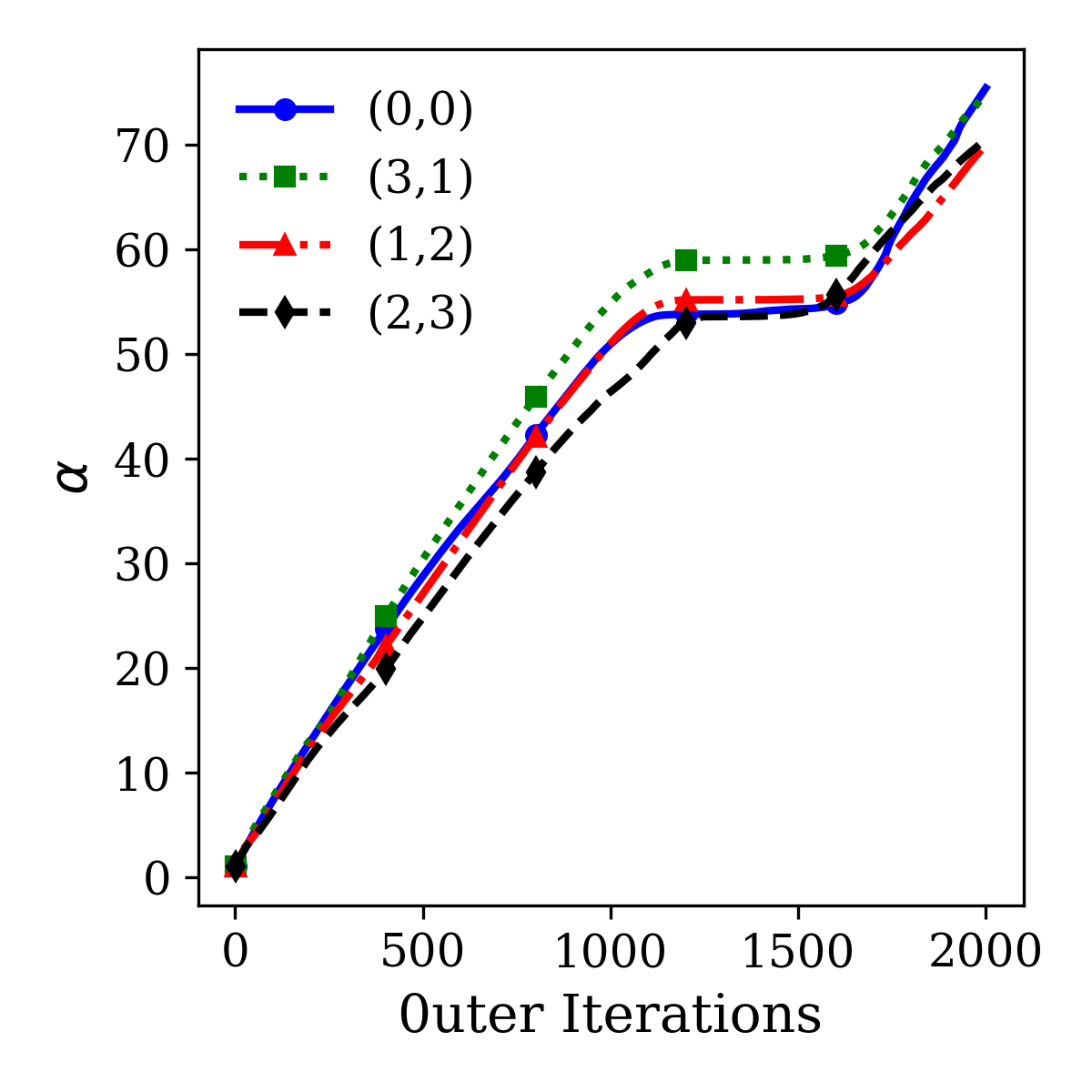}}\quad
    \subfloat[]{\includegraphics[scale=0.45]{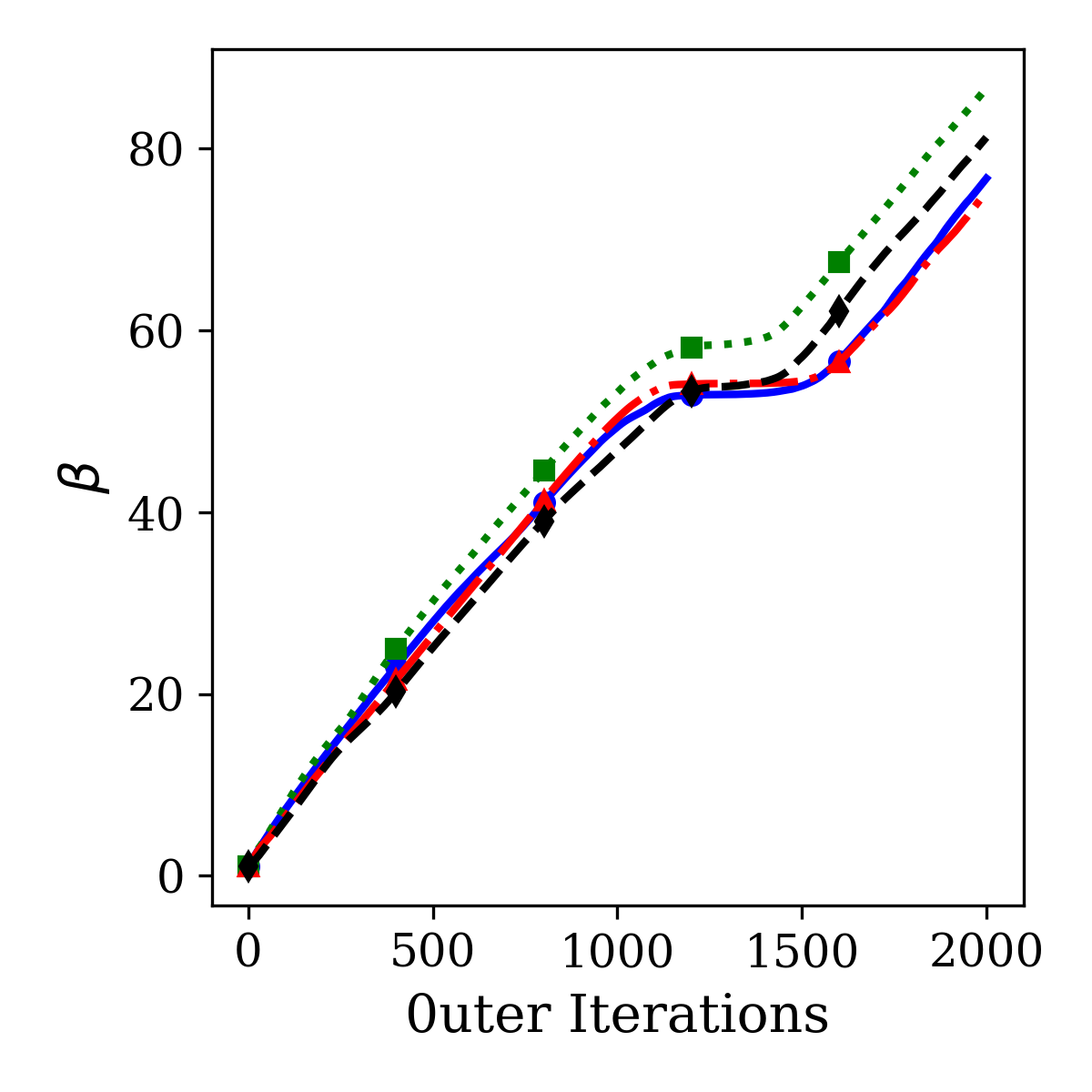}}\quad
    \caption{Helmholtz equation with a complex-valued solution: Evolution of (a) PDE constraints, (b) boundary condition constraints, (c) objective functions, (d) relative $\mathit{l^2}$ norms, (e) interface parameter $\alpha$, (f) interface parameter $\beta$ for four randomly-selected subdomains.
}
\label{fig:helmholtz_complex_update}
\end{figure}

\textcolor{black}{
For ease of presentation, we select four subdomains out of the total 16. Figure~\ref{fig:helmholtz_complex_update} presents the evolution of the individual loss terms, relative $\mathit{l^2}$ norms and the  two interface parameters chosen. From Fig.~\ref{fig:helmholtz_complex_update}(a-d), we observe that both boundary and PDE constraints and the objective function as well as the relative error $\mathcal{E}_r$ rapidly decrease after around 1000 outer iterations. For outer iterations ranging from \(1000\) to \(1600\), interface parameters $\alpha$ and $\beta$ shown in Fig.~\ref{fig:helmholtz_complex_update}(e) and Fig.~\ref{fig:helmholtz_complex_update}(f), respectively, plateau at a value around 50. After \(1600\) outer iterations, the interface parameters continue their upward trend; however, the objective functions converge to low values, indicating that the interface operators are being satisfactorily met.}

\subsubsection{\textcolor{black}{Helmholtz equation with a real-valued, high-wavenumber solution}}\label{sec:helmholtz}

Following the work of multilevel FBPINNs \cite{DOLEAN2024}, we consider the same 2D Helmholtz problem with a Dirichlet boundary condition and a strong, Gaussian-like point source term centered in the domain \(\Omega = \{(x, y) ~ | ~ 0 \leq x, y \leq 1\}\). \citet{DOLEAN2024} investigated this problem by systematically increasing its model complexity and showed that beyond a certain level, various methods, including the multilevel FBPINN, perform poorly with some methods worse than the others. This challenging Helmholtz problem with a real-valued, high-wavenumber solution is described as follows:
\begin{equation}
    \begin{aligned}
    \nabla^2 u + k^2 u & = \frac{1}{2\pi \sigma^2} \exp{ -\frac{\|\mathbf{x} - 0.5\|_2^2}{2\sigma^2} }, && \text{in} \quad \Omega, \\
     u & = 0, && \text{on} \quad \partial \Omega,
    \label{eq:helmholtz_2d}
    \end{aligned}
\end{equation}
where \(k = \dfrac{2^L\pi}{1.6}\) is the wave number, and \(\sigma = \dfrac{0.8}{2^L}\) defines the width of the Gaussian source. The complexity of the problem is controlled by the parameter $L$. Here we consider $L=5$, which corresponds to the 5-level FBPINN predictions in \citet{DOLEAN2024}.

We use the same uniform global mesh of \(160 \times 160\) as adopted in \cite{DOLEAN2024}. We partitioned our domain with a \(5 \times 5\) decomposition. In our Algorithm~\ref{alg:adaptive_training_algorithm}, the penalty scaling factor for boundary condition constraints, \(\eta_B\), is set to 10, and the smoothing constant \(\zeta\) is set to 0.999. Our neural network and optimizer configurations remain identical to those we have used in the Poisson's equation investigations. 
The exact solution to this problem is not known. Therefore, to evaluate the quality of the predictions, the same finite difference method, as adopted in \cite{DOLEAN2024},  on a mesh of \(360 \times 360\) is used to obtain the reference solution. 
\begin{figure}[t!]
    \centering
    \subfloat[]{\includegraphics[scale=0.45]{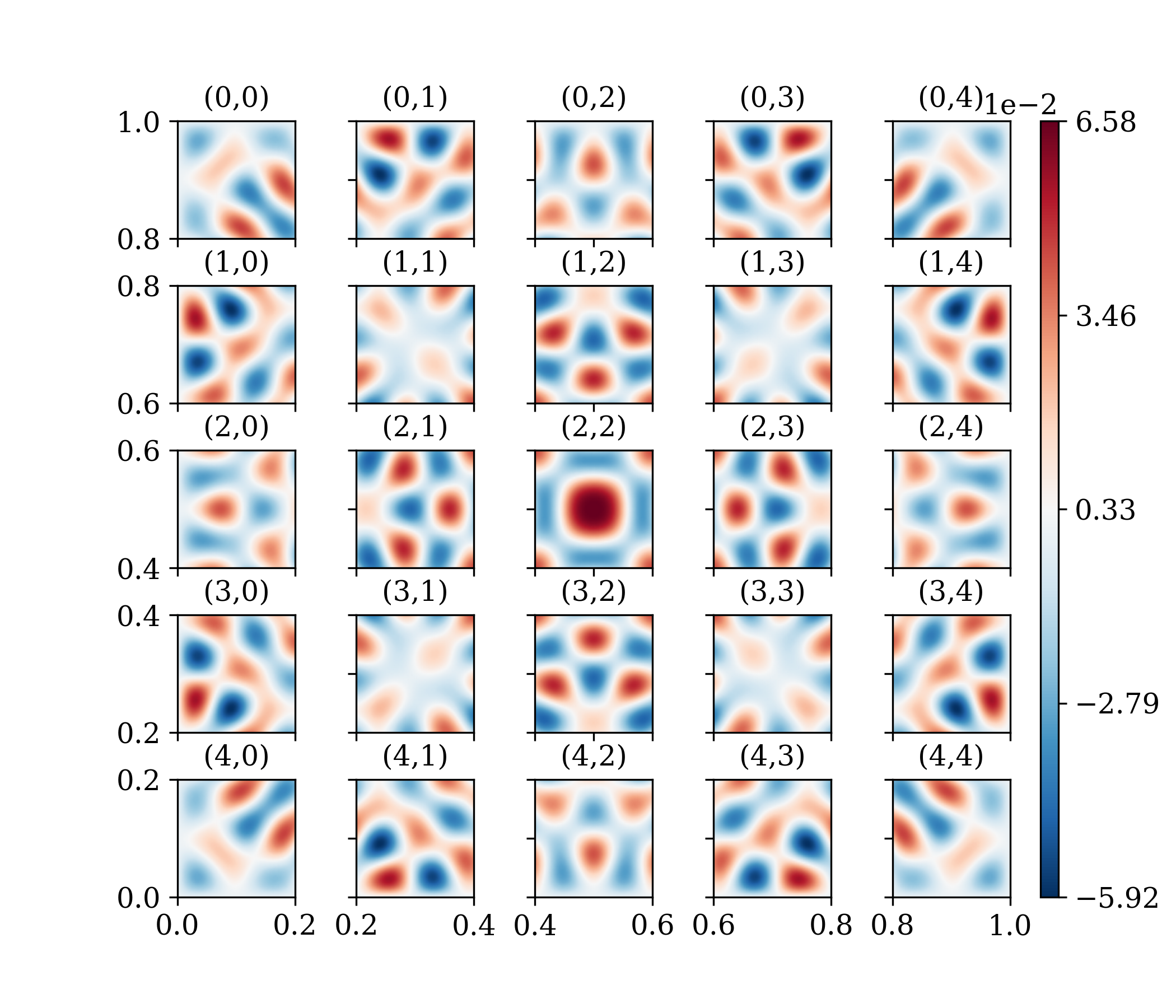}}\quad
    \subfloat[]{\includegraphics[scale=0.45]{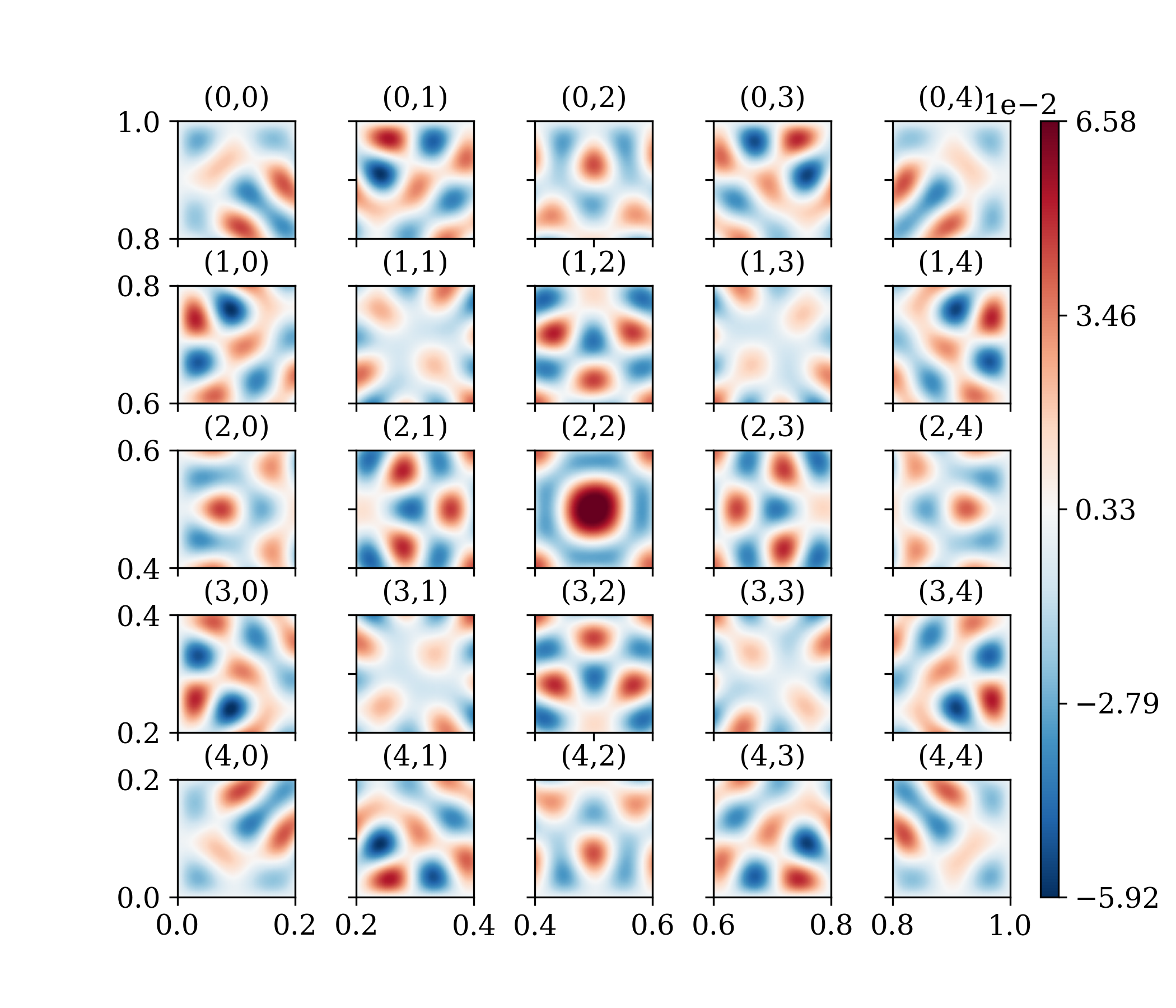}}\quad
    \subfloat[]{\includegraphics[scale=0.45]{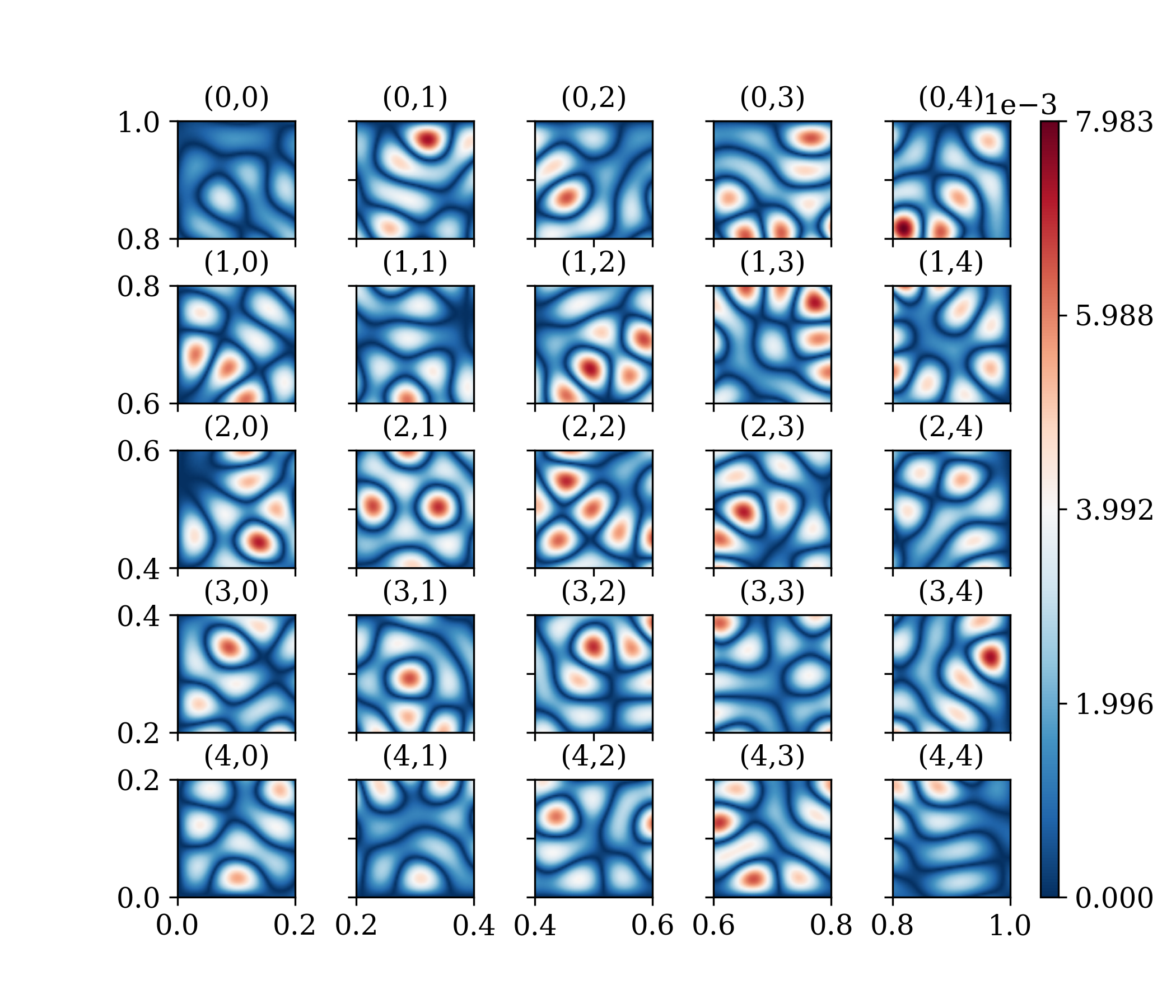}}\quad
    \caption{Helmholtz equation with a real-valued, high-wavenumber solution: (a) finite difference solution on a $5\times5$ partitioned domain, (b) predicted solution, (c) absolute point-wise error. }
    \label{fig:multiscale_helmholtz_solution_l5}
\end{figure}

Figure~\ref{fig:multiscale_helmholtz_solution_l5}a presents the finite difference solution. The predicted solution, and the absolute errors relative to the finite difference solution are shown in Fig.~\ref{fig:multiscale_helmholtz_solution_l5}b and Fig~\ref{fig:multiscale_helmholtz_solution_l5}c, respectively. Overall, the predicted solution from our DDM accurately captures all the key features of the reference solution, as seen by comparing Figs.~\ref{fig:multiscale_helmholtz_solution_l5}(a-b). However, the error distribution in Fig.~\ref{fig:multiscale_helmholtz_solution_l5}(c) is relatively uniform across most subdomains. We should note that when \(L \geq 5\), this problem tends to exhibit inconsistencies even when using the finite difference method with increasing mesh resolution, using the code provided in \citet{DOLEAN2024}.

\begin{figure}[t!]
\centering
    \subfloat[]{\includegraphics[scale=0.45]{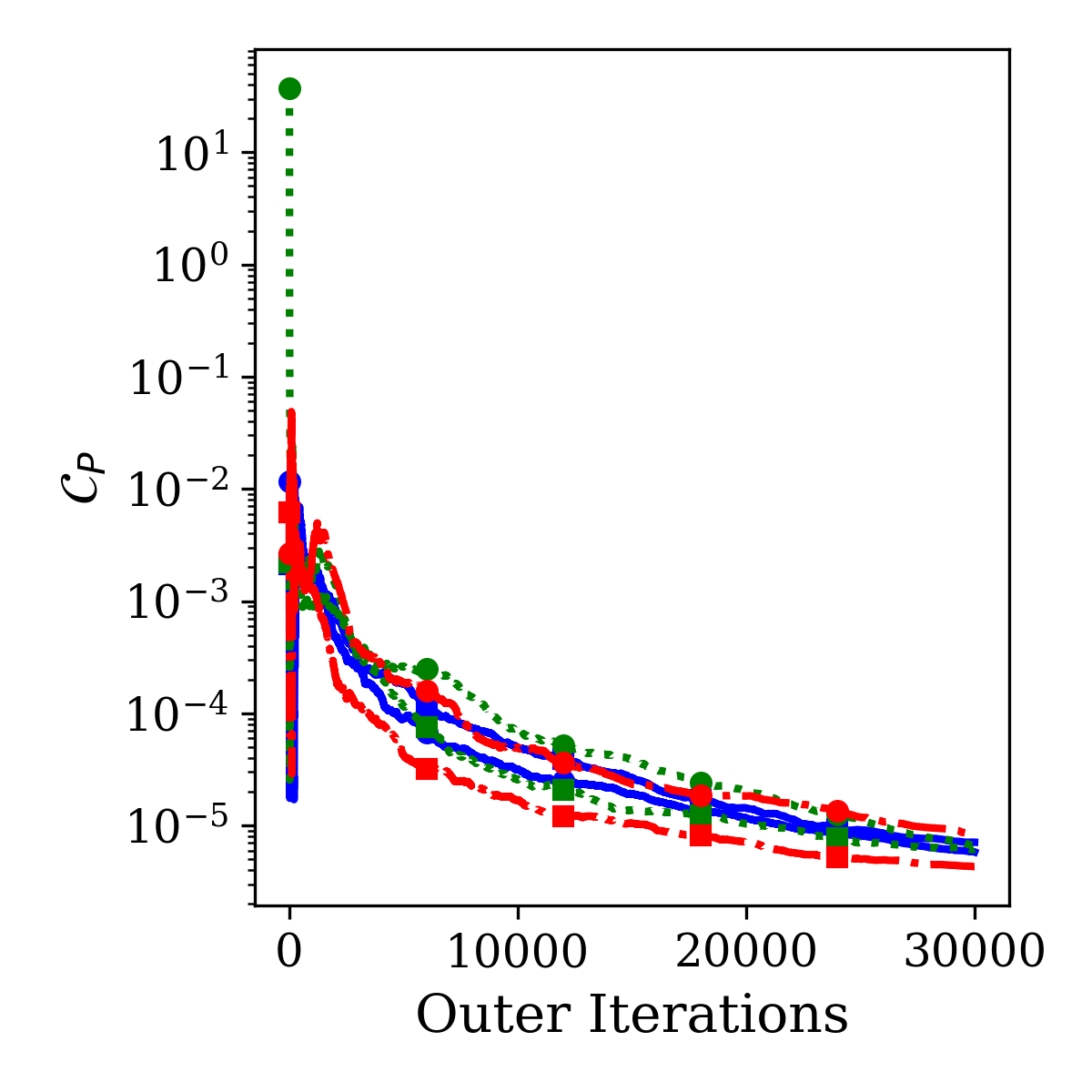}}\quad
    \subfloat[]{\includegraphics[scale=0.45]{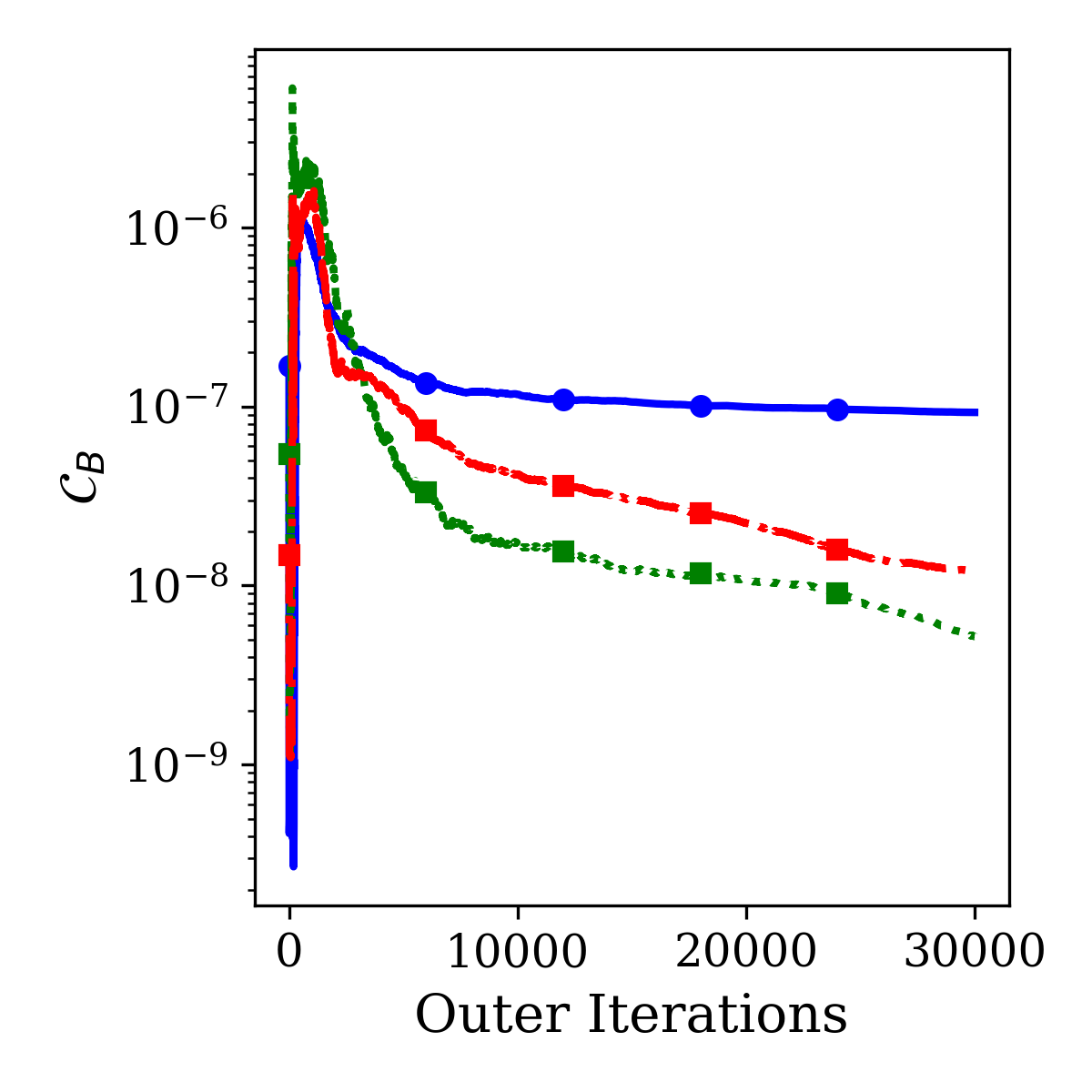}}\quad
    \subfloat[]{\includegraphics[scale=0.45]{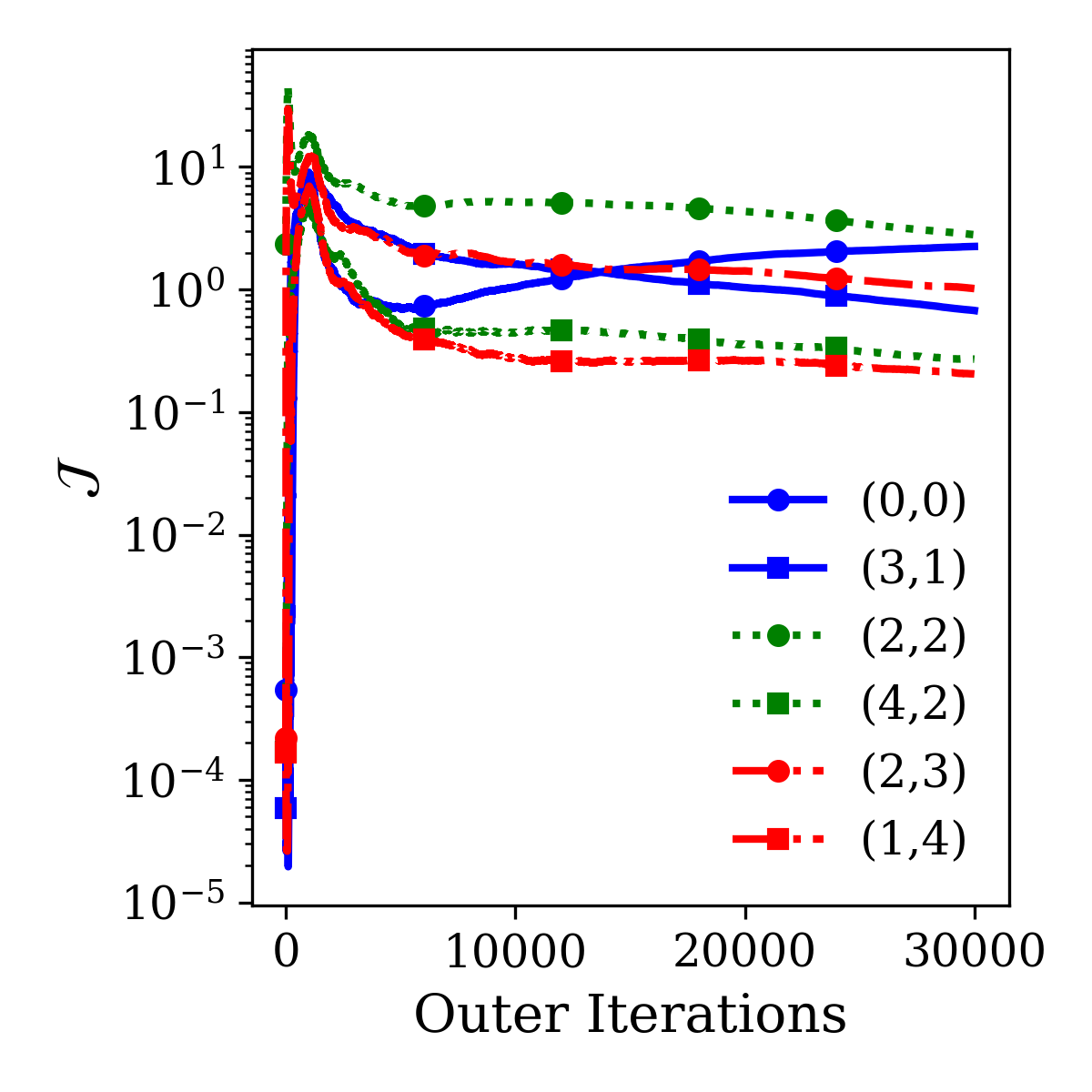}}\quad
    \subfloat[]{\includegraphics[scale=0.45]{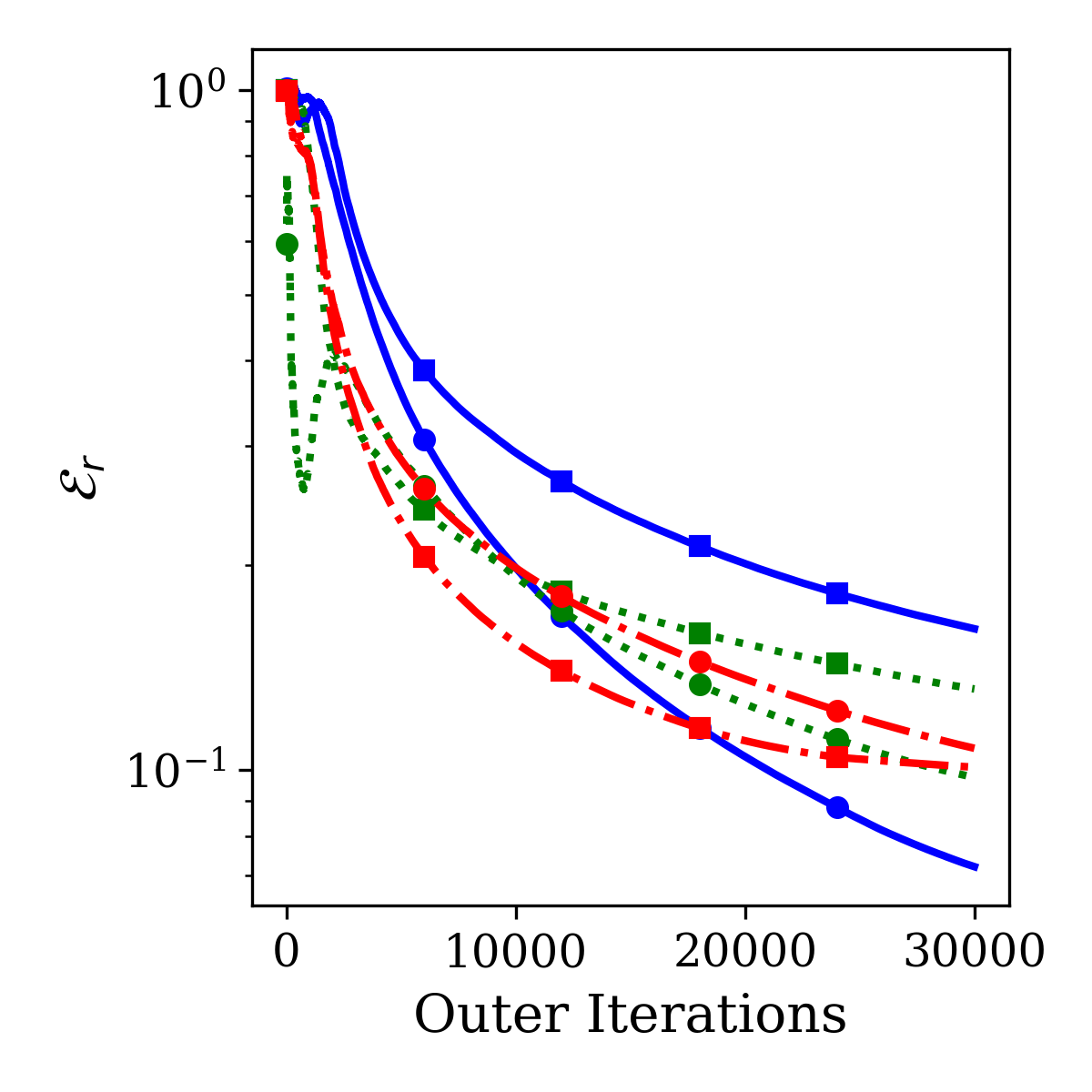}}\quad
    \subfloat[]{\includegraphics[scale=0.45]{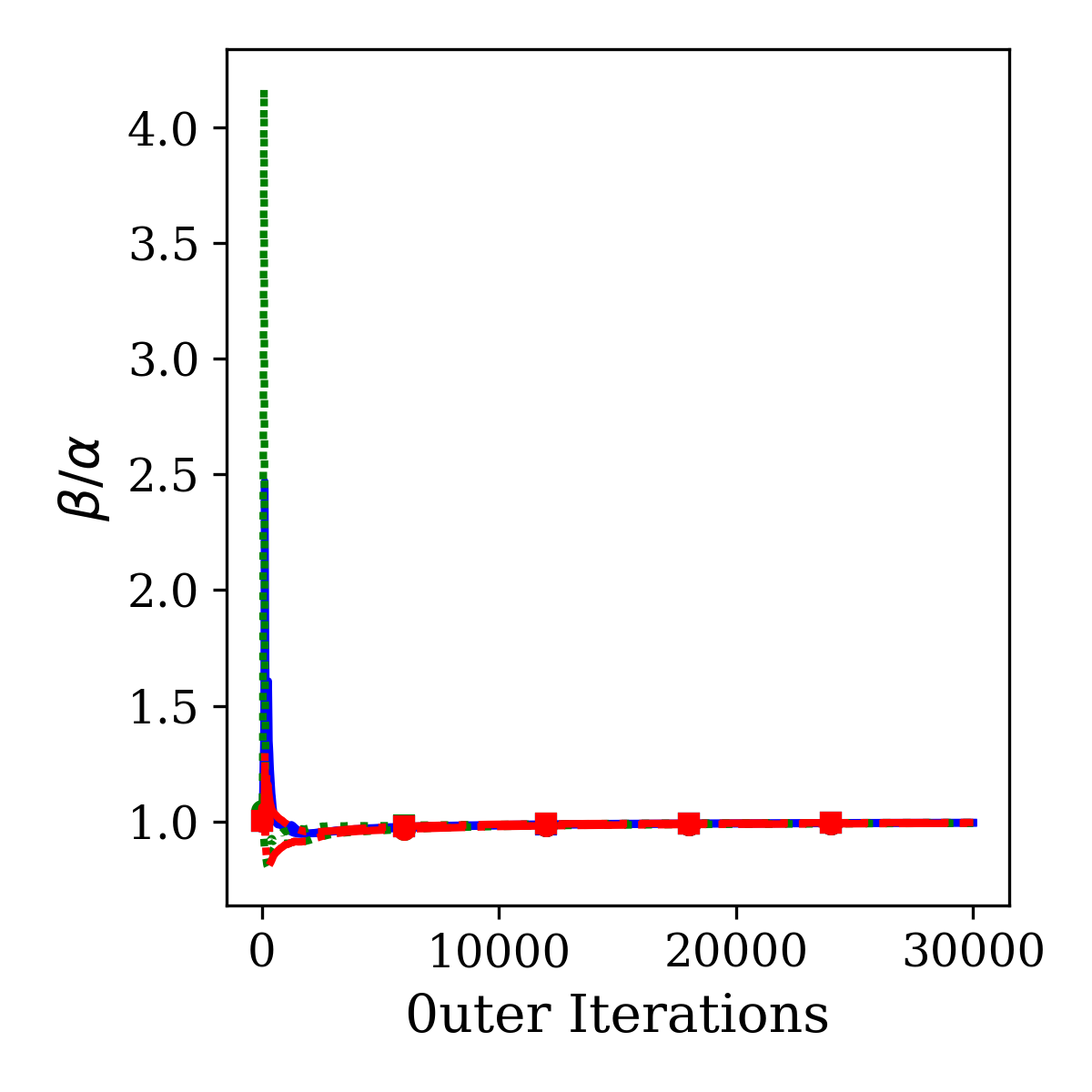}}\quad
    \subfloat[]{\includegraphics[scale=0.45]{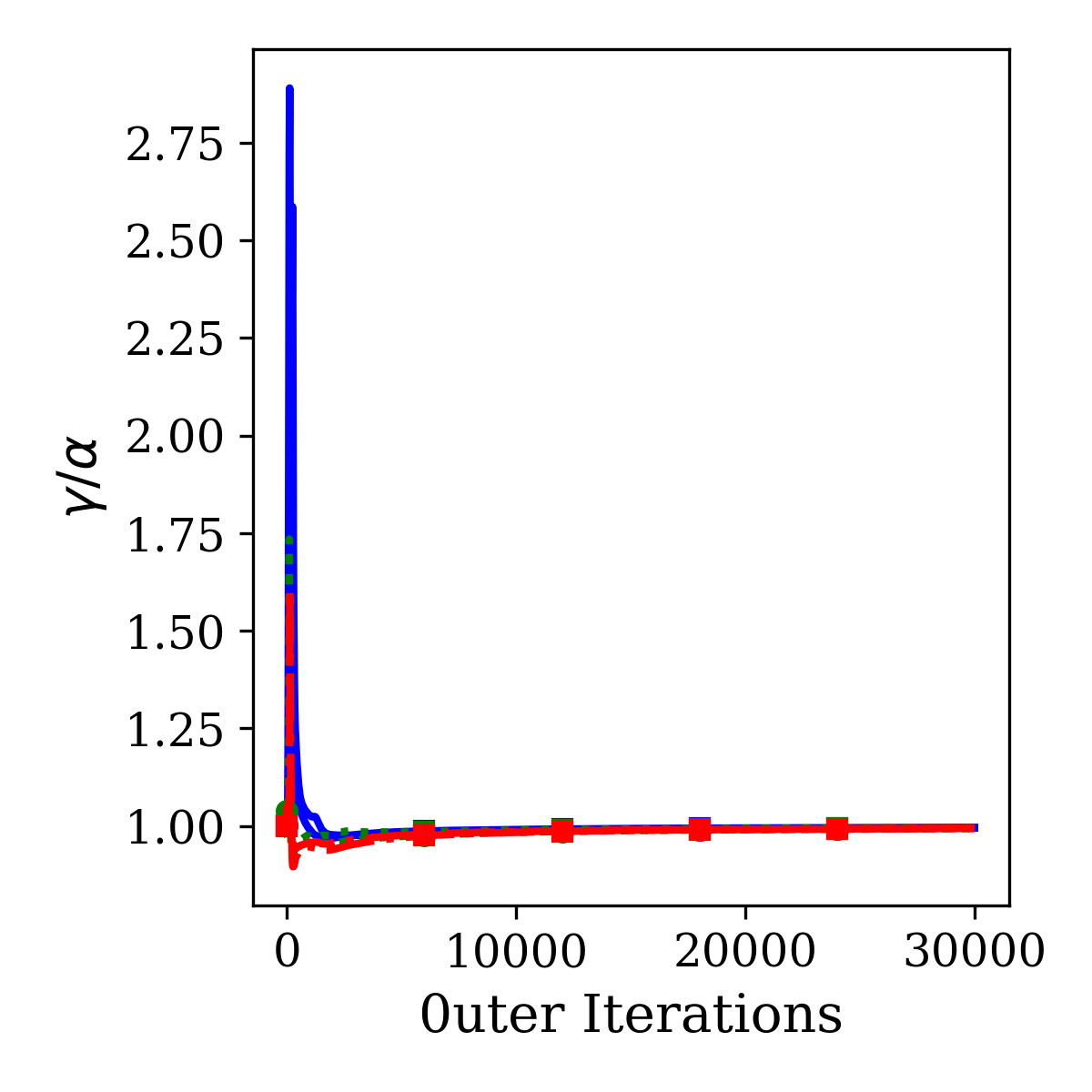}}\quad

    \caption{Helmholtz equation with a real-valued, high-wavenumber solution: Evolution of (a) PDE constraints, (b) boundary condition constraints, (c) objective functions, (d) relative $\mathit{l^2}$ norms, (e) the ratio $\beta / \alpha$, (e) the ratio $\gamma / \alpha$ for six specific subdomains.
}
\label{fig:helmholtz_l5_update}
\end{figure}

To illustrate the performance of the method, we select six specific subdomains based on the symmetry of the solution, and track the evolution of the loss terms, the relative \(\mathit{l^2}\) norms, and the interface parameter ratios, as shown in Figure~\ref{fig:helmholtz_l5_update}. 
At the early stages of training, most subdomains, except the central one (2,2), fall into local minima with low values for PDE, boundary constraints and objective functions, as shown in Figures~\ref{fig:helmholtz_l5_update}(a-c). This behavior arises from the challenging problem setup: the source term, with its narrow width and high magnitude, is concentrated in the central subdomain, while the surrounding subdomains have near-zero source terms.
The high wave number further complicates the reduction of the PDE constraint in the central subdomain. 
As the PDE constraint in this region decreases to around \(10^{-3}\), its solution improves, gradually propagating information to neighboring subdomains. This leads to an increase in the objective functions (interface losses) in adjacent subdomains, pulling them out of their local minima, as reflected in the noticeable rise in constraint values.
The interplay between satisfying the boundary conditions and improving the physics constraints drives the reduction of the relative \(\mathit{l^2}\) norms, as seen in Figure~\ref{fig:helmholtz_l5_update}(d). This process continues over 30,000 iterations towards convergence of $\mathcal{E}_r$.

Examining the convergence characteristics of the interface parameter ratios in Fig.~\ref{fig:helmholtz_l5_update}(e-f), it is evident that the parameters for the Neumann and tangential derivative continuity operators, \(\beta\) and \(\gamma\), have a significant impact on information communication in the early stages. After that, both ratios, \(\dfrac{\beta}{\alpha}\) and \(\dfrac{\gamma}{\alpha}\), converge to values around 1 after 6,000 outer iterations.

\subsection{Inverse Problems}
We now apply our PECANN framework with DDM to learn the solution of inverse PDE problems. Specifically, the objective is to infer the unknown, spatially varying thermal conductivity in 2D steady-state heat conduction across multilayered and functionally graded materials.
Both materials generally exhibit non-uniform microstructures, resulting in anisotropic and heterogeneous macroscopic properties \cite{Somasundharam2018invmaterials}. 
In the field of heat transfer, accurately measuring or estimating the thermal properties, such as thermal conductivity and specific heat capacity, of these materials is essential for design optimization of thermally controlled systems \cite{RAUSCH2013610, DALESSANDRO2023123666}.

The governing equation for the underlying heat conduction problem is written as:
\begin{equation}
\begin{aligned}
    \frac{\partial u}{\partial x} \kappa(x,y) \frac{\partial u}{\partial x} + \frac{\partial u}{\partial y} \kappa(x,y) \frac{\partial u}{\partial y} & = f(x,y) && \text{in} \quad \Omega
\end{aligned}
\label{eq:conduction_equation}
\end{equation}
where $u$ is the temperature field and $\kappa(x,y)$ is the space-dependent thermal conductivity that needs to be inferred. Dirichlet boundary conditions are applied.
It is crucial to highlight that the Neumann operator is the continuity of heat fluxes across the interfaces, expressed as \(\kappa(x,y) \frac{\partial u}{\partial \boldsymbol{n}}\), while tangential derivative continuity operator is still the tangential derivative of $u$.

\subsubsection{Inferring thermal conductivity in multilayered materials}
Multilayered materials, composed of layers with varying properties, are widely used in engineering applications such as integrated circuits \cite{kim2021extremely}, radiation shielding \cite{WAS2014355}, and environmental protection \cite{GRZEBIENIARZ2023101033} due to their ability to offer customized material characteristics. These materials typically exhibit significant discontinuities in thermal conductivity, $\kappa$, at the interfaces between layers, complicating the direct application of governing differential equations at interface points. Consequently, random sampling across the entire domain for training a single neural network model becomes impractical. Domain decomposition methods offer an effective solution by aligning each subdomain with an individual layer, enabling a localized approach to manage discontinuities. This method ensures accurate and efficient management of layer-specific properties and extends beyond merely accelerating the training process. 

We consider a steady-state 2D heat conduction problem within a double-layered material in the domain $\Omega = \{(x,y) \mid 0 \leq x \leq 1, 0 \leq y \leq 1\}$, which is divided at $x = 0.5$. The exact solution is given by:
\begin{equation}
[u, \kappa] = \begin{cases}
        [88x^2(1-x)y, \frac{3}{22}] & \text{for } 0 \leq x \leq 0.5, \\
        [(50x-3)(1-x)y, 1] & \text{for } 0.5 < x \leq 1,
    \end{cases}
\end{equation}
While the solution $u$ is continuous at the interface $x = 0.5$, it is not differentiable due to a sharp variation across the interface. However, the heat flux, represented by $\kappa \frac{\partial u}{\partial \textbf{n}}$, remains continuous across the interface. This continuity of flux is a crucial physical requirement that must be adhered to in our modeling approach.

\begin{figure}
    \centering
    \includegraphics[scale=0.6]{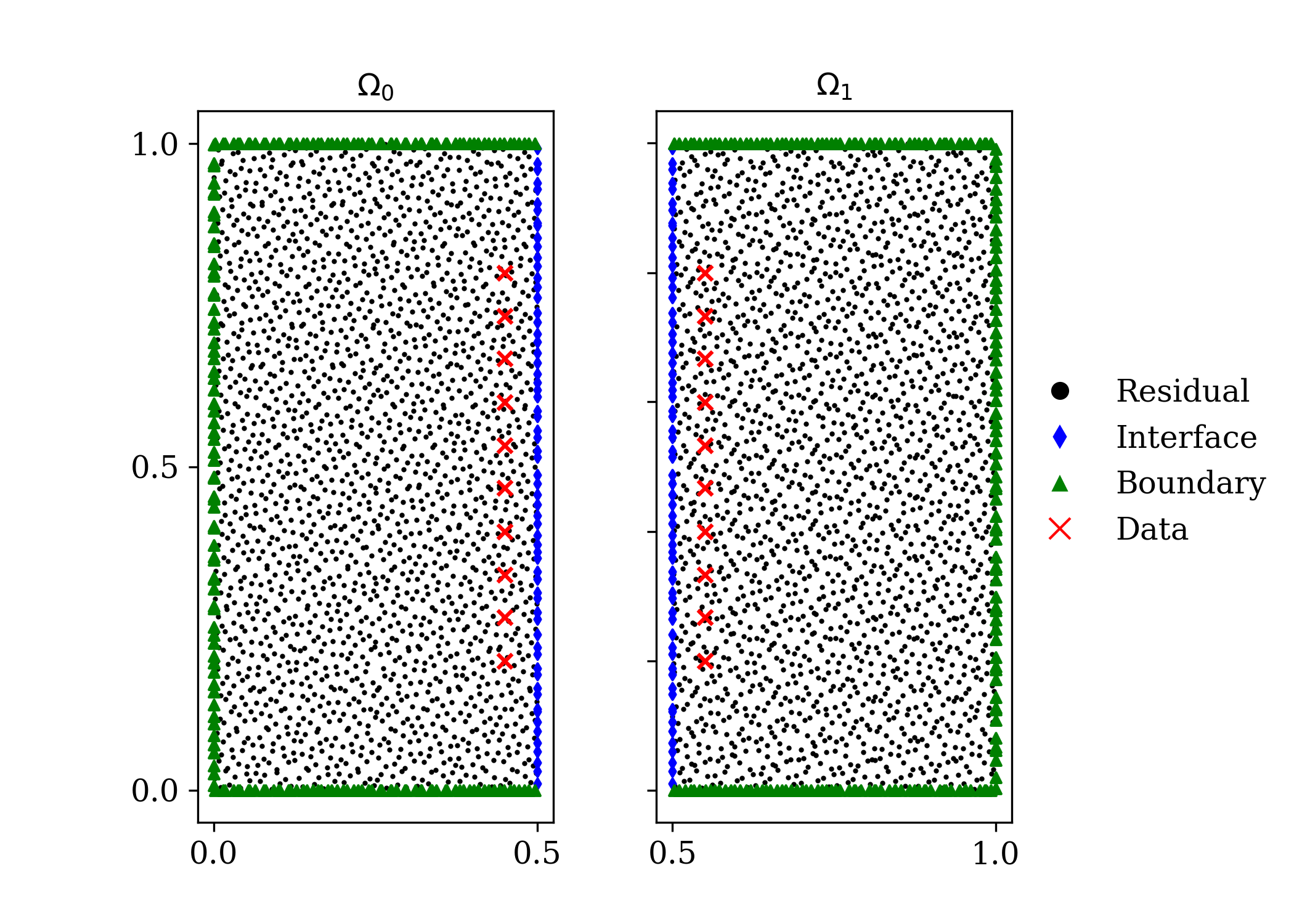}
    \caption{Computational domain for the inverse problem for multilayered materials: residual, interface, data and boundary points.}
    \label{fig:inv_multi_layers_points}
\end{figure}

The technical challenge lies in inferring two independent $\kappa$ values from sparse data. 
We generate synthetic measurement data at ten locations using pseudo sensors that are equidistantly positioned. 
These sensors are placed between \(y = 0.2\) and \(y = 0.8\), situated 0.05 units away from the interface.
Using a $2\times1$ domain decomposition, each subdomain is assigned to a distinct layer.
The global domain includes $64^2$ randomly distributed collocation points, comprising residual, boundary, and interface points, as shown in Figure~\ref{fig:inv_multi_layers_points}. For evaluation, we employ a uniform $128 \times 128$ mesh. 
The same neural network architecture, hyperparameters, and optimizer settings from the forward Poisson's problem, as shown in Figure~\ref{fig:poisson_simple_solution}, are used here. 
It is important to note that the additional data constraint shares the same default penalty scaling factor, $\eta_B = 1$, as the boundary condition constraint. The unknown conductivity of each layer is treated as a model parameter to be learned, i.e., $\kappa_k \in \theta_k$ for $k \in \{1,2\}$, with initial values randomly assigned between 0 and 1.

\begin{figure}[t]
\centering
    \subfloat[]{\includegraphics[scale=0.5]{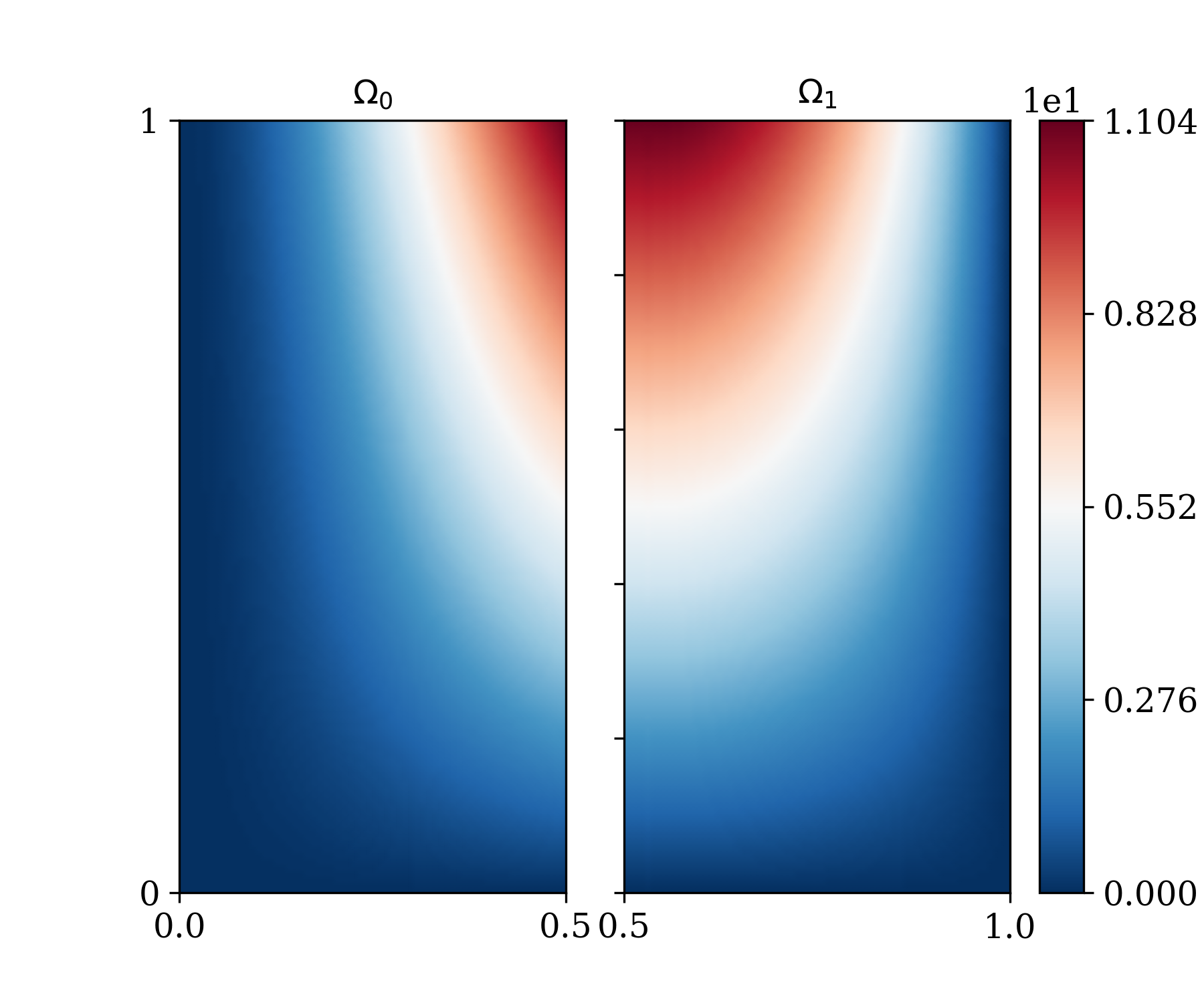}}\quad
    \subfloat[]{\includegraphics[scale=0.5]{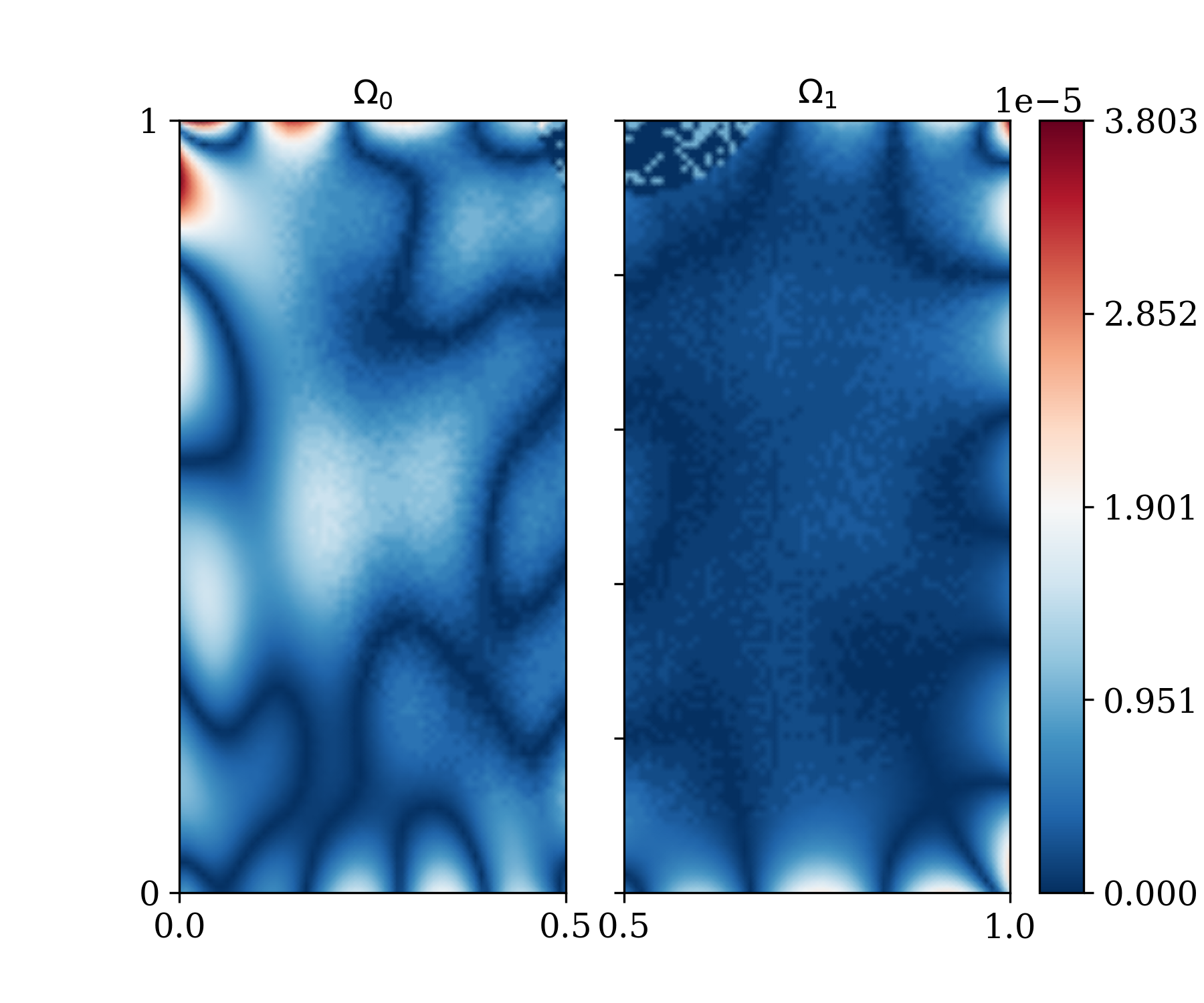}}\quad
    \subfloat[]{\includegraphics[scale=0.5]{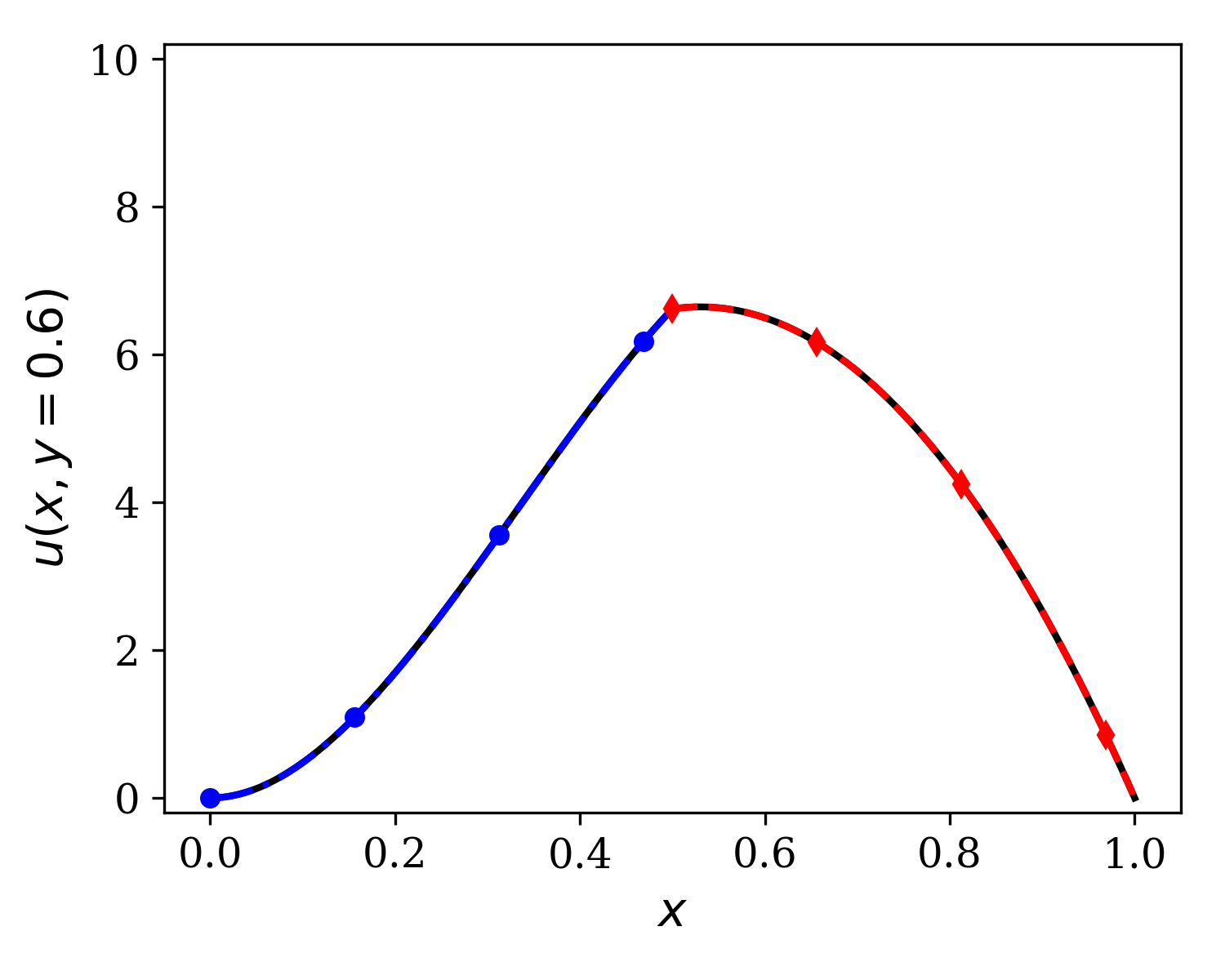}}\quad\quad\quad\quad
    \subfloat[]{\includegraphics[scale=0.5]{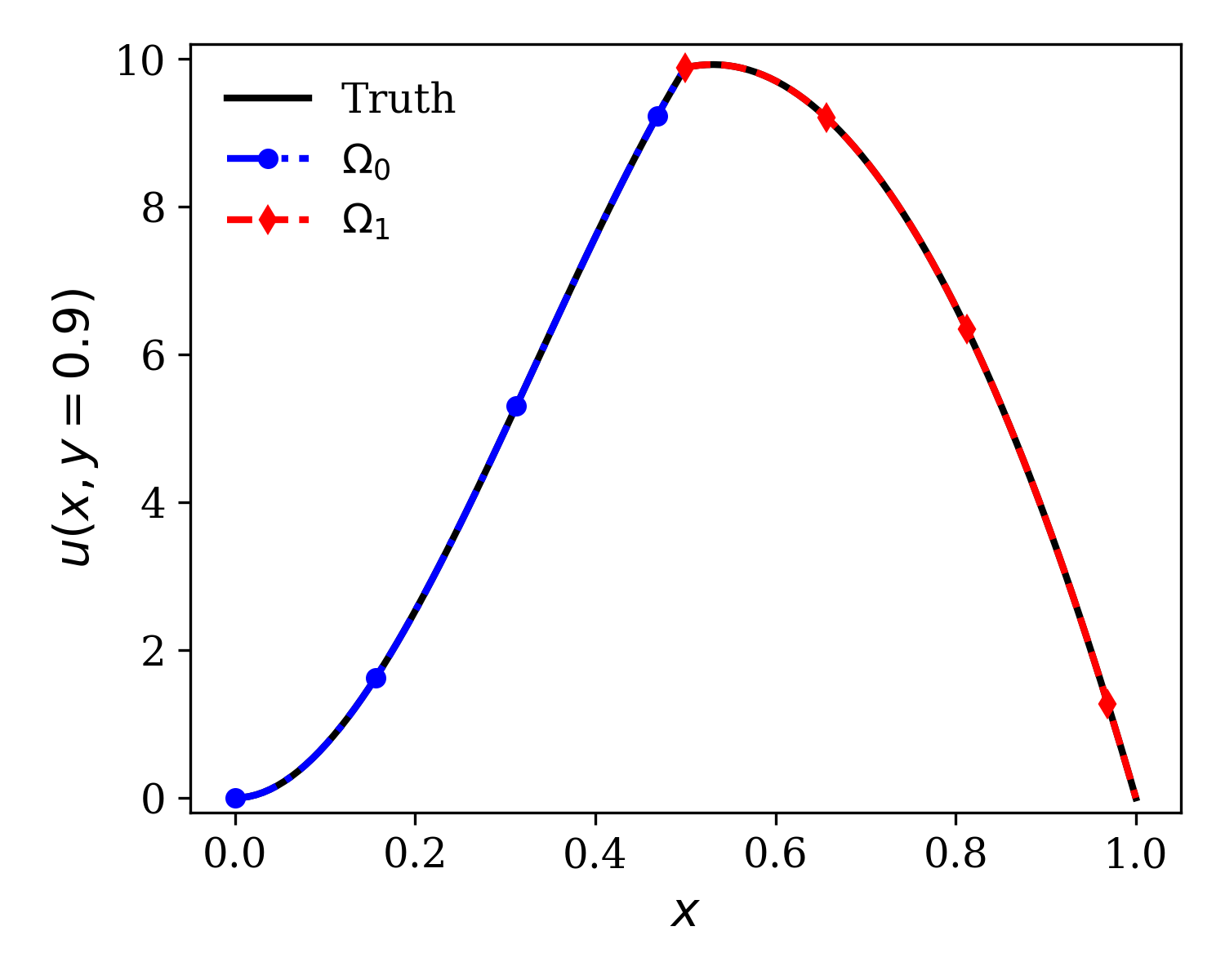}}\quad
    \caption{Inverse identification of spatially varying conductivity for multilayered materials: (a) Predicted temperature, (b) Absolute point-wise error in temperature prediction, (c) Predicted profile at $y=0.6$, (d) Predicted profile at $y=0.9$. The relative $l^2$ norm is $2.507\times10^{-5}$.}
    \label{fig:inv_multi_layers_solution}
\end{figure}

Figure~\ref{fig:inv_multi_layers_solution} presents the predicted temperature distribution and the absolute point-wise error, along with two temperature profiles from a single trial. 
Despite the relatively high range of temperature in Fig.~\ref{fig:inv_multi_layers_solution}(a), the maximum absolute error in Fig.~\ref{fig:inv_multi_layers_solution}(b) is reduced to around the order of $10^{-5}$, demonstrating remarkably high accuracy. 
In Figs.~\ref{fig:inv_multi_layers_solution}(c-d), the predicted temperature profiles at $y = 0.6$ and $y = 0.9$ closely match the exact solution for both layers, clearly capturing the expected non-differentiability at the layer interfaces.

\begin{figure}[!t]
\centering
    \subfloat[]{\includegraphics[scale=0.45]{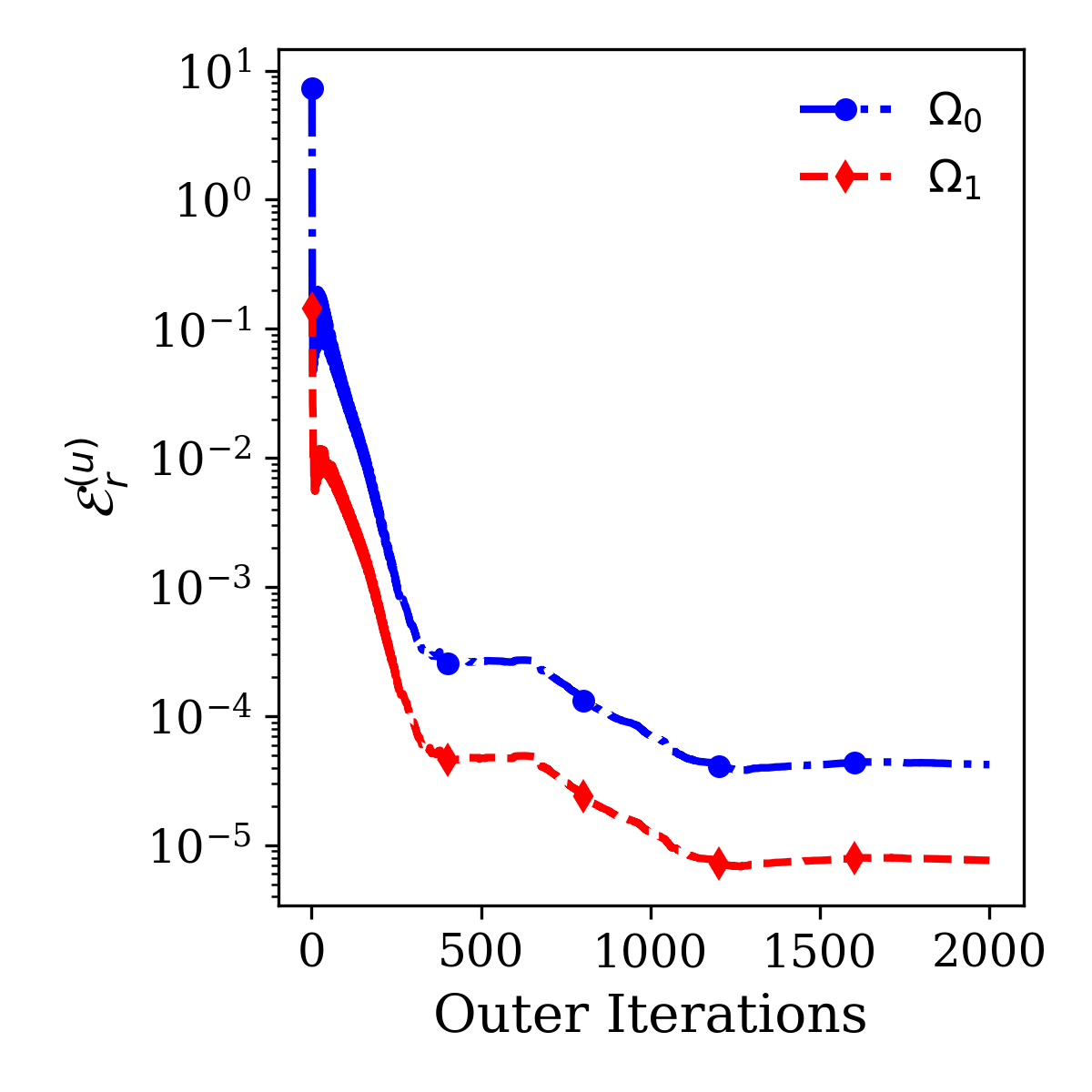}}\quad
    \subfloat[]{\includegraphics[scale=0.45]{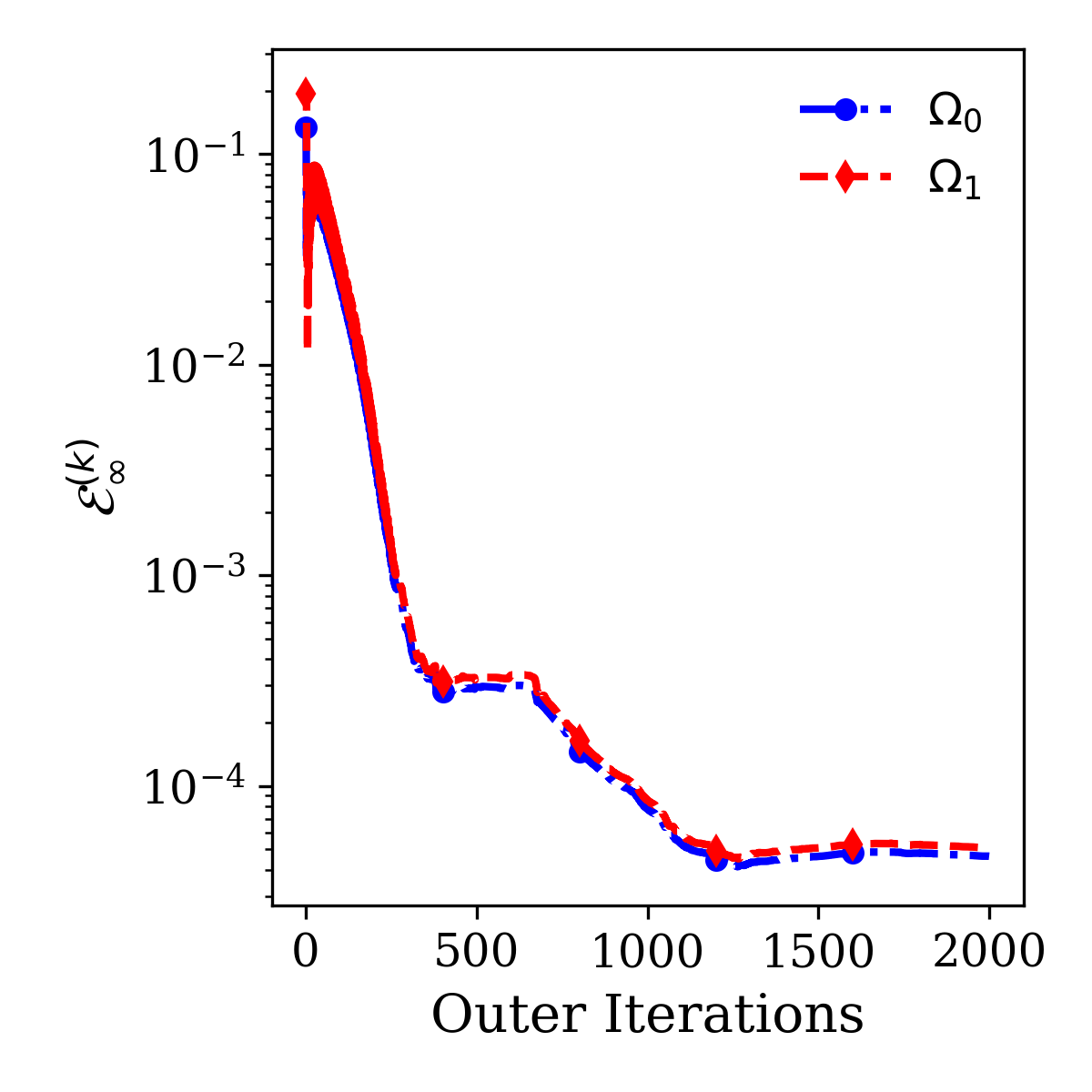}}\quad
    \subfloat[]{\includegraphics[scale=0.45]{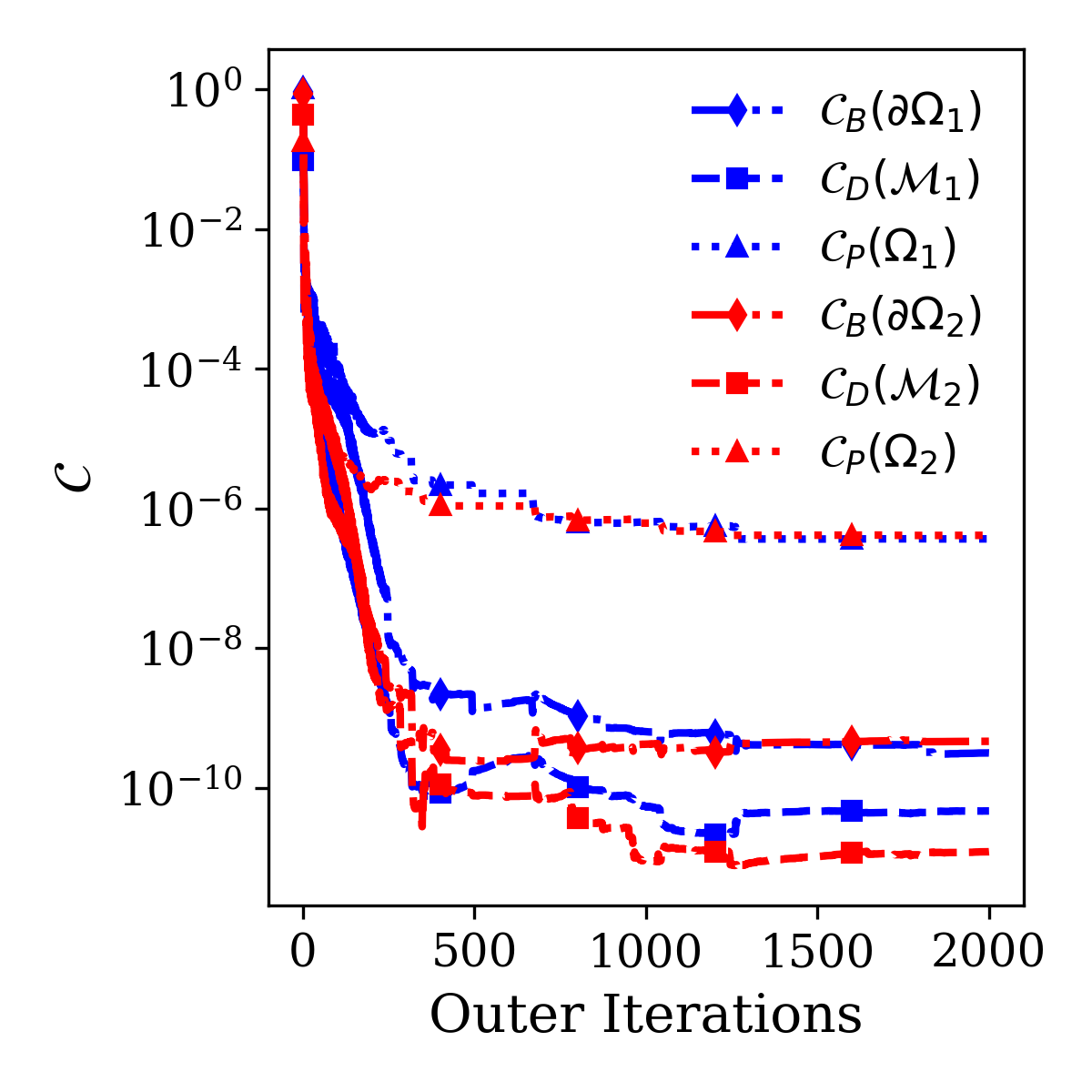}}\quad
    \subfloat[]{\includegraphics[scale=0.45]{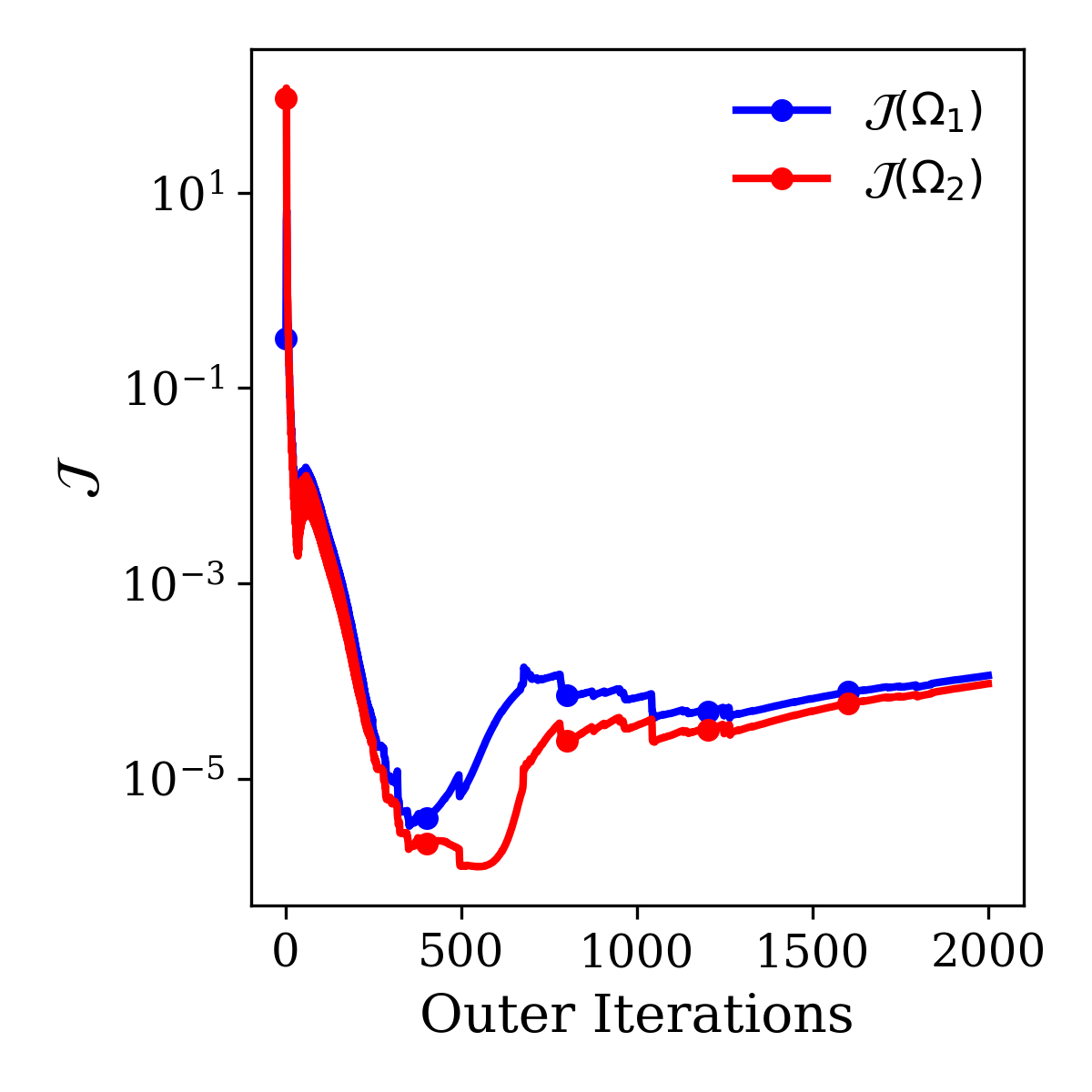}}\quad
    \subfloat[]{\includegraphics[scale=0.45]{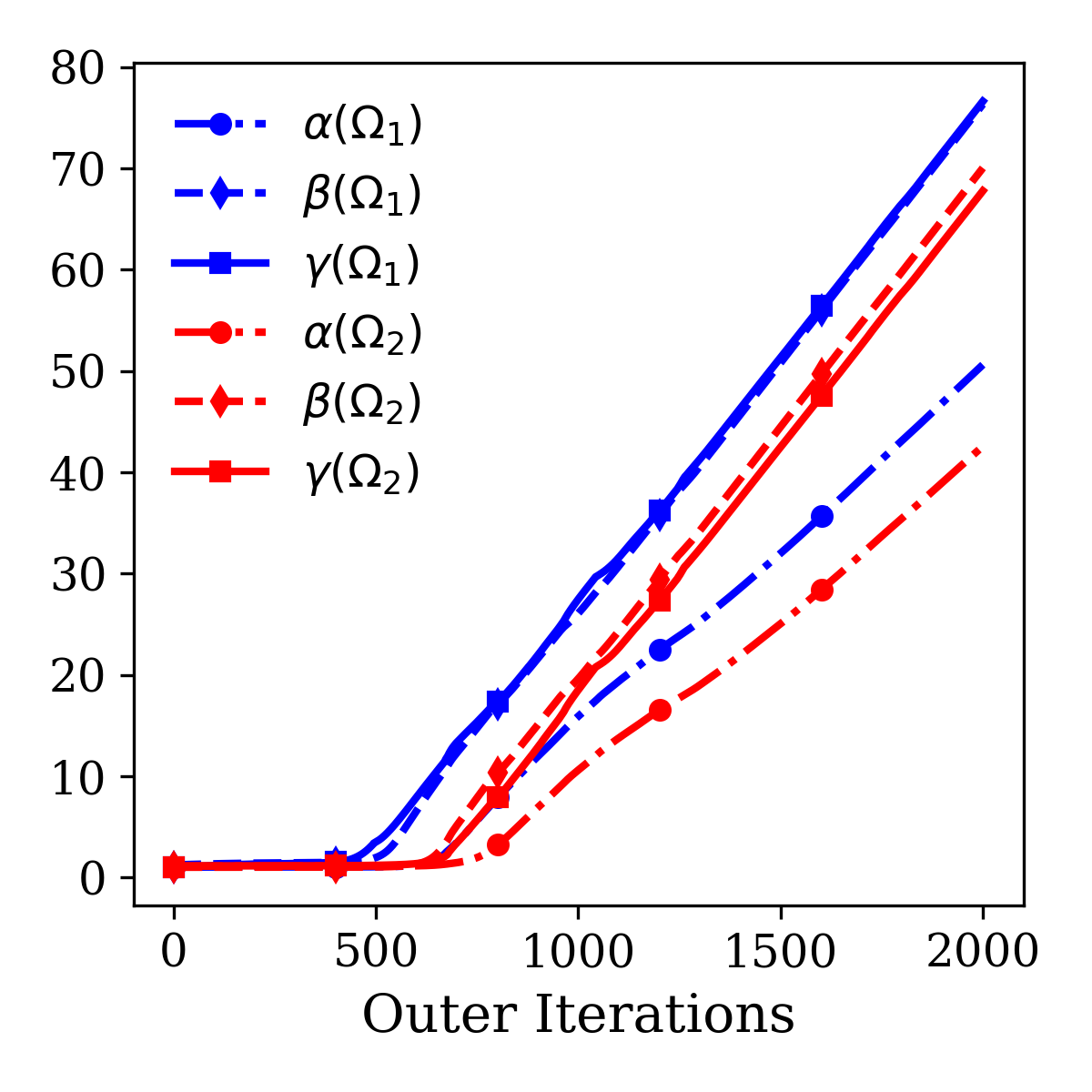}}\quad
    \subfloat[]{\includegraphics[scale=0.45]{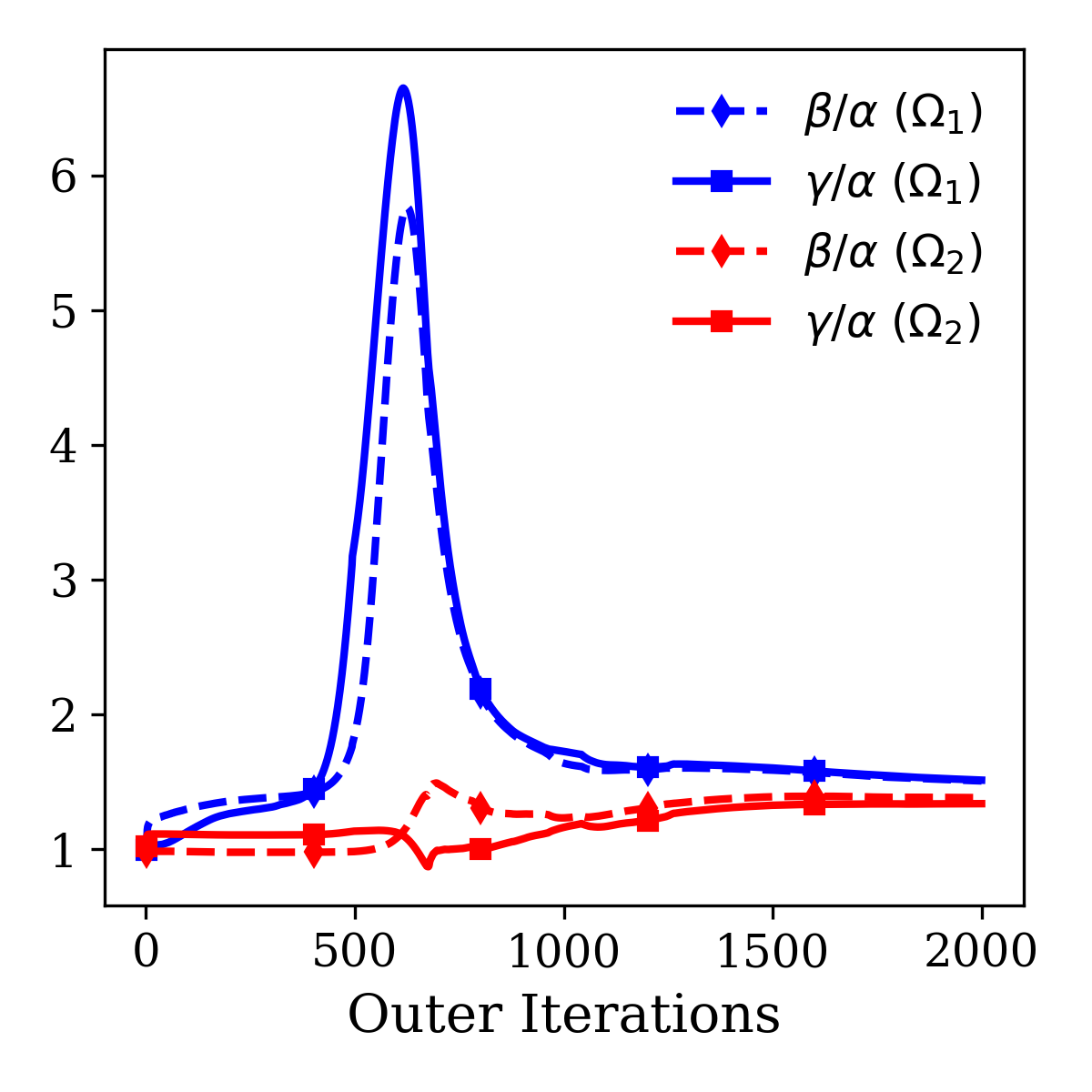}}\quad
    \caption{Inverse identification of spatially varying conductivity for multilayered materials: Evolution of (a) relative $\mathit{l^2}$ norms of temperature, (b) $\mathit{l^\infty}$ norms of conductivities (c) constraints, (d) objectives, (e) interface parameters, (f) ratios of interface parameters for both layers.}
    \label{fig:inv_multi_layers_update}
\end{figure}

Figure~\ref{fig:inv_multi_layers_update} depicts the evolution of two accuracy metrics, constraints, objective functions, and interface parameters, as well as their ratios for both layers. The relative $\mathit{l^2}$ norms of the temperature and the absolute errors of two predicted conductivities, $\mathit{l^\infty}$, are shown in Figs.~\ref{fig:inv_multi_layers_update}(a-b), respectively. 
In the beginning, both metrics exhibit rapid convergence, with the errors in $\mathcal{E}_r^{u}$ and $\mathcal{E}_\infty^{\kappa}$, separately decreasing by approximately five and three orders of magnitude. After a brief stabilization phase, further reductions occur around 700 outer iterations, coinciding with a slight decrease in the boundary condition and data constraints, in Fig.~\ref{fig:inv_multi_layers_update}(c), and a notable rise in the objective functions, in Fig.~\ref{fig:inv_multi_layers_update}(d).
The increase in objective function values is attributed to the growth of interface parameters, seen in Fig.~\ref{fig:inv_multi_layers_update}(e). 
As discussed in Section~\ref{sec:interface_loss}, increasing \(q^j\) helps maintains a strong gradient when the corresponding operator $\mathcal{O}_i^j$ converges to small values, thereby enhancing convergence rate. 
Eventually, the objective functions stabilize, and both accuracy metrics reach convergence after approximately 1200 outer iterations. Additionally, Fig.~\ref{fig:inv_multi_layers_update}(f) shows that the interface ratios also stabilize between 1 and 2.

\subsubsection{Inferring thermal conductivity in functionally graded materials}
Functionally graded materials (FGMs) are innovative materials characterized by gradual changes in composition, constituents, or microstructures along one or more spatial directions, leading to tailored variations in properties and functions for optimized performance \cite{ZHANG2019138209}.
In aerospace applications, FGMs outperform multilayered materials by offering better synergy and flexibility, which helps them handle extreme conditions like high wear, fatigue, and severe thermal stress more effectively \cite{Popoola16, SVETLIZKY2022142967}.
The heat transfer properties have been investigated via traditional numerical methods \cite{Olatunji2012FGMs, Xi2019FGMs} and machine learning \cite{sulaiman2024machine}.

\begin{figure}
\centering
    \subfloat[]{\includegraphics[scale=0.5]{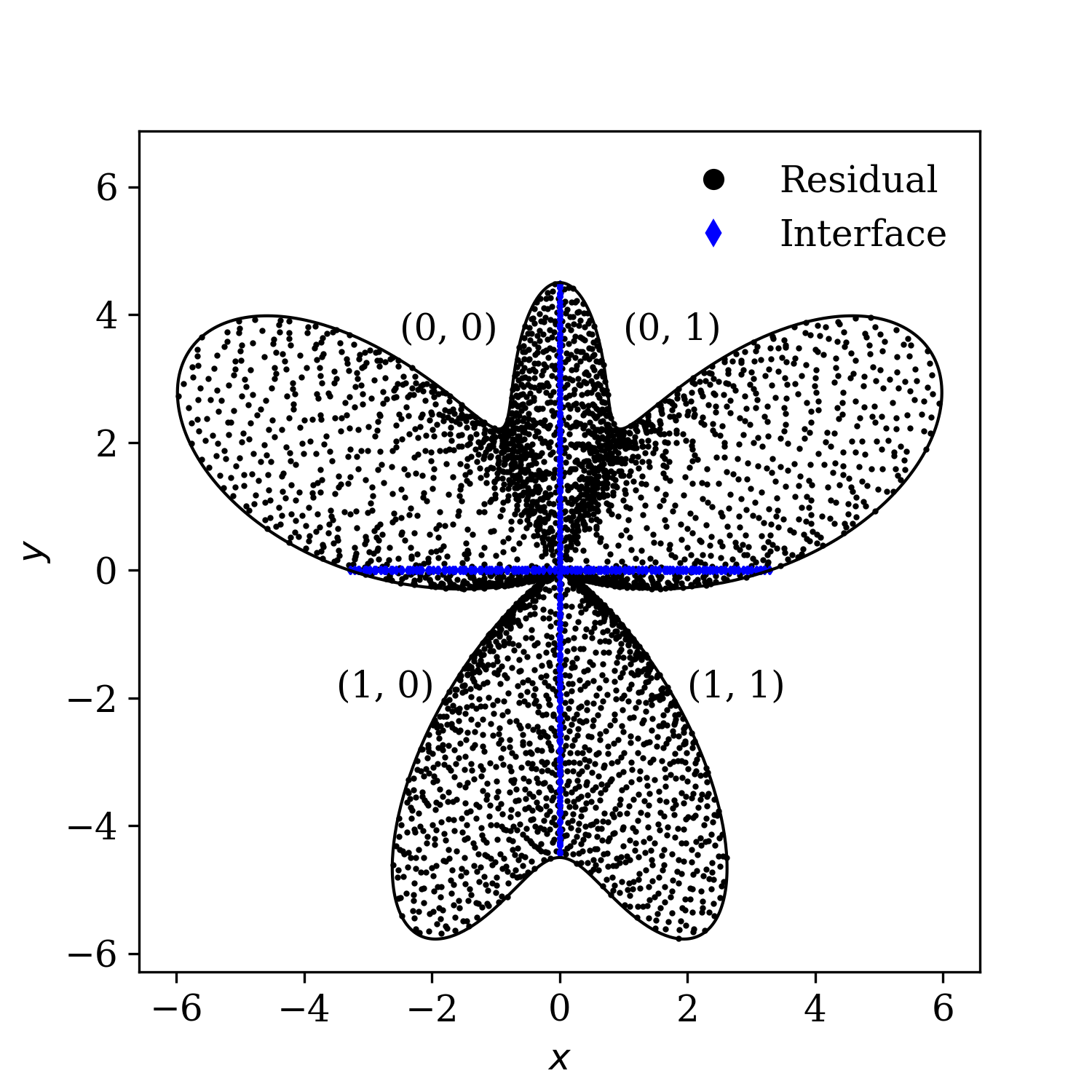}}\quad
    \subfloat[]{\includegraphics[scale=0.5]{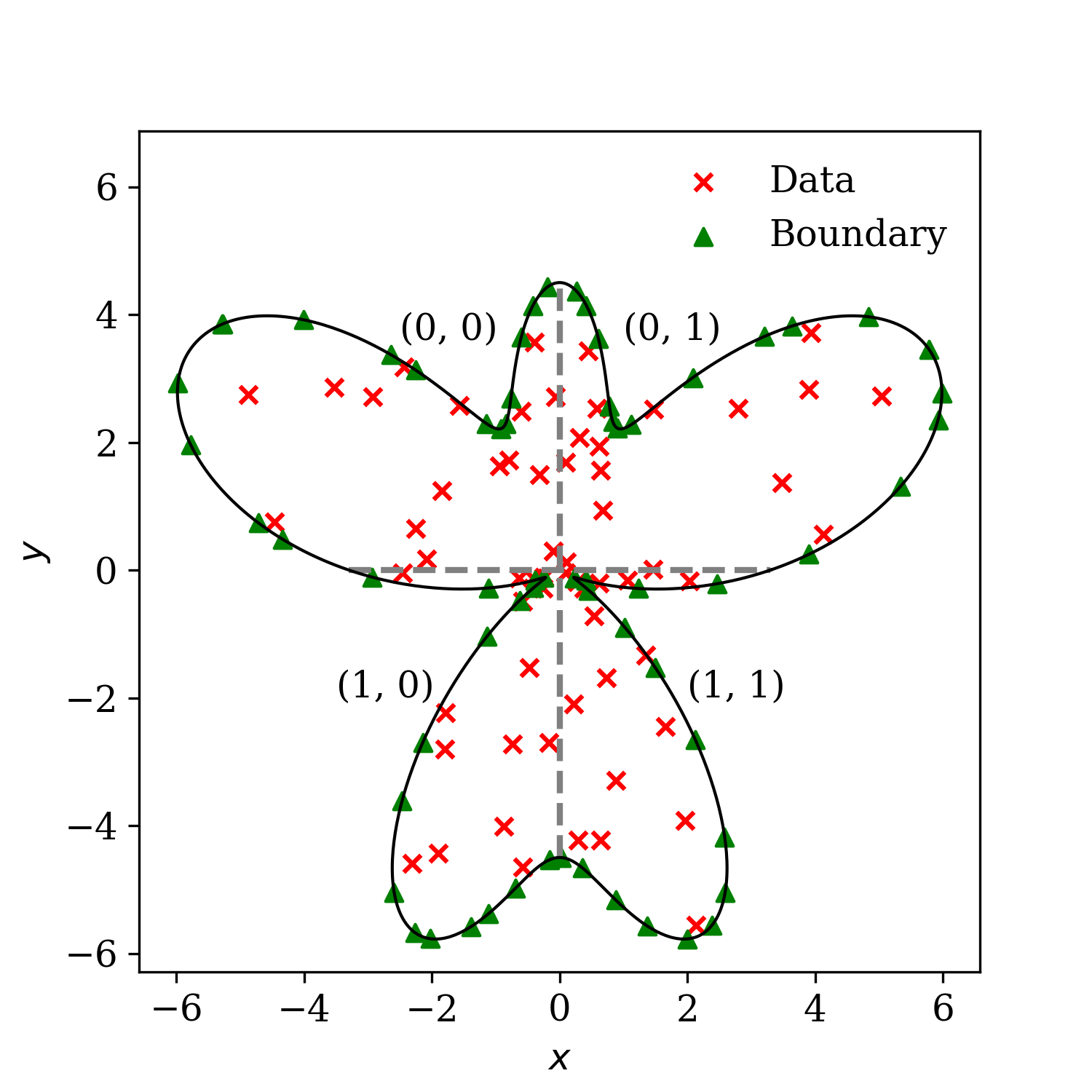}}\quad
    \caption{Inverse identification of spatially varying conductivity in functionally graded materials: (a) residual and interface points, (b) data and boundary points with $2\times2$ domain decomposition of a butterfly shape.}
    \label{fig:inverse_butterfly_points}
\end{figure}

Here, we adopt a similar problem used by \citet{shukla2021parallel_pinns}. Instead of the map of the United States of America, we consider a butterfly-shaped domain $\Omega$.
The geometry of the domain is defined as:
\begin{equation}
    \Omega = \{(x,y) | x = 3.3\rho(\theta)\cos(\theta), y=4.5\rho(\theta)\sin(\theta)\},
\end{equation}
where $\rho(\theta) = 1 + \cos(\theta) \sin(4\theta)$ for $0 \leq \theta \leq 2\pi$.
The predefined analytical expressions for temperature and conductivity are:
\begin{equation}
\begin{aligned}
    u(x, y) & = 20\exp(-0.1y), \\
    k(x, y) & = 20 + \exp(0.1y) \sin(0.5x).
\end{aligned}
\end{equation}
The challenge of this problem lies in inferring $k(x, y)$ from sparse measurements of temperature within the domain. The conductivity is also known at selected locations along the boundary.
With $2\times2$ domain decomposition, each of the four subdomains utilizes 1024 random residual points to represent the physics, with 64 points on each interface to facilitate information exchange between neighboring subdomains.
The collocation points are shown in Fig.~\ref{fig:inverse_butterfly_points}(a).
Additionally, each subdomain includes 16 synthetic temperature measurement points and 16 boundary points, where conductivity is known, in Fig.~\ref{fig:inverse_butterfly_points}(b).
The number of evaluation points used to assess the models is 65,536.

We use a straightforward neural network architecture with one hidden layer containing 20 neurons, and two output units for predicting both the temperature $u$ and the thermal conductivity $\kappa$. Model parameters and interface parameters are updated using Adam optimizers, with learning rates of $10^{-3}$ and $10^{-4}$, respectively. It is important to note that the Dirichlet operators at the interfaces now account for both $u$ and $\kappa$, each with its own Dirichlet parameter. Without this consideration, the prediction of $\kappa$ can exhibit unphysical gaps around the interfaces, even if the predictions within the subdomains are accurate.

\begin{figure}[t!]
\centering
    \subfloat[]{\includegraphics[scale=0.5]{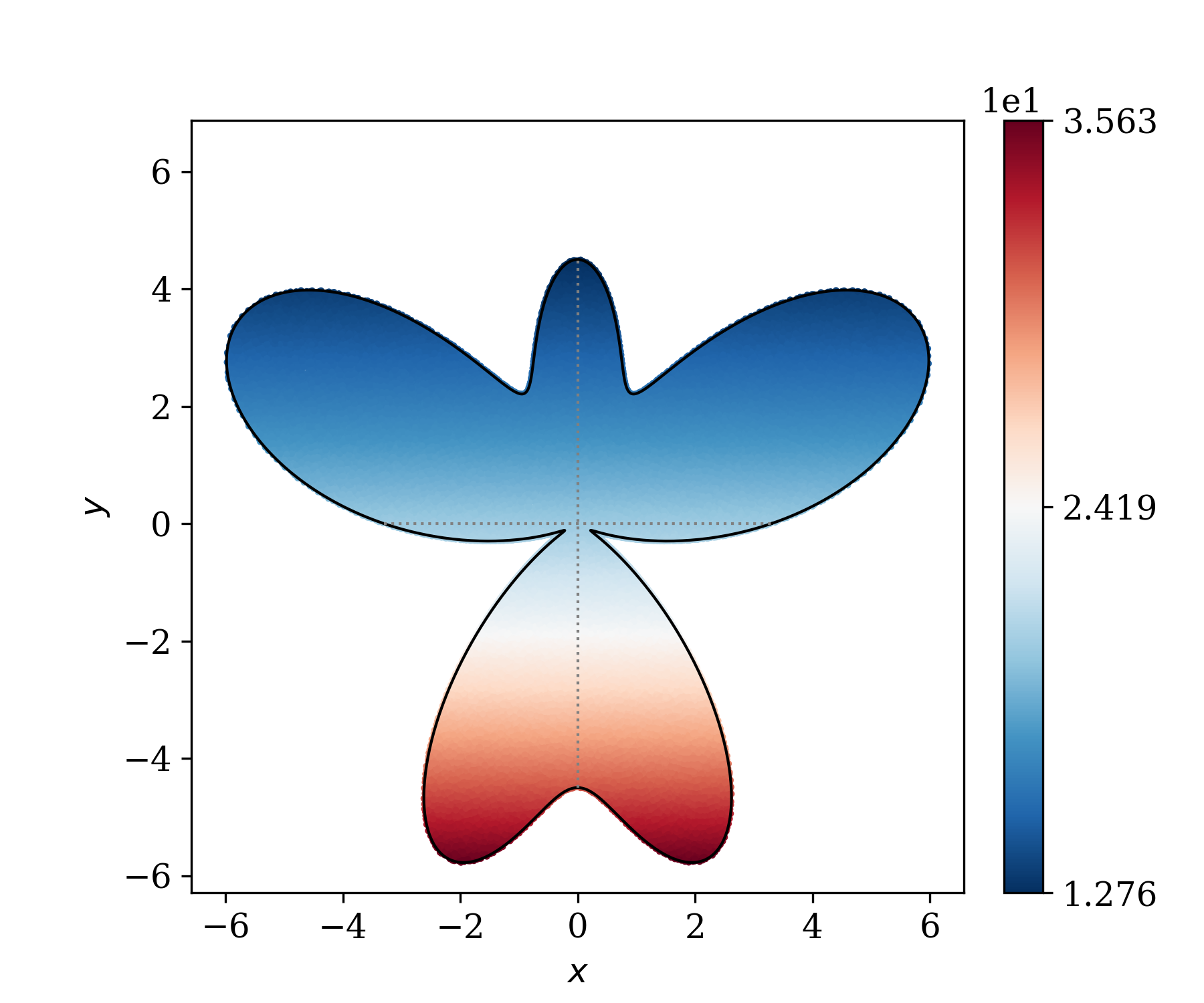}}\quad
    \subfloat[]{\includegraphics[scale=0.5]{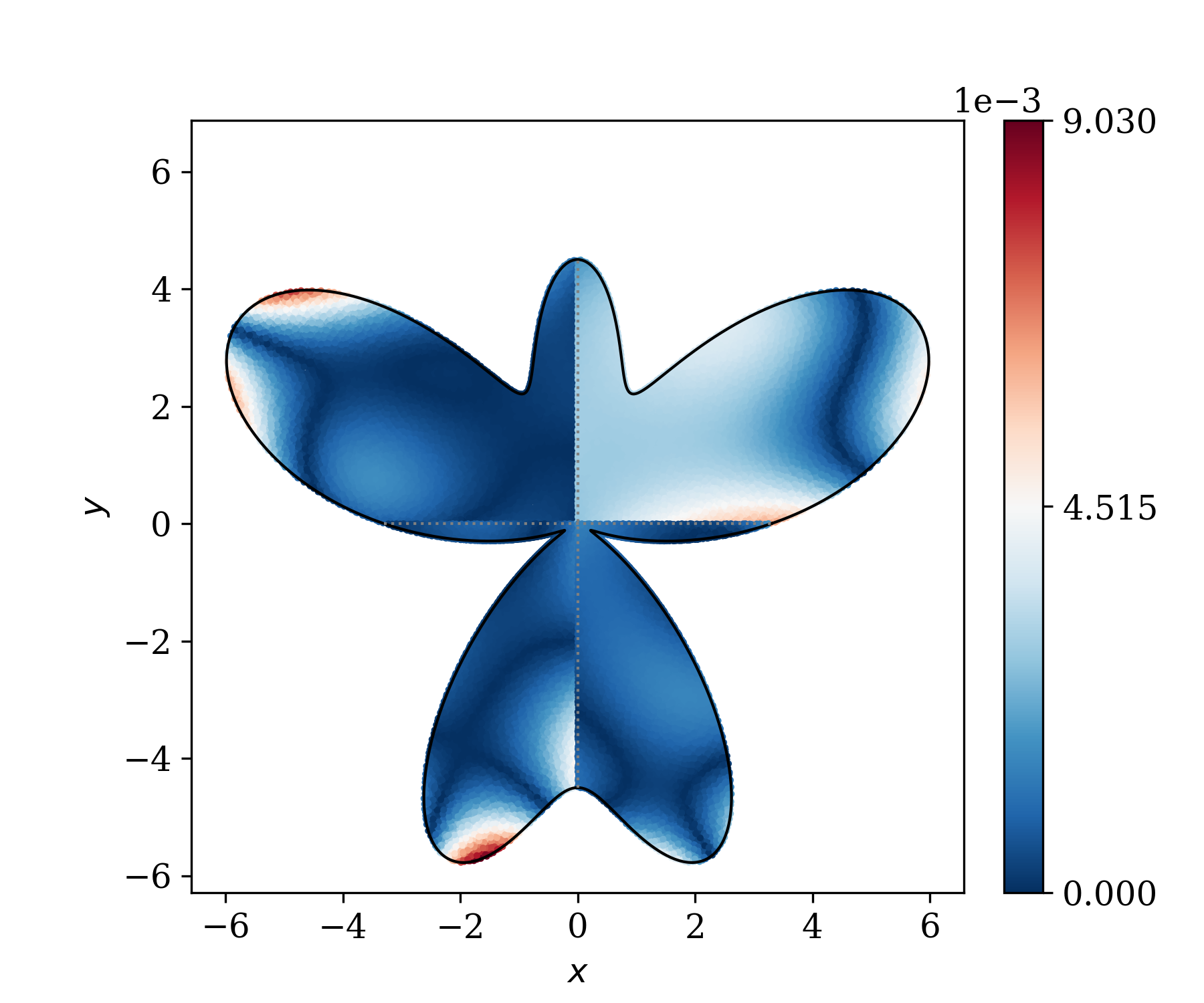}}\quad
    \subfloat[]{\includegraphics[scale=0.5]{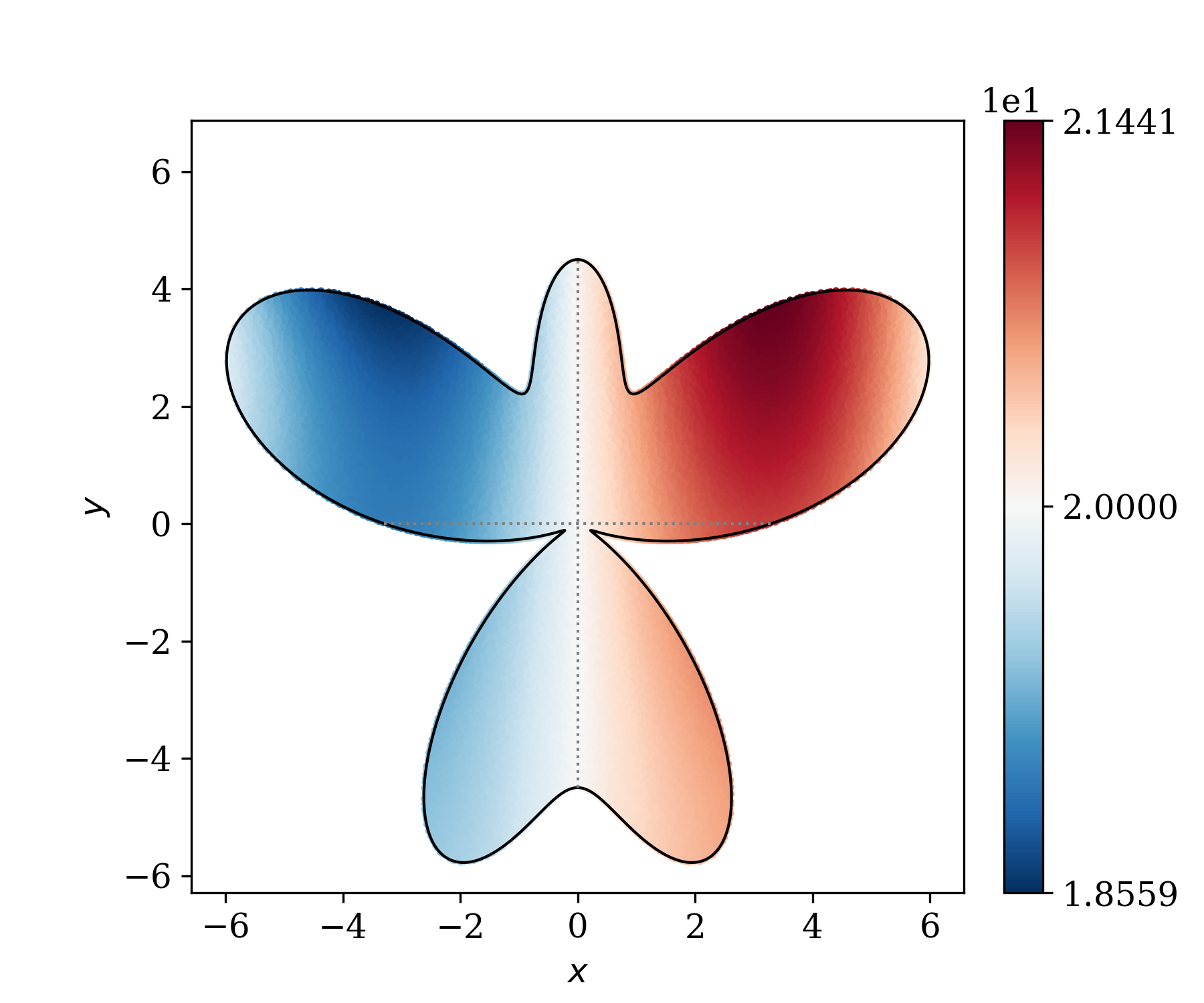}}\quad
    \subfloat[]{\includegraphics[scale=0.5]{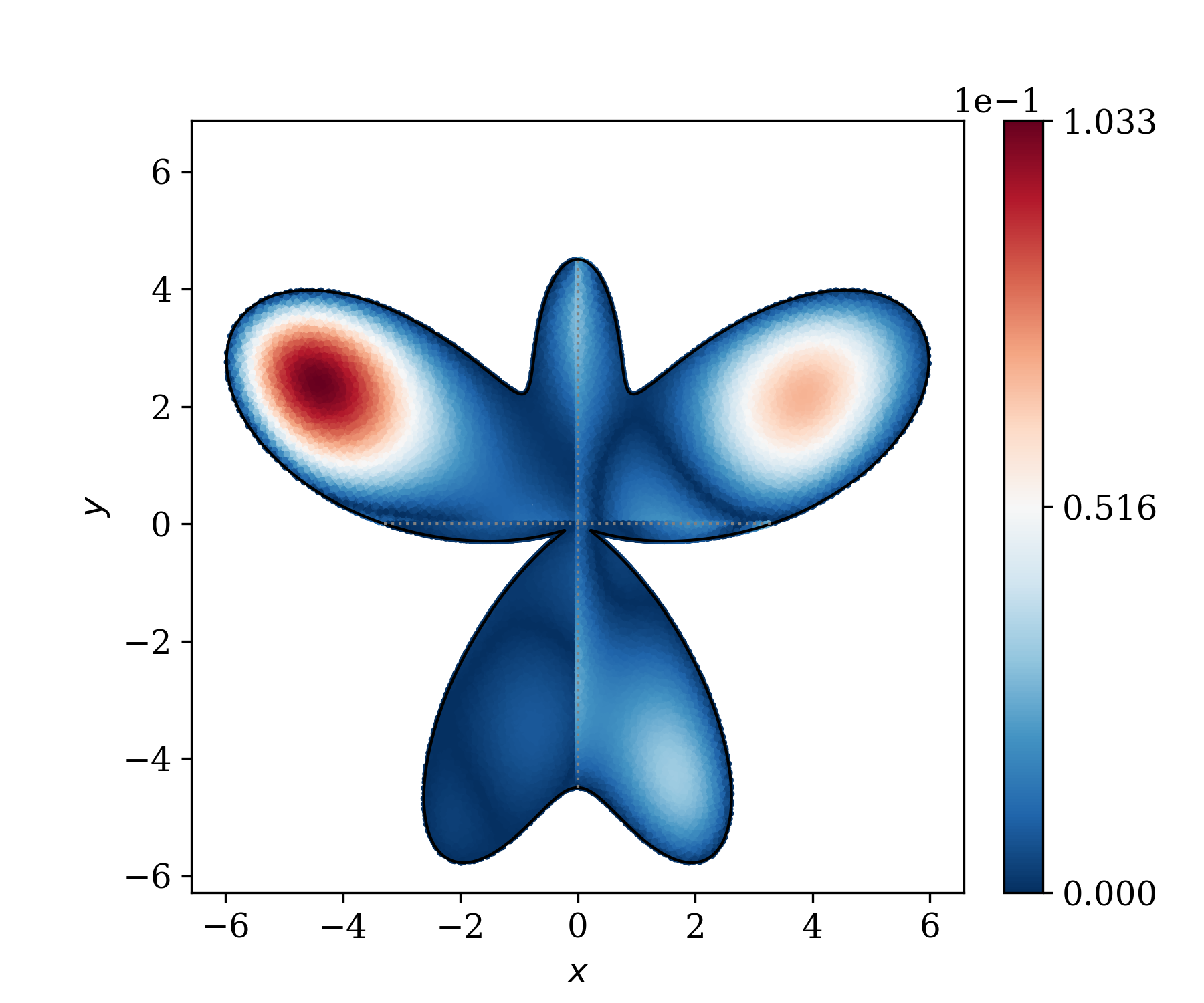}}\quad
    \caption{Inverse identification of spatially varying conductivity in functionally graded materials: (a) Predicted temperature, (b) absolute point-wise error in temperature prediction, (c) predicted thermal conductivity, (d) absolute point-wise error in conductivity prediction. The relative $l^2$ norms for $u$ and $\kappa$ are $8.703\times10^{-5}$ and $1.032\times10^{-3}$, respectively.}
    \label{fig:inverse_butterfly_solution}
\end{figure}

Figure~\ref{fig:inverse_butterfly_solution} displays the predicted distributions of temperature and thermal conductivity, along with their absolute errors. In Fig.~\ref{fig:inverse_butterfly_solution}(b), the temperature prediction errors are relatively uniform across the four subdomains, generally within the order of $10^{-3}$ and are primarily localized near the boundaries. In contrast, Fig.~\ref{fig:inverse_butterfly_solution}(d) reveals that the errors in conductivity predictions are mostly concentrated in the top subdomains, particularly in $(0,0)$, where the maximum error is $1.033 \times 10^{-1}$. Comparing these errors with the true ranges shown in Figs.~\ref{fig:inverse_butterfly_solution}(a) and~\ref{fig:inverse_butterfly_solution}(c), both predictions demonstrate high accuracy, with temperature predictions being more precise than those for conductivity.

\begin{figure}[h!]
\centering
    \subfloat[]{\includegraphics[scale=0.45]{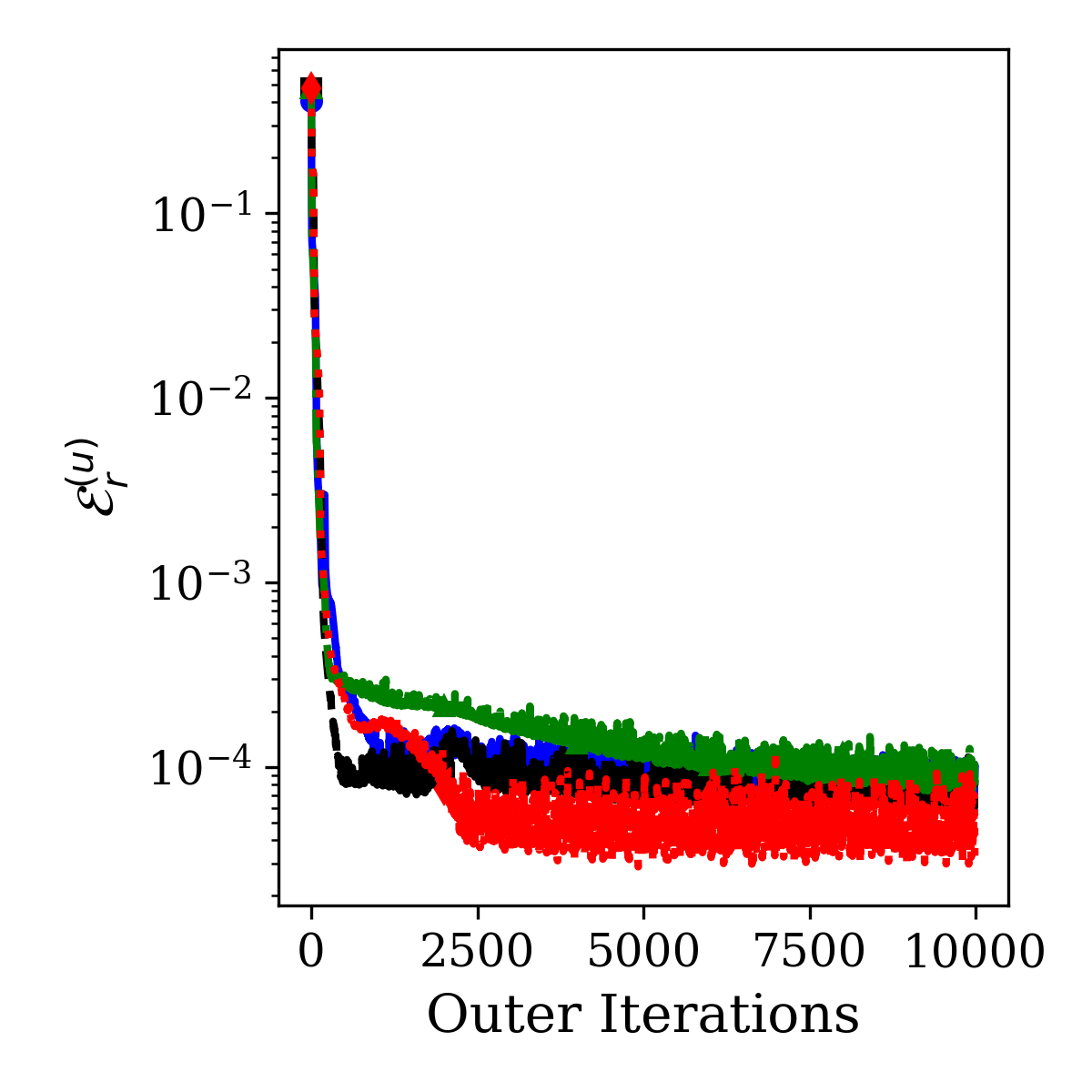}}\quad
    \subfloat[]{\includegraphics[scale=0.45]{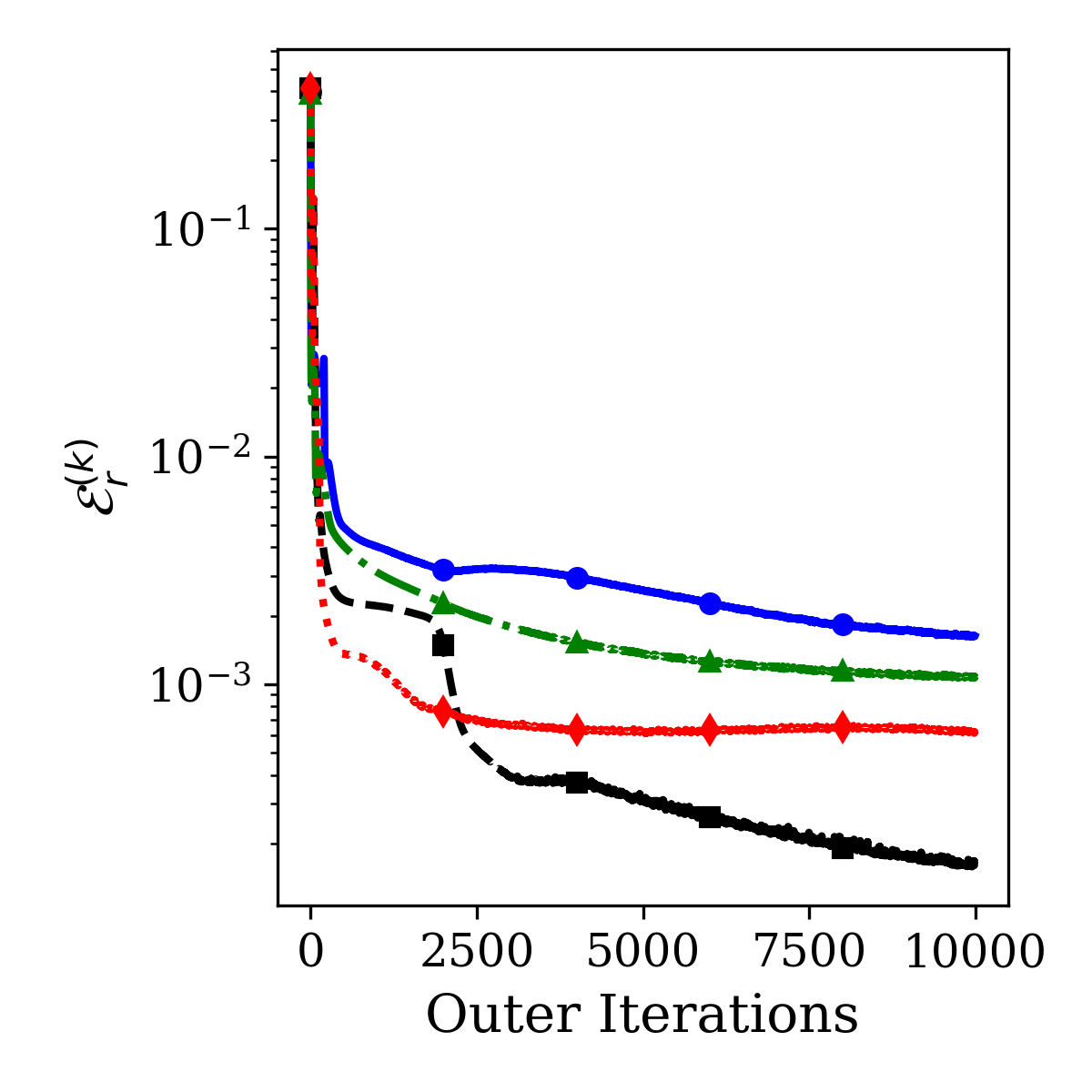}}\quad
    \subfloat[]{\includegraphics[scale=0.45]{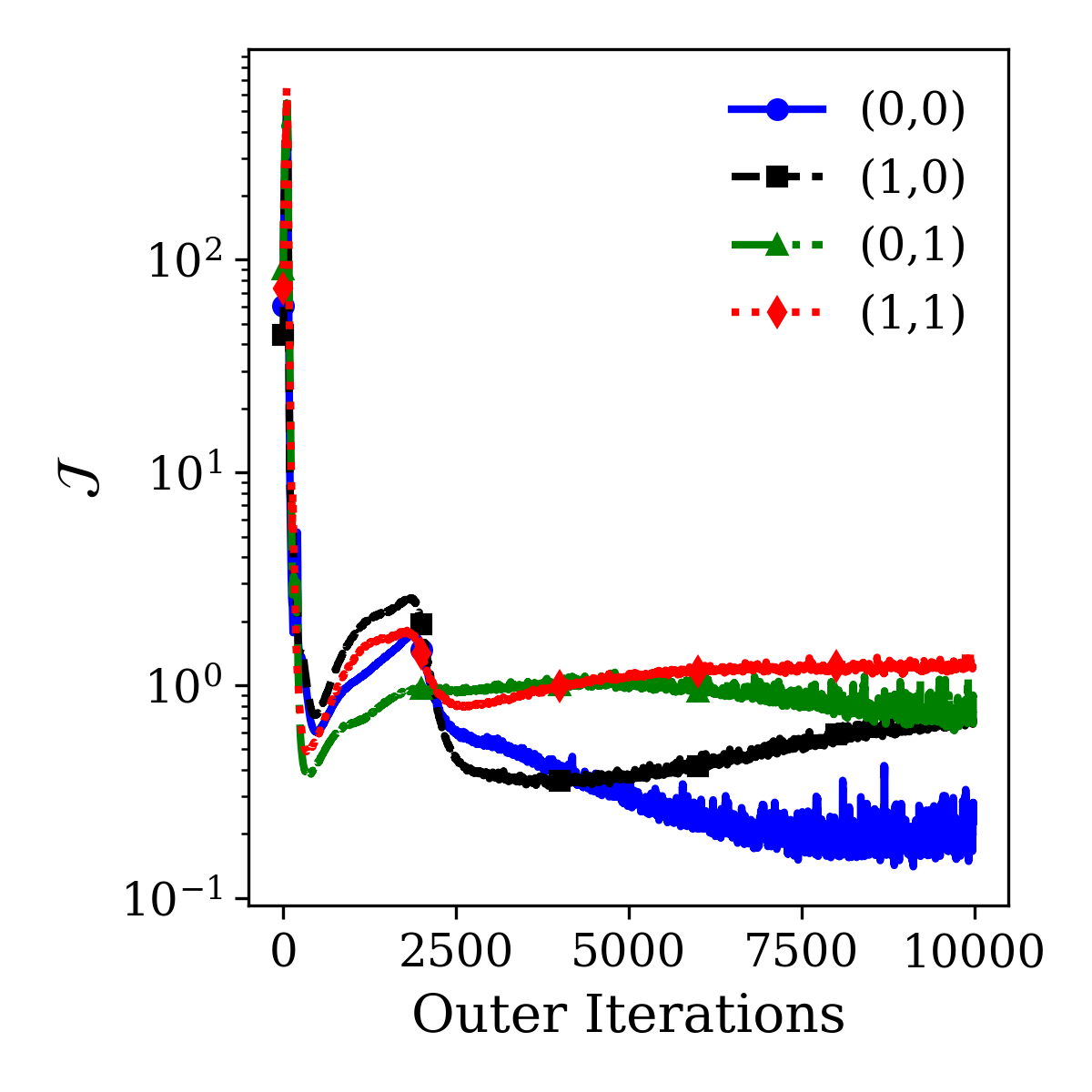}}\quad
    \subfloat[]{\includegraphics[scale=0.45]{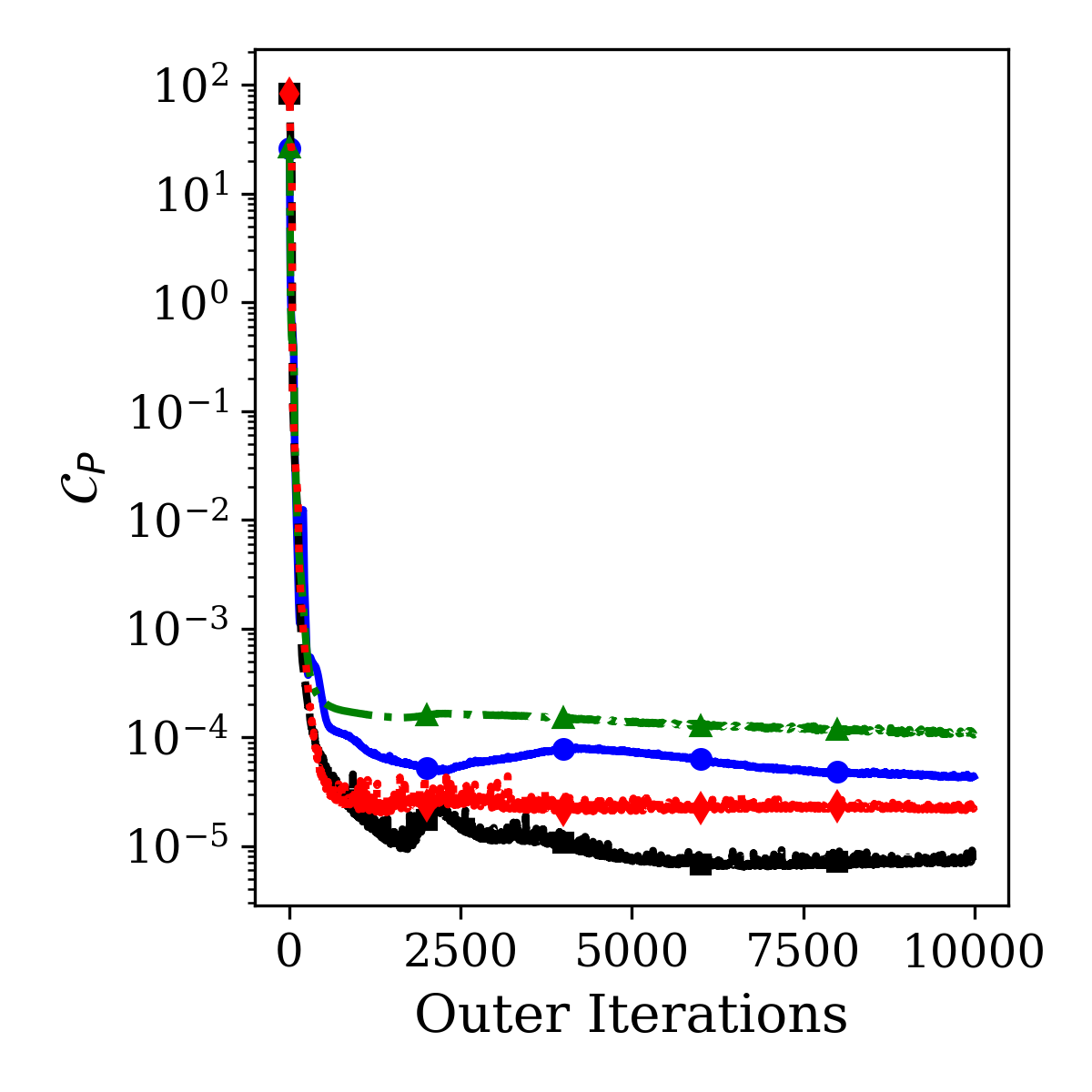}}\quad
    \subfloat[]{\includegraphics[scale=0.45]{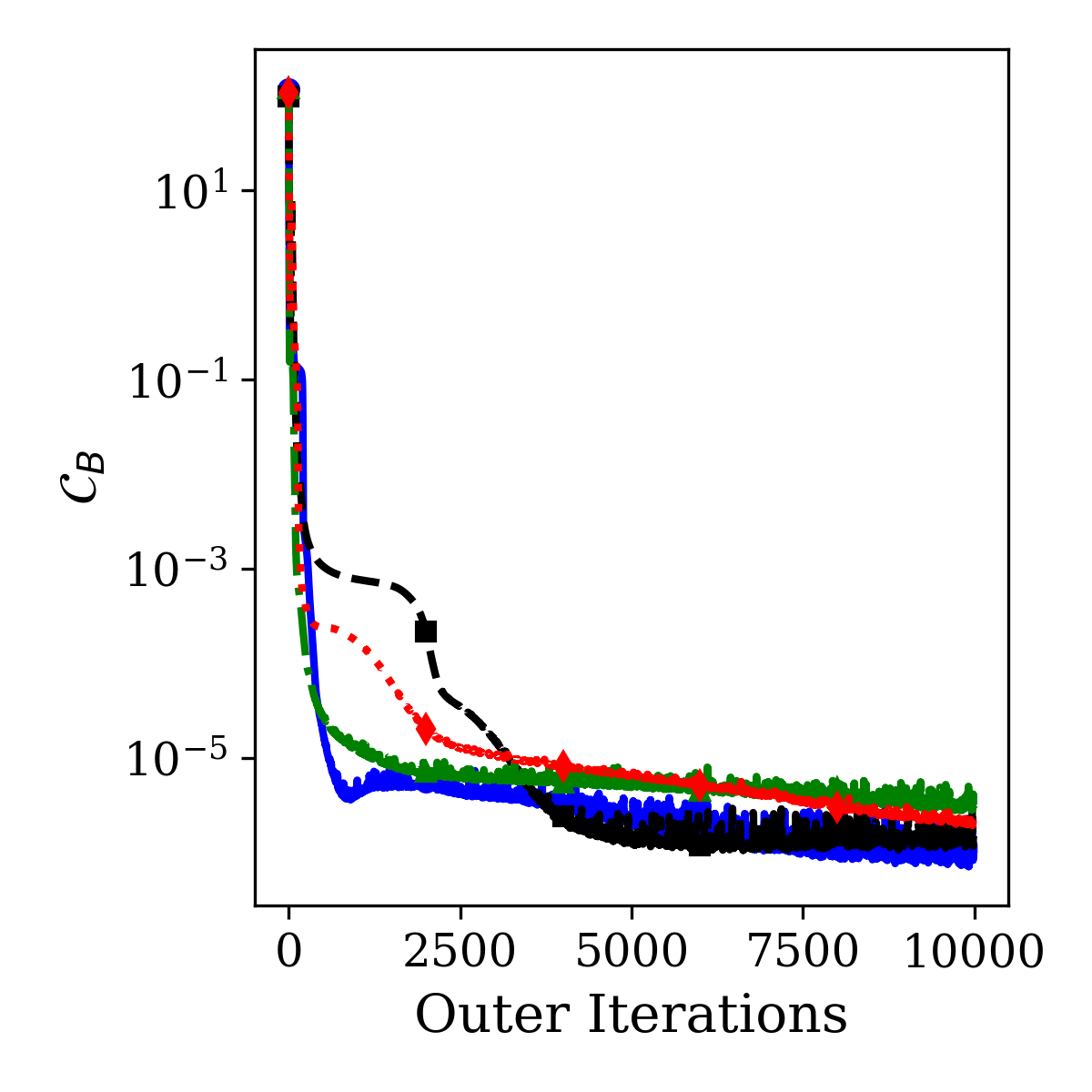}}\quad
    \subfloat[]{\includegraphics[scale=0.45]{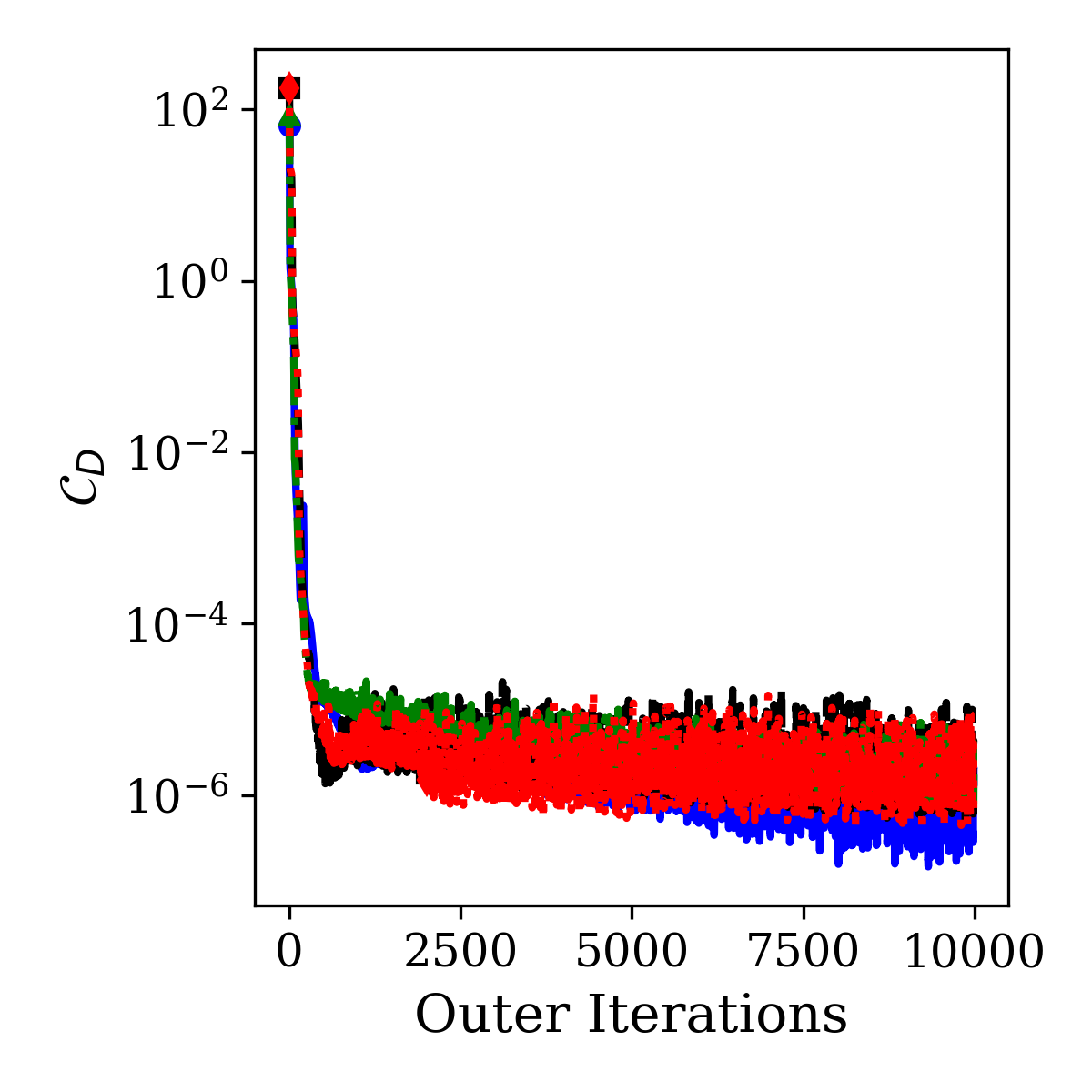}}\quad
    \subfloat[]{\includegraphics[scale=0.45]{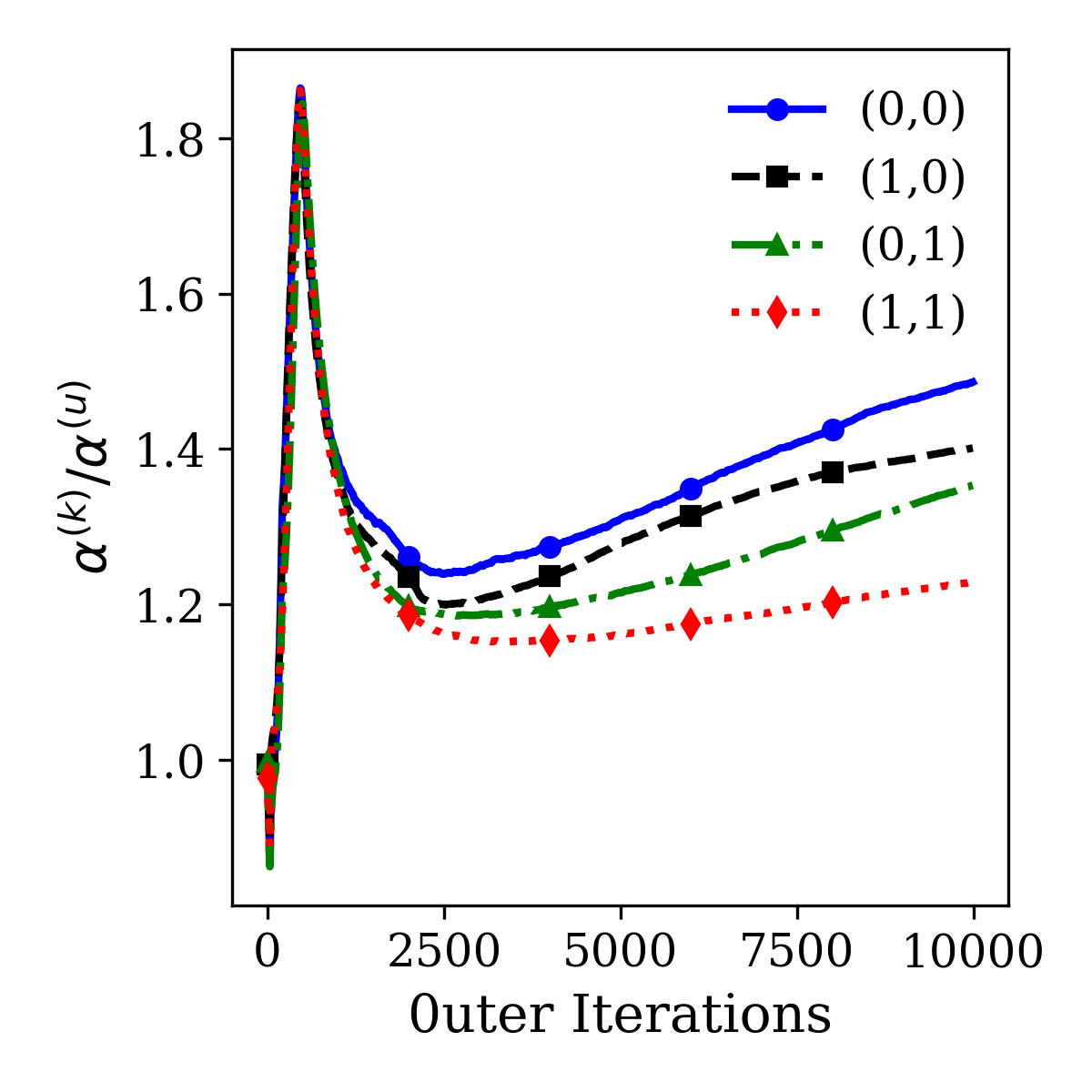}}\quad
    \subfloat[]{\includegraphics[scale=0.45]{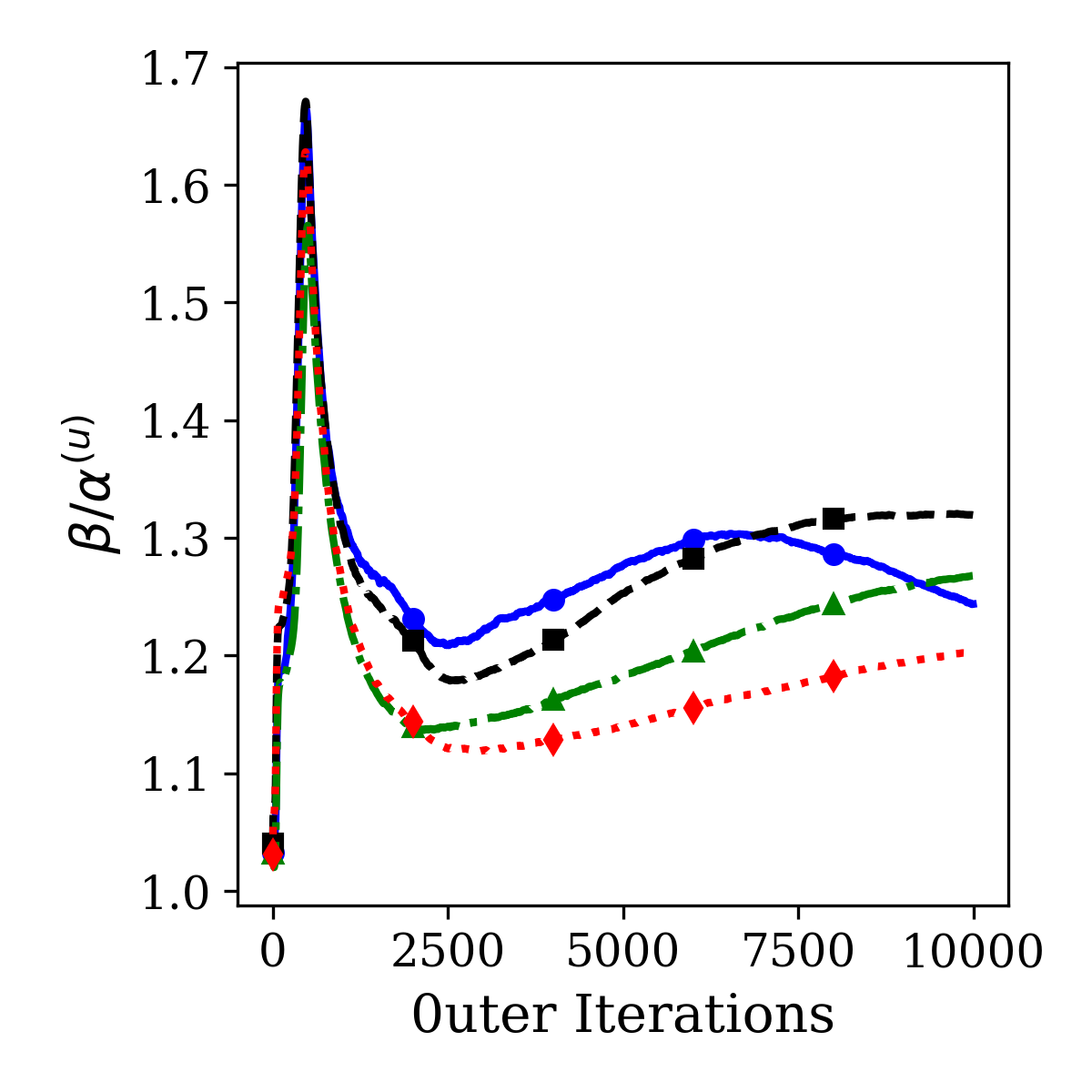}}\quad
    \subfloat[]{\includegraphics[scale=0.45]{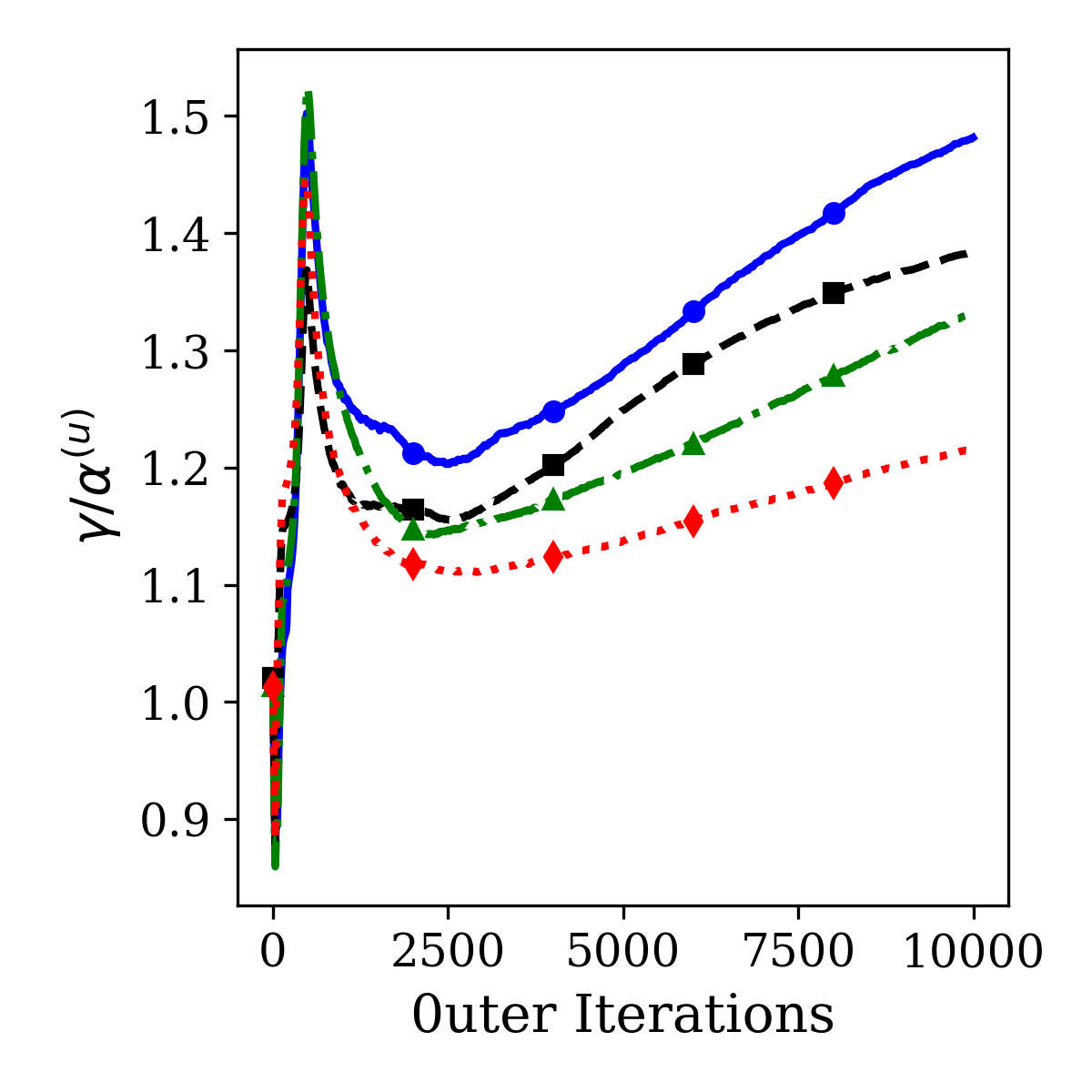}}\quad
    \caption{Inverse identification of spatially varying conductivity in functionally graded materials: Evolution of (a) relative $\mathit{l^2}$ norms for temperature, (b) relative $\mathit{l^2}$ norms for conductivity, (c) objective functions, (d) PDE constraints, (e) boundary condition constraints, (f) data constraints, (g) interface parameter ratio  $\alpha^{(k)}/\alpha^{(u)}$, (h) $\beta/\alpha^{(u)}$, (i) $\gamma/\alpha^{(u)}$ for all the four subdomains.}
    \label{fig:inverse_butterfly_update}
\end{figure}

Figure~\ref{fig:inverse_butterfly_update} shows the evolution of the relative $\mathit{l^2}$ norms for temperature and conductivity, the objective functions, three constraints, and the interface parameter ratios across all subdomains. As seen in Fig.~\ref{fig:inverse_butterfly_update}(a), $\mathcal{E}_r^{u}$ converges first, oscillating around $10^{-4}$ after approximately 2500 outer iterations, while $\mathcal{E}_r^{\kappa}$, in Fig.~\ref{fig:inverse_butterfly_update}(b), continues to decrease gradually. Around the same outer iterations, the objective functions, which initially rise, begin to decrease rapidly in Fig.~\ref{fig:inverse_butterfly_update}(c).
The PDE, BC and data constraints, in Figs.~\ref{fig:inverse_butterfly_update}(d-f), reduce to approximately $10^{-6}$, $10^{-6}$, and $10^{-5}$, respectively. Among these, subdomain (1,0) exhibits a slower decay in $\mathcal{C}_B$, but ultimately achieves the lowest $\mathcal{E}_r^{\kappa}$ over the iterations.
Due to the inclusion of the Dirichlet interface operator for $\kappa$, Fig.~\ref{fig:inverse_butterfly_update}(g) shows the ratios of the interface parameters $\alpha^{(\kappa)}$ to $\alpha^{(u)}$. $\dfrac{\alpha^{(\kappa)}}{\alpha^{(u)}}$, as well as the other interface parameter ratios, $\dfrac{\beta}{\alpha^{(u)}}$ and $\dfrac{\gamma}{\alpha^{(u)}}$ in Figs.~\ref{fig:inverse_butterfly_update}(h-i), fluctuate between 1.2 and 1.5 after an initial spike. As noted earlier, the convergence of the relative $\mathit{l^2}$ norms does not necessarily require the convergence of the interface parameter ratios.

\subsection{Parallel Performance}
We assess the parallel efficiency of our domain decomposition method, using the same Poisson's equation problem as outlined in Section \ref{sec:experiment_psn_simple}. Using the Adam optimizer, we evaluate both weak and strong scaling performances, varying the number of subdomains from 2 to 32. The performance numbers are derived from the average time costs recorded over 5 independent trials for each configuration. The numerical experiments are conducted on an Intel Xeon Platinum 8462Y+ architecture. Each subdomain is managed by an MPI process (mpi4py 3.1.5 package) bound to a single CPU core. A Cartesian communicator is created via \textit{Create\_cart} function and the interface information is sent and received via \textit{Isend} and \textit{Irecv} functions, respectively. 

\textcolor{black}{Our parallel implementation is straightforward, and we have not explored more advanced techniques, such as overlapping communication with computation, to improve scalability. Additionally, our performance analysis is limited to a maximum of 32 processes. It is well known that DDM typically do not scale efficiently to a large number of subdomains without a coarse space correction \cite{dolean2015introduction}, which has not been pursued in this study. These limitations should be taken into account when judging the scalability of our approach based on the following results.}

\subsubsection{Weak Scaling Analysis}
The weak scaling analysis aims to maintain a constant workload per processor as the overall problem size increases \cite{Gropp1999MPI}. This approach is particularly suited for memory-bound issues that surpass the memory capacity of a single computational node. For this purpose, each subdomain is responsible for an equal number of random residual points, $16284$, with 128 random points placed on each side of the subdomain, regardless of whether it is a boundary or an interface.

\begin{figure}
\centering
    \subfloat[]{\includegraphics[scale=0.45]{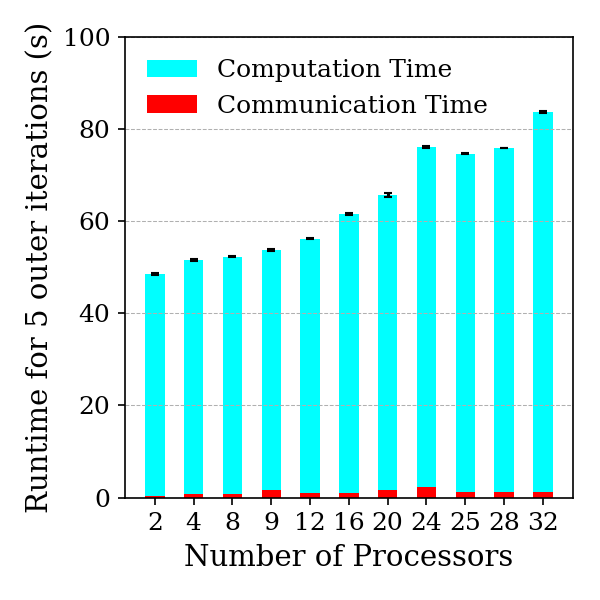}}\quad
    \subfloat[]{\includegraphics[scale=0.45]{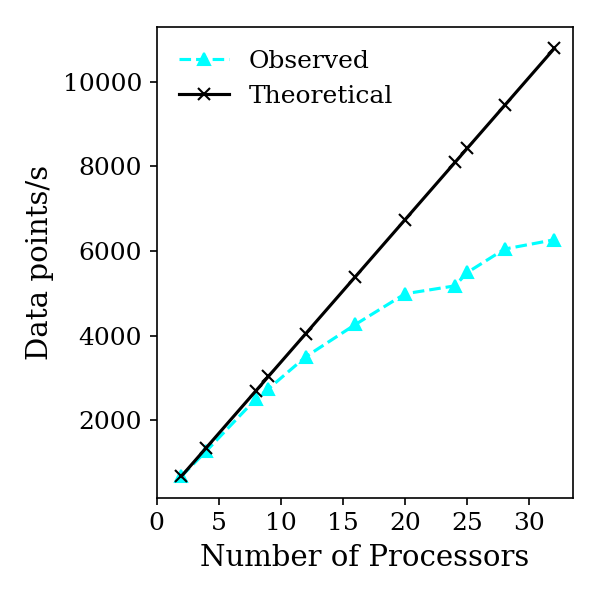}}\quad
    \subfloat[]{\includegraphics[scale=0.45]{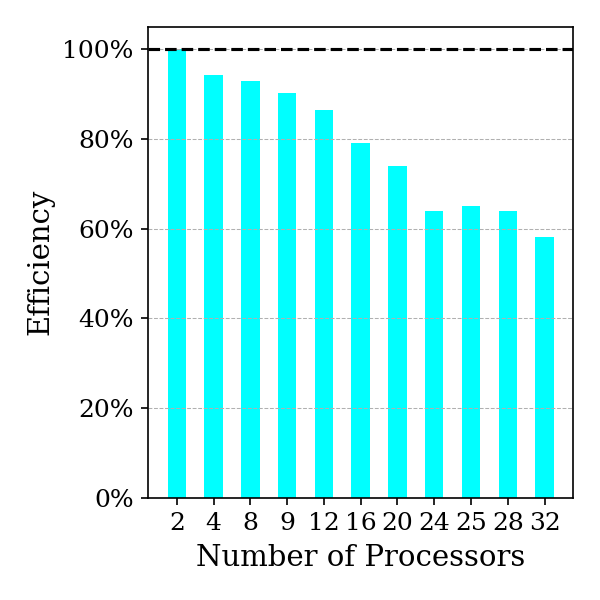}}\quad
    \caption{Weak scaling analysis: (a) computation and communication time as a function of processors employed in the computations, (b) number of data points processed per second, (c) parallel efficiency. The same forward problem given in Section \ref{sec:experiment_psn_simple} is adopted.}
    \label{fig:weak_scaling}
\end{figure}

Figure~\ref{fig:weak_scaling}(a) captures the runtime for five outer iterations, segmented into computation and communication times. 
The total runtime slightly increase, approximately 60 seconds, across various domain decompositions, with computation time significantly exceeding communication time. This predominance is attributed to the design of our domain decomposition training procedure, in which communications between neighboring subdomains are scheduled for after at least 50 inner epochs. This is possible because our PECANN framework aims to infer the parameters of a generalized interface condition constrained by the PDE and its boundary conditions, as opposed to learning the parameters of a neural network constrained by an uncertain interface transmission condition.

Due to the two-dimensional Cartesian decomposition strategy, the number of interfaces per subdomain varies from 1 to 4, resulting in noticeable increases in communication time. This is particularly evident when scaling up from a setup with 2 subdomains (i.e. 2$\times$1 partitioning) to one with 9 subdomains (i.e. 3$\times$3 partitioning).
Although influenced by our community-shared, computing cluster environment, communication time generally stabilizes. However, it exhibits fluctuations as the number of subdomain decompositions increases.
Using a $2\times1$ decomposition as the baseline and excluding executions at boundary and interface points, Figure~\ref{fig:weak_scaling}(b) shows the approximate number of data points processed per second. 
The observed data closely follows theoretical predictions up to 12 processors, highlighting the speed of our algorithm's execution.
Meanwhile, Figure~\ref{fig:weak_scaling}(c) illustrates the weak scaling efficiency, defined as:
\begin{equation}
    E_w = \frac{T_w^{(2)}}{T_w^{(N)}},
\end{equation}
where $T_w^{(2)}$ is the total time required for the baseline with 2 processors, and $T_w^{(N)}$ is the total time required for the simulation with $N$ processors. 
Ideally, the weak scaling efficiency should remain constant at 100\%.
The configurations maintain an efficiency of approximately 80\% up to 16 processors, indicating effective, albeit not optimal, utilization of computational resources.

\subsubsection{Strong Scaling Analysis}
Strong scaling, a key performance metric where the problem size remains fixed while the number of processors increases, is ideal for compute-bound problems \cite{shukla2021parallel_pinns}. We adhere to this strategy by maintaining a constant global domain, which includes a total of $128\times128$ collocation points.
The exact number of points for each subdomain follows the rounding off calculation as outlined in the Section~\ref{sec:experiment_psn_multi}, for different decomposition.

\begin{figure}[ht]
\centering
    \subfloat[]{\includegraphics[scale=0.45]{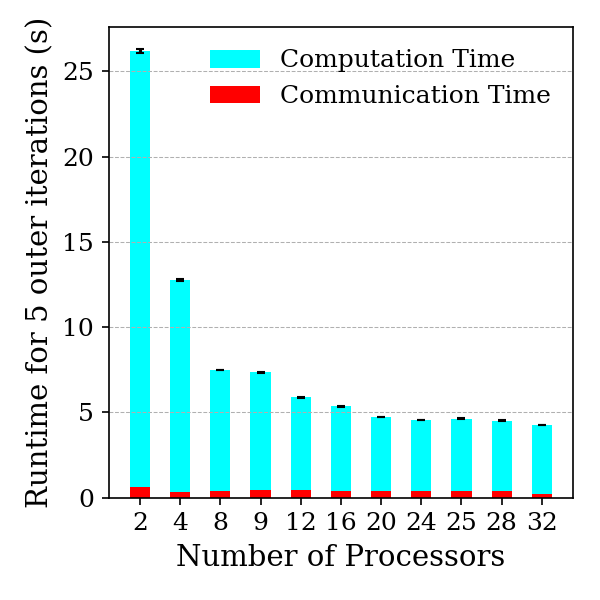}}\quad
    \subfloat[]{\includegraphics[scale=0.45]{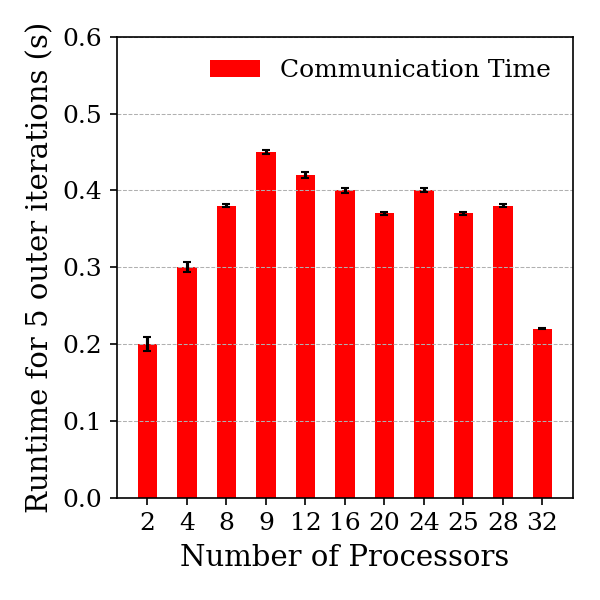}}\quad \\
    \subfloat[]{\includegraphics[scale=0.45]{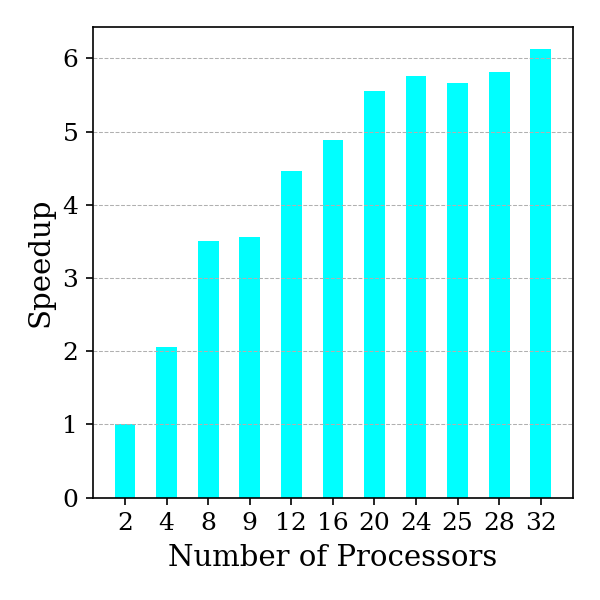}}\quad
    \subfloat[]{\includegraphics[scale=0.45]{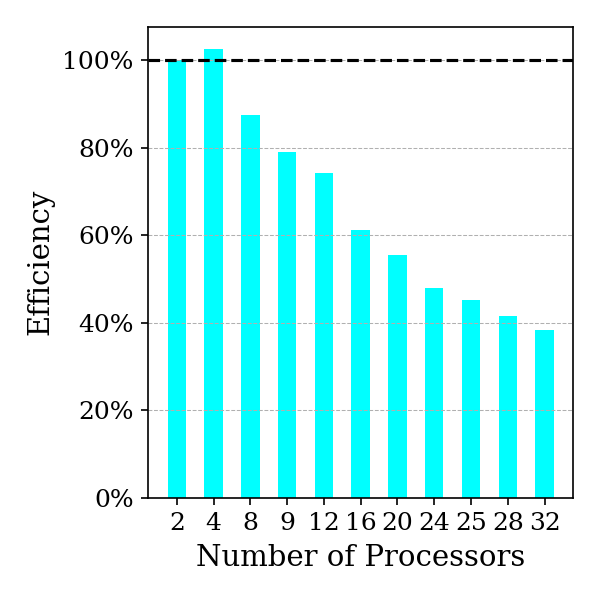}}\quad
    \caption{Strong scaling analysis: (a) Computation and communication time of strong scaling, (b) (zoomed-in) communication time  (c) speedup, (d) parallel efficiency. The same forward problem given in Section \ref{sec:experiment_psn_simple} is adopted.}
    \label{fig:strong_scaling}
\end{figure}

From Figure~\ref{fig:strong_scaling}(a), we observe a consistent reduction in the total time cost, including computation and communication, as the number of processors increases. This decrease, attributable to the reduced workload per processor, suggests an efficient redistribution of workload across a growing number of processors. 
However, focusing on communication time in Fig.~\ref{fig:strong_scaling}(b), the decomposition from 2 to 9 shows a notable rise. This is due to the increasing complexity of the interface data structure. For example, a \(2 \times 1\) decomposition has the same number of interface points per subdomain as a \(2 \times 2\) decomposition in strong scaling. However, to ensure correct exchange direction, interface information like the scalar \(u\) is stored in a matrix, with each column representing an interface. Thus, a \(2 \times 1\) decomposition involves one column vector to be sent and received, while a \(2 \times 2\) decomposition involves a matrix with two columns, requiring a \textit{for} loop and conditional statements in the code.
The \(3 \times 3\) decomposition marks the point where a subdomain contains four interfaces, resulting in the highest communication time observed in Figure~\ref{fig:strong_scaling}(b). After this point, the communication time decreases due to fewer interface points on the subdomains. We have not pursued more sophisticated ways to store interface data, which could have resulted in better performance in fetching data and sending. Nonetheless, the performance of strong scaling is primarily determined by computation time, which benefits from the reduced number of residual points in each subdomain with further partitioning.

Using the total time consumed, $T_s^{(N)}$, for each simulation with $N$ processors, we calculate the corresponding speedup as follows:
\begin{equation}
    s^{(N)} = \frac{T_s^{(2)}}{T_s^{(N)}},
\end{equation} 
as shown in the Figure~\ref{fig:strong_scaling}(c).
The trend becomes more apparent by the nearly linear increase in speedup relative to the $2\times1$ baseline. The trend holds up consistently up to 20 processors.
Strong scaling efficiency is further computed using the formula:
\begin{equation}
    E_s = \frac{s^{(N)}}{N/2},
\end{equation}
The results are shown in Figure~\ref{fig:strong_scaling}(d). 
Ideally, the weak scaling efficiency should remain constant at 100\%.
The efficiency graph shows a gradual decline from the ideal 100\% efficiency level, except the super-linearity of the $2\times2$ decomposition.
Such trend is not too surprising for strong scaling analysis because of the reduced workload per processor. While the communication times are marginal, this decline is potentially exacerbated by the computational overhead associated with the Adam optimizer that does scale with increased processor count. Improving strong scaling performance could potentially be enhanced by leveraging advanced features of the MPI library, particularly those optimized for multi-core CPU architectures. However, exploring these enhancements falls outside the scope of our current study.

\section{Conclusion}
Domain decomposition methods (DDMs) are crucial for advancing physics-informed or constrained machine learning techniques to solve large-scale partial differential equations (PDEs) in a distributed setting. We introduce a novel Schwarz-type, non-overlapping DDM to learn the solution of forward and inverse PDE problems using physics-informed neural networks. Specifically, we leverage the physics and equality-constrained artificial neural networks (PECANN) framework \cite{PECANN_2022} in a novel way, formulating a constrained optimization problem for each subdomain. An augmented Lagrangian method with a conditionally adaptive strategy, unique to this study, is employed to transform each constrained optimization problem into a dual unconstrained problem. 

Unlike the original PECANN method, which uses boundary conditions exclusively to constrain the PDE loss term, our DDM allows both the boundary conditions and governing PDE to jointly constrain a generalized interface loss term within the PECANN framework. This formulation also differs from other physics-informed neural network approaches with DDM, where subdomain interface conditions are used alongside boundary conditions to softly constrain the PDE solution.

Our proposed interface transmission condition is formulated as a linear combination of Dirichlet, Neumann, and tangential derivative continuity operators, each with subdomain-specific interface parameters to be optimized. \textcolor{black}{Crucially, we derive an approximate interface loss term based on the mean squared error estimator in which the cross product terms of the interface operators are neglected, resulting in a formulation (i.e. Eq. \ref{eq:app_interface_loss}) more suitable for optimization. The PDE solution is then learned via a feed-forward neural network while minimizing this approximate interface loss term simultaneously. Subdomains are independently trained using separate neural networks with the same architecture for a fixed number of epochs before synchronizing with neighboring subdomains, thus minimizing communication overhead while promoting the independent learning of interface parameters. We find that the augmented Lagrangian method with the proposed conditionally adaptive strategy plays a crucial role in the optimization process. It is also worth noting that our proposed interface transmission condition may not be directly applicable as a DDM for classical numerical methods, such as finite differences, due to the fundamentally different approach to predicting the solution. In our approach, the solution is learned by minimizing the approximate interface loss term, rather than by directly enforcing the interface transmission conditions.} 

We have applied our method to learn the solution of a variety of forward and inverse PDE problems in parallel with up to 64 subdomains using a Message Passing Interface model. A promising aspect of our method, facilitated by the proposed approximate interface loss term, is its capability to learn solutions to both Poisson's and Helmholtz equations, including those with high wavenumber as well as complex-valued solutions. This unified framework for tackling PDEs with fundamentally different characteristics stands in sharp contrast to the state-of-the-art in conventional DDMs, which usually require distinct interface transmission conditions for equations like Poisson’s and Helmholtz. It is also important to note that, despite the common underlying framework, the objective functions remain specific to each PDE due to the PDE-constrained nature of minimizing a subdomain-specific interface loss term. \textcolor{black}{Additional numerical experiments would be valuable for thoroughly evaluating the potential of our approach in solving PDEs with varied characteristics and increasing complexity.}

\textcolor{black}{Our method exhibits consistent generalization capabilities as the number of subdomains increases; however, this trend is inherently limited and cannot continue indefinitely. Our parallel performance analysis has so far been limited to 32 processes, and further testing is needed to assess parallel scalability beyond this number. It is well-documented that Schwarz-type domain decomposition methods often experience scalability challenges when applied across a large number of subdomains \cite{dolean2015introduction}, which we expect to be the case for our method as well. In such cases, introducing a coarse space or level correction is recommended to improve scalability of parallel solutions with numerous subdomains. A promising strategy to address this challenge could be the approach proposed by \citet{coarse_correction_deepddm} in the context of deep DDM \cite{deep_ddm_2020}. In our approach, a coarse level can be introduced by representing the solution with a neural network that spans the entire domain. Each subdomain can communicate information sampled from its interior points to the coarse-space neural network in a data-driven manner, which helps to constrain the solution at the coarse level. The augmented Lagrangian method within our PECANN framework is particularly well-suited for assimilating such data into the learning process. When solving within each subdomain, the generalized interface conditions can be enriched by incorporating information from the global coarse-space solution. This information can be weighted appropriately, with the weight being learned alongside other subdomain-specific interface parameters.} 

\textcolor{black}{It is essential to highlight the limitations of the proposed method. Solving PDEs using neural networks is significantly more time-consuming than conventional numerical methods, particularly for forward problems. This time burden becomes even greater for complex PDEs with multi-scale or high-wavenumber features in their solutions, where prolonged training times are necessary to achieve accurate predictions. Furthermore, neural network-based methods often require multiple runs to ensure solution accuracy, further increasing computational demands. Although we observe that increasing the number of subdomains enhances the accuracy of learned solutions due to our method’s generalization capabilities, this improvement has inherent limits. A coarse-space correction could potentially address this limitation, which we plan to explore in future work, along with an in-depth scalability assessment involving a larger number of subdomains and applications to complex three-dimensional problems.}

\section*{Acknowledgments}
This material is based upon work supported by the National Science Foundation under Grant No. (1953204) and by University of Pittsburgh Center for Research Computing, RRID:SCR\_022735, through the resources provided. Specifically, this work used the H2P cluster, which is supported by NSF award number OAC-2117681.

\section*{Declaration of competing interests}
\noindent The authors declare that they have no known competing financial interests or personal relationships that could have appeared to influence the work reported in this paper.

\section*{Data availability}
\noindent All the code related to this study is available as open-source software at this site \url{https://github.com/HiPerSimLab/PECANN-DDM}.

\section*{Declaration of generative AI and AI-assisted technologies in the writing process}

\noindent During the preparation of this work the author(s) used ChatGPT in order to to improve language and readability. After using this tool/service, the author(s) reviewed and edited the content as needed and take(s) full responsibility for the content of the publication.

\section*{Appendix A. Performance comparison of different approaches}
\subsubsection*{A.1. Poisson's equation with a simple step source term}
We adopt the same problem setting outlined in \citet{hu_xpinn_2022} to compare the accuracy of our proposed method against the XPINN approach \cite{jagtap2020xpinns}, as well as evaluate the different interface loss function formulations as introduced in Section \ref{sec:interface_loss}. 
The Poisson equation is considered over the domain \(\Omega = \{ (x,y) \mid 0 \leq x, y \leq 1 \}\), with Dirichlet boundary conditions and a step-function source term \(s\) defined over the subdomain \(\Omega_1 = \{ (x,y) \mid 0.25 \leq x, y \leq 0.75 \}\):
\begin{equation}
    \begin{aligned}
        g(x, y) &= 0, \quad \forall (x, y) \in \partial \Omega  \\
        s(x, y) & = \begin{cases} 
        1, & \text{if } (x, y) \in \Omega_1, \\
        0, & \text{elsewhere}
        \end{cases}
    \end{aligned}
\end{equation}
This setup benefits naturally from the decomposition of the domain into the interior subdomain \(\Omega_1\) with the non-zero constant source term and the surrounding subdomain \(\Omega_2\).

In addition to the deep neural network model with 9 hidden layers and 20 units per layer used in \citet{hu_xpinn_2022}, we implement a shallower network for each decomposed subdomain with 2 hidden layers and 20 units per layer, using the \(\tanh\) activation function. We use L-BFGS optimizer with a learning rate of \(0.1\) for model parameters and Adam optimizer for interface parameters. The training dataset consists of 400 residual points, 80 boundary points, and 80 interface points, consistent with \citet{hu_xpinn_2022}. The trained models are tested on 1,002,001 domain points. In our DDM, we set the number of outer iterations to 1000 for interface communication and the minimum number of inner epochs to 50 for independent training of neural networks before any communication.

\begin{table}
\small
    \centering
    \caption{Relative $\mathit{l^2}$ norms of error for the solution of Poisson's equation with a step source term. Data for PINN and XPINN methods are from \cite{hu_xpinn_2022}.}
    \begin{tabular}{lccc}
        \hline
        Algorithm & (layers, units/layer) & Relative $\mathit{l^2}$ norm & Mean time cost(s) \\
        \hline
        PINN \cite{hu_xpinn_2022} (no DDM) & (9, 20) & $5.55e-2 \pm 2.94e-2$ & $1.58e+3$\\
        PECANN (no DDM) & (9, 20)& $5.21e-2 \pm 6.76e-3$ & $1.63e+3$\\
        PECANN (no DDM) & (2, 20)& $1.04e-2 \pm 2.40e-3$ & $9.41e+2$\\
        \hline
        XPINN3 \cite{hu_xpinn_2022} & (9, 20) & $1.11e-1 \pm 1.56e-2$ & --\\
        \textbf{Proposed DDM, Eq.~\ref{eq:objective_final}} & (9, 20) & $8.51e-3 \pm 3.25e-3$ & $3.74e+3$ \\
        \textbf{Proposed DDM, Eq.~\ref{eq:objective_final}} & (2, 20) & $\textcolor{black}{5.91e-3 \pm 2.49e-3}$ & $1.29e+3$\\
        DDM (abs. interface loss, Eq.~\ref{eq:abs_interface_loss}) & (2, 20) & $4.81e-2	\pm 5.83e-3$ & $4.90e+3$\\
        DDM (full interface loss, Eq.~\ref{eq:full_interface_loss}) & (2, 20) & \textcolor{black}{$2.23e+3 \pm 1.45e+3$} & $2.19e+3$\\
        \hline
    \end{tabular}
    \label{tab:benchmark_poisson_l2}
\end{table}

The first three rows of Table~\ref{tab:benchmark_poisson_l2} compare the relative $l^2$ norms of the error and its standard deviation over five trials for the PINN and PECANN methods without domain decomposition. In the next three rows, we compare the accuracy of our DDM with the best results obtained by the XPINN approach as reported in \cite{hu_xpinn_2022}. Also included in Table~\ref{tab:benchmark_poisson_l2} are the relative error levels produced by the use of three different formulations of the interface loss term in our DDM.

In contrast to the PINN approach, where a 9-layer architecture was adopted, a simpler 3-layer architecture is sufficient with the PECANN approach to significantly reduce the mean relative $l^2$ error to nearly $10^{-2}$, along with a smaller deviation. 
With a 9-layer architecture, the performance of PECANN is slightly worst than the 3-layer performance and only slightly better than the performance of PINN with 9-layer. This drop in performance occurs despite achieving lower converged values on the objective function in PECANN, indicating that the complexity of this deep network, combined with insufficient sampling, may lead to overfitting. 

From Table~\ref{tab:benchmark_poisson_l2}, we observe that the XPINN approach with the best-tuned weights (i.e. XPINN3 in \cite{hu_xpinn_2022}) experiences a reduction in accuracy compared to PINN predictions without any domain decomposition. For this problem, \citet{hu_xpinn_2022} concluded that all three variants of the XPINN perform worse than PINN because they perform poorly either on the boundary or on the interface. In contrast, our proposed DDM with the approximate interface loss (Eq.~\ref{eq:objective_final}), outperforms PECANN, achieving the lowest relative $l^2$ error, on the order of magnitude $10^{-3}$.
We attribute the superior performance of our DDM over PECANN with no decomposition to the robustness of the underlying constrained optimization formulation, which uses an adaptive augmented Lagrangian method that weighs each constraint in a principled fashion through adjusting Lagrange multipliers and the penalty method of each constraint.

\begin{figure}[t]
    \centering
    \subfloat[]{\includegraphics[scale=0.5]{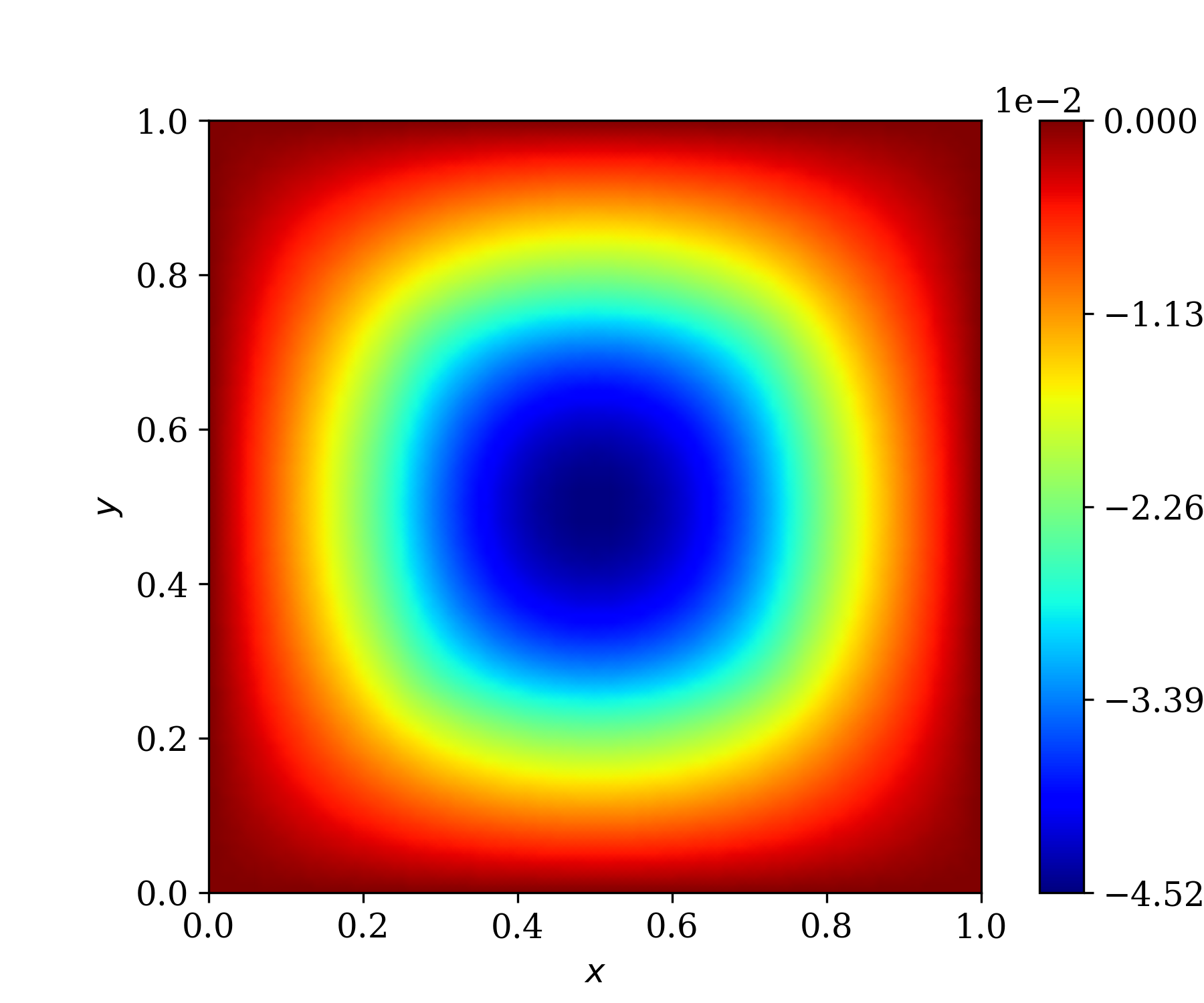}}
    \subfloat[]{\includegraphics[scale=0.5]{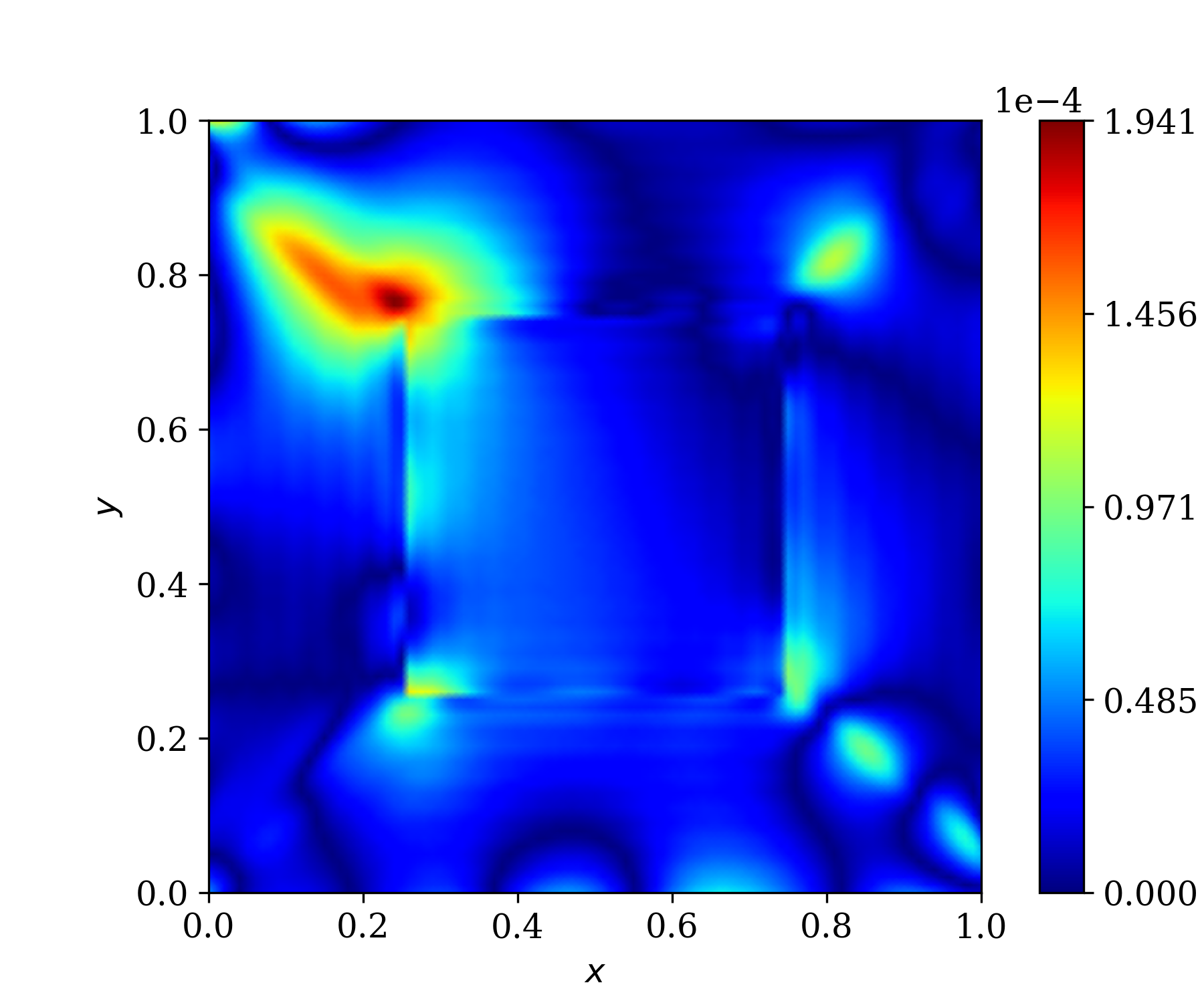}}
    \caption{Poisson's equation with a step source term: (a) predicted solution obtained with the proposed DDM with the approximate interface loss function (Eq. \ref{eq:objective_final}), (b) corresponding absolute point-wise error.}
    \label{fig:benchmark_poisson_solution}
\end{figure}

Figure~\ref{fig:benchmark_poisson_solution} presents the predicted solution and the absolute point-wise error distribution for the best trial of the proposed DDM with the approximate interface loss function, respectively. The results from Fig.~\ref{fig:benchmark_poisson_solution}(b) reveal that the maximum error is reduced to \(1.941 \times 10^{-4}\), with the primary error regions concentrated near the interface. This observation validates the satisfaction of boundary and physics constraints over the interface condition.

To validate the effectiveness of the approximate interface loss function (Eq.~\ref{eq:objective_final}), we also tested the absolute (Eq.~\ref{eq:abs_interface_loss}) and the full interface loss functions (Eq.~\ref{eq:full_interface_loss}) in the same framework. From Table \ref{tab:benchmark_poisson_l2} we observe that the absolute interface loss function demonstrates a good level of accuracy, supporting the feasibility of using the proposed interface conditions. However, the full interface loss function results in substantial errors. Given these findings, we employ the approximate interface loss function (Eq. \ref{eq:objective_final}) in the proposed domain decomposition method.

The fourth column of Table \ref{tab:benchmark_poisson_l2} shows the total training time in seconds. Although the training time for DDM is longer than for PECANN without DDM, this is expected. The DDM formulation is more complex than PECANN without DDM. In DDM, the objective function is based on the interface conditions, constrained by the PDE and boundary conditions, while in PECANN without DDM, the objective function is simply the residual form of the PDE constrained by the boundary conditions. Additionally, in DDM, interface parameters must be inferred, which is not required in PECANN without DDM. The advantages of DDM become more evident in complex problems, where without DDM, a much deeper network, more training epochs, and additional collocation points would be required to achieve acceptable results.

\clearpage


\end{document}